%% file: main.tex
\documentclass[fleqn,10pt]{wlscirep}

\usepackage[utf8]{inputenc}
\usepackage[T1]{fontenc}

\title{Language-encoded network topology enables large language models to reason about complex networks}

\author[1]{Ucchwas Talukder Utsha}
\author[1]{Sakib Mostafa}
\author[2,3,4]{James Zou}
\author[1,*]{Md Tauhidul Islam}
\affil[1]{Department of Radiation Oncology, Stanford University, Stanford, California, USA}
\affil[2]{Department of Electrical Engineering, Stanford University, Stanford, California, USA}
\affil[3]{Department of Biomedical Data Science, Stanford University School of Medicine, Stanford, California, USA}
\affil[4]{Department of Computer Science, Stanford University, Stanford, California, USA}

\affil[*]{correspondence: tauhid@stanford.edu}

\usepackage{array}
\usepackage{booktabs}
\usepackage{multirow}
\usepackage{float}
\usepackage{placeins}
\usepackage{algorithm}
\usepackage{algorithmic}
\usepackage{tikz}
\usetikzlibrary{
  positioning,
  arrows.meta,
  shapes.geometric,
  fit,
  backgrounds,
  calc
}

\usepackage{subcaption}

\hypersetup{
  colorlinks=true,
  allcolors=blue!55!black,
  breaklinks=true
}

\definecolor{accent}{HTML}{2A6FB0}
\definecolor{bgInput}{HTML}{DCEBF7}
\definecolor{bgAnalyze}{HTML}{D8EEE4}
\definecolor{bgCompile}{HTML}{FBE8CE}
\definecolor{bgPipe}{HTML}{E7E2F0}
\definecolor{bgLLM}{HTML}{F8DDD7}
\definecolor{bgOracle}{HTML}{D6EED9}
\definecolor{inkEdge}{HTML}{5B6068}

\newcommand{\glyph}[1]{\texttt{\small #1}}

\newcommand{\bioglyph}{\textsc{BioGlyph}}

\begin{abstract}
\input{sections/abstract}
\end{abstract}

\begin{document}

\flushbottom
\maketitle
\thispagestyle{empty}


\section*{Introduction}
\input{sections/intro}

\section*{Results}
\input{sections/results}

\section*{Discussion}
\input{sections/discussion}

\section*{Methods}
\input{sections/methods}

\section*{Data availability}

Every network we analyzed is public. Section~\ref{sec:suppresults} of the Supplementary Information describes the four networks of the main text, and Table~\ref{tab:datasets} gives their sizes. The protein interaction networks are STRING v12\cite{szklarczyk2023string} (the budding yeast, human and \emph{E.\ coli} physical subnetworks), HuRI\cite{luck2020reference} and PP-Pathways\cite{agrawal2018large}. ego-Facebook, email-Eu-core, Wiki-Vote and the Enron email graph\cite{klimt2004enron} come from the SNAP collection\cite{leskovec2014snap}, and ChCh-Miner and DG-AssocMiner come from BioSNAP\cite{zitnik2018biosnap}. The western US power grid analyzed by Watts and Strogatz\cite{watts1998collective} comes from Newman's network data collection. Cora and CiteSeer are the standard citation networks\cite{sen2008collective}, and Amazon-Photo and Coauthor-CS follow Shchur et al.\cite{shchur2018pitfalls}; we load those four through PyTorch Geometric\cite{fey2019fast} and ogbn-arxiv through the Open Graph Benchmark package\cite{hu2020open}. The disease comorbidity network is HuDiNe\cite{hidalgo2009dynamic}. The glioblastoma patient-similarity network is built by similarity network fusion\cite{wang2014similarity} from the public TCGA cohort\cite{weinstein2013cancer}, and the glioblastoma single-cell network is built from the Smart-seq2 expression matrix deposited under GEO accession GSE131928. The knockout screen runs on the Reactome functional interaction network\cite{wu2010human}. We validated the roles against external labels, none of which we showed to any model: SGD gene essentiality\cite{cherry2012saccharomyces}, DepMap 24Q4 gene effect\cite{tsherniak2017defining,dempster2021chronos}, GTEx tissue expression\cite{gtex2020gtex} and approved gene symbols from HGNC\cite{tweedie2021genenames}. We release the compiled descriptions, the question sets with their exact answers, every stored model reply and the scored result tables behind every number in this paper together with the analysis code.

\section*{Code availability}
The code used for all analyses is available to editors and reviewers through a private Code Ocean capsule. The capsule contains the complete BioGlyph implementation, including data preprocessing, representation construction, model training and evaluation pipelines. An interactive browser-based demonstration is available at \url{https://islamlab.org/bioglyph}, allowing users to reproduce the worked examples and apply BioGlyph to their own networks without local installation. Upon publication, the Code Ocean capsule will be made publicly available under a persistent digital object identifier.

\FloatBarrier
\bibliography{references}


\clearpage
\setcounter{figure}{0}
\setcounter{table}{0}
\renewcommand{\thefigure}{S\arabic{figure}}
\renewcommand{\thetable}{S\arabic{table}}
\renewcommand{\theHfigure}{S\arabic{figure}}
\renewcommand{\theHtable}{S\arabic{table}}
\setcounter{section}{0}
\renewcommand{\thesection}{S\arabic{section}}

\section*{Supplementary Information}
\suppressfloats[t]

\input{sections/supplementary}

\FloatBarrier

\end{document}

%% file: sections/abstract.tex
Networks describe diverse systems in biology and beyond, from protein interactions and social relationships to power grids and citation records. Reasoning about such systems requires understanding their structure: which elements are central, which connections bridge otherwise separate communities, how the network is organized, and how its structure changes when elements are removed. Although large language models (LLMs) excel at reasoning over natural language, they often struggle with such structural questions when networks are represented as edge lists, sentences or tables of numerical measurements, because the structural meaning of these representations must be inferred. Here we introduce BioGlyph, a method that compiles network topology into an interpretable and transferable language of structural roles. BioGlyph combines graph-partitioning algorithms and structural measurements to identify roles such as hubs, community cores and cross-community connectors, and uses fixed rules to translate their algorithmic signatures into a universal vocabulary. The resulting representation describes each network element through its structural role, supporting evidence and semantic consequences, while leaving both the original network and the LLM unchanged. Across twenty networks spanning five different domains including biological, social and information systems, BioGlyph substantially improves the ability of open LLMs to answer structural reasoning questions, outperforming edge-based, numerical and learned representations by up to 26 percentage points in system accuracy. Ablation experiments show that the improvement comes from explicitly encoding structural roles in semantically interpretable terms by Bioglyph. The performance improvement becomes even more prominent in dense, community-structured networks and diminishes in sparse networks whose topology can be more readily inferred from direct textual representations. Applied to a budding-yeast protein-interaction network, BioGlyph also exposes biologically meaningful organization: cross-community connectors are enriched for essential genes, whereas peripheral proteins are depleted. BioGlyph thus provides a common, interpretable representation through which both language models and scientists can reason about complex network structure.

%% file: sections/intro.tex
\begin{figure}[!t]
\centering
\includegraphics[width=\linewidth]{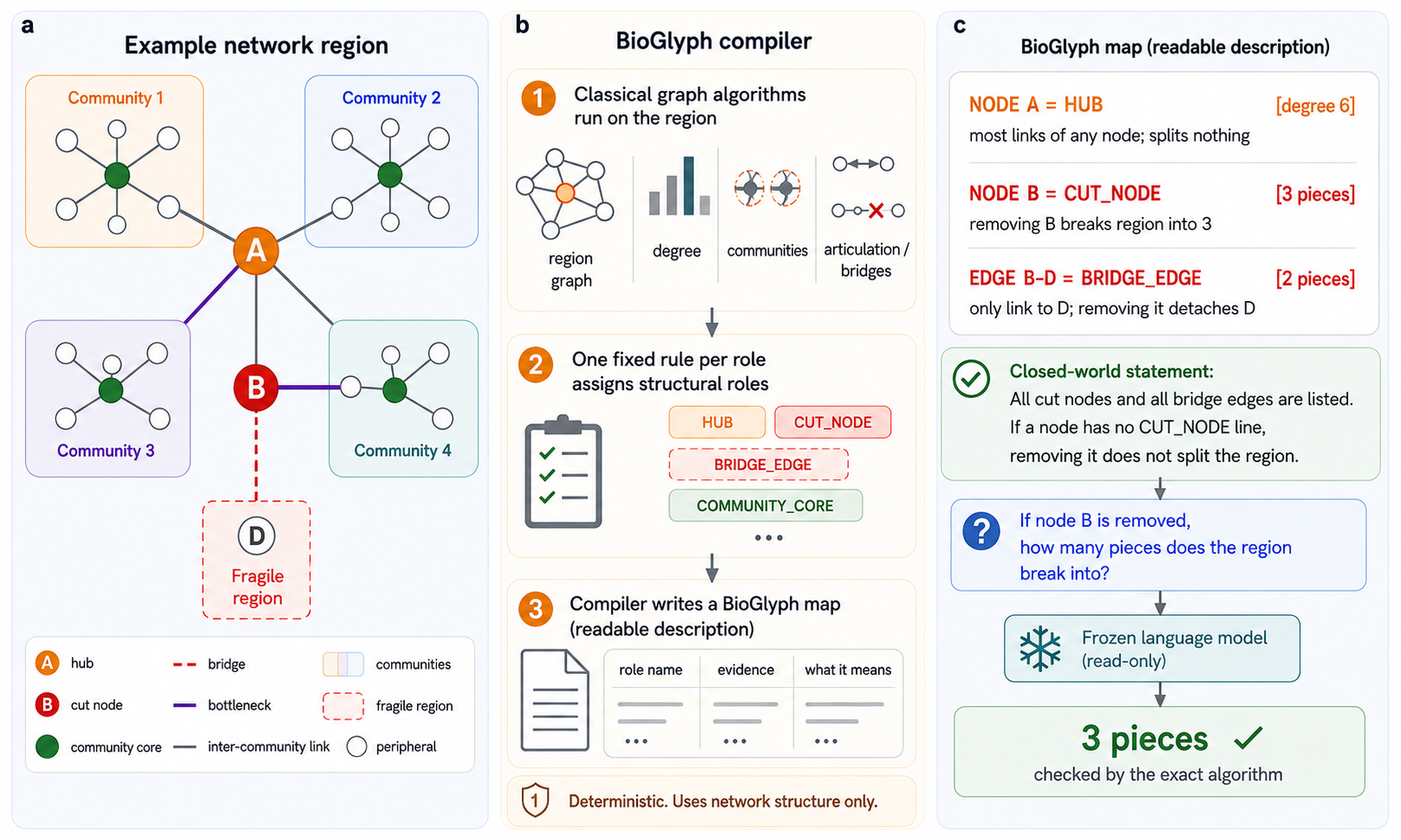}
\caption{\textbf{BioGlyph translates network structure into interpretable roles.}
\textbf{a}, An illustrative network of 31 nodes organized into five communities. BioGlyph identifies complementary structural roles, including hubs, cut nodes, bridges, community cores, cross-community connectors and bottleneck links. For example, node A is a highly connected hub, whereas removal of node B separates the network into three components. The dashed outline indicates a fragile community connected to the rest of the network through a single link.
\textbf{b}, BioGlyph applies classical network-analysis algorithms and fixed rules to identify eleven structural roles. The rules use either exact perturbation tests or thresholds derived from the network structure. The procedure is deterministic for a given random seed and uses network topology alone, without node labels or biological annotations.
\textbf{c}, BioGlyph converts the resulting structural information into a human- and language-model-readable description. Each entry specifies the structural role, the evidence supporting it and, where relevant, the consequence of removing the node or link. The resulting description provides an interpretable summary of network organization that can be used by a frozen language model to answer questions about connectivity, importance and perturbation. Answers can then be checked against exact graph algorithms.}
\label{fig:pipeline}
\end{figure}

Networks provide a natural way to describe how parts of a system are connected. Protein interactions inside a cell, gene-regulatory circuits, and connections in the brain can all be represented as graphs. The same representation is used for power grids, transport systems, citation records, the World Wide Web, and social networks. In a graph, nodes represent the parts of a system and edges represent the relationships between them. Decades of network science have shown that the behavior of a system depends not only on the individual nodes, but also on the pattern of connections among them\cite{watts1998collective,barabasi1999emergence}. Several structural patterns appear across very different networks. Hubs have unusually many connections. Cut nodes, or articulation points, separate a connected network when removed. Bridges provide the only link between otherwise disconnected regions. Many networks also form dense communities connected by a small number of nodes that link one community to another\cite{girvan2002community,fortunato2010community}. Such structural features can have direct scientific meaning. Network position has been linked to gene essentiality\cite{jeong2001lethality,yu2007importance}, infrastructure vulnerability\cite{albert2000error}, and the flow of information through social systems. Understanding a network therefore requires more than knowing which nodes are connected. The important question is often what those connections imply for the system, especially when a node or edge is removed.

Large language models are strong general-purpose systems for reasoning over text. Models trained on large text collections can learn new tasks from examples in a prompt\cite{brown2020language} and can solve multi-step problems through explicit reasoning\cite{wei2022chain}. Network reasoning presents a different challenge. When a graph is written as an edge list or as sentences describing which nodes are connected, a language model must recover the relevant structure from the text before answering the question. The model must determine, for example, whether a node connects two regions, whether removing an edge separates the graph, or which candidate is most important for connectivity. Language models often make errors on such problems even when standard graph algorithms can compute the answers exactly\cite{wang2023can,fatemi2024talk}.

Perturbation-based questions make this limitation particularly clear. Can two nodes still reach each other after a a third node is removed? Does deleting a single edge split the network? Which node causes the greatest disruption when removed? Related questions arise when a router fails in a communication system or a connection is disrupted in the brain. They are especially important in biomedicine, where many questions can be framed as network perturbations. What happens when a gene is upregulated or downregulated? Which protein is most important for maintaining a pathway? Which component connects two biological processes? Which proteins are most vulnerable to disruption? Answering these questions requires understanding how each component is positioned within the wider biological system. A list of interactions alone does not explain why a protein is important or predict what will happen when it is perturbed. In each setting, the effect of a perturbation depends on the structure of the network.

Current approaches generally provide network information to LLMs either as connections or as learned numerical representations. Connections can be written as edge lists, structured data or natural-language descriptions \cite{guo2023gpt4graph,wang2023can,fatemi2024talk}. These formats preserve the underlying network, but the model still has to infer important structural properties from the connections. Other approaches learn numerical representations of nodes or network regions and provide these representations to language models \cite{perozzi2024let,tang2024graphgpt,chen2024llaga}. Although such representations can capture complex patterns, they are difficult for researchers to interpret. A numerical representation does not readily explain why a protein is important, which other proteins it connects, or what might happen when it is removed. These limitations raise a simple question: can network structure be expressed directly in terms that both language models and researchers can understand?

Here we introduce BioGlyph, a method that translates network structure into a simple language of structural roles (Fig.~\ref{fig:pipeline}). BioGlyph first uses established network-analysis algorithms to identify properties such as hubs, community cores, cut nodes, bridges and proteins that connect different communities. It then uses fixed rules to describe these properties in words. Each protein or interaction can therefore be described not only by its structural role, but also by the measurements supporting that role and by what its position means for the surrounding network. For example, rather than providing only the number of interactions of a protein, BioGlyph can describe it as a hub or as a connector between two communities and explain how the network would change if that protein were removed. The method uses only the network itself and does not require biological labels or task-specific training. The LLM is also kept unchanged. BioGlyph therefore changes how network information is presented rather than changing the underlying model. By expressing network properties in words, BioGlyph makes the results of network analysis easier to interpret and incorporate into a broader reasoning process.

We evaluated BioGlyph across eight networks representing biological, social and information systems. We asked LLMs questions about network connectivity, importance and the consequences of removing nodes or interactions, while keeping the underlying network and model fixed and changing only how the network was represented. We compared BioGlyph with edge lists, natural-language descriptions, numerical network measurements and learned network representations. BioGlyph consistently improved performance on questions that required understanding network structure, and approached the performance of a graph neural network trained specifically on the task. The improvement was particularly pronounced in dense networks with many interconnected communities, where directly reading the individual connections is difficult. In sparse networks, where the connections are already relatively easy to read, the advantage was smaller. These results indicate that explicitly describing meaningful structural properties can make complex network information easier for LLMs to use.

We next asked whether the same structural descriptions could reveal biologically meaningful patterns independently of their use for language-model reasoning. In the budding-yeast protein-interaction network, BioGlyph identified structural roles using network connectivity alone, without using gene-essentiality annotations. Proteins that connected different communities were substantially more likely to be essential, whereas peripheral proteins were less likely to be essential. Cross-community connectors were also more strongly enriched for essential genes than proteins selected solely by degree at the same set size. These findings suggest that the structural vocabulary captures biologically meaningful organization rather than simply providing a convenient format for prompting an LLM.

Together, these results show that network structure can be translated into a common language that is accessible to both computational models and researchers. BioGlyph preserves information obtained from established network-analysis methods while making important structural relationships easier to interpret, communicate and use in biomedical reasoning. This provides a general framework for using LLMs with biological networks and, more broadly, for connecting quantitative network analysis with human-readable scientific interpretation.

%% file: sections/results.tex

\subsection*{BioGlyph improves reasoning over the yeast interactome and identifies biologically meaningful roles}

We first evaluated BioGlyph on STRING-Yeast, a budding-yeast protein-interaction network comprising 3{,}384 proteins and 43{,}030 physical associations\cite{szklarczyk2023string}. Independent gene-essentiality annotations enabled us to test whether roles inferred solely from network topology correspond to biological function. The benchmark comprised 156 questions, 10.3\% of which referred to a protein absent from the network and therefore required the model to state that the question could not be answered. For each question, we retrieved the network region surrounding the named proteins and represented it in six forms: BioGlyph; no network information (R0); an edge list (R1); the same edges expressed as sentences (R2); a table of raw graph measurements (R3); and structural roles learned using GraphSAGE (R4). GraphSAGE was the strongest of the four graph encoders evaluated; results for all four are reported in Section~\ref{sec:suppresults}. Each representation was provided to the same two frozen 8-billion-parameter language models, Qwen3-8B and Llama-3.1-8B. Answers were graded against exact graph-algorithm outputs rather than by another language model.

\begin{figure}[!t]
\centering
\includegraphics[width=\linewidth]{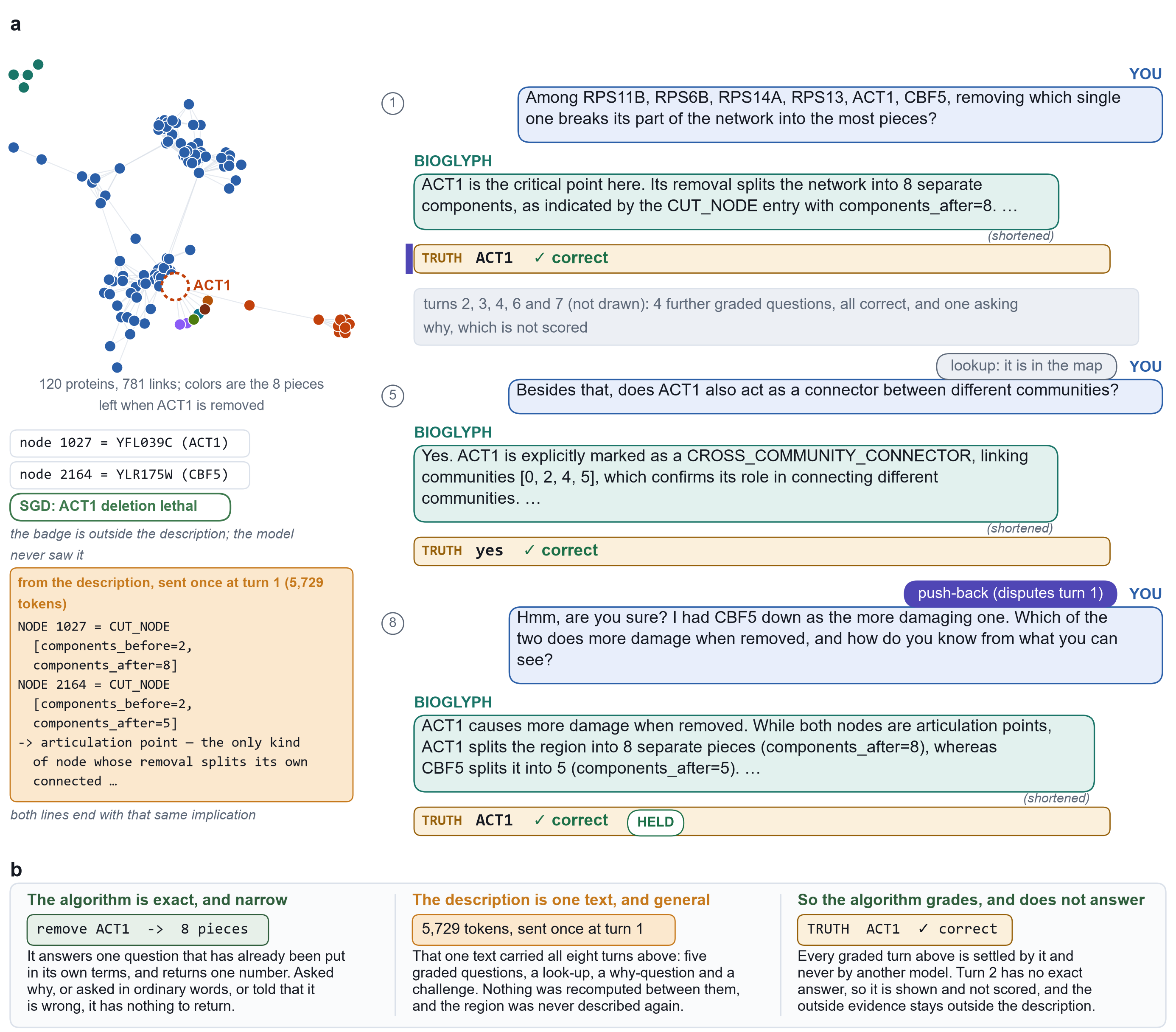}
\caption{\textbf{A single BioGlyph description supports multi-turn reasoning about the yeast interactome.}
\textbf{a} A STRING-Yeast region containing 120 proteins and 781 links was compiled once and provided to a frozen Qwen3-8B. Nodes are coloured according to the eight connected components that remain after removal of ACT1. Three of eight conversation turns are shown. Turn 1 asks which of six proteins would cause the greatest disruption if removed; turn 5 asks whether ACT1 also connects distinct communities; and turn 8 challenges the initial answer by proposing CBF5 instead. Gene names are shown for readability: node 1027 corresponds to YFL039C (ACT1), and node 2164 to YLR175W (CBF5). The model received only node identifiers and did not receive the SGD annotation. Responses are taken from stored model outputs and shortened only at sentence boundaries. Each graded response was evaluated against an exact graph-algorithm output.
\textbf{b} The same BioGlyph description was used throughout all eight turns. Exact graph algorithms supplied the grading reference; no language model graded another model. Further multi-turn examples are provided in Section~\ref{sec:talking}.}
\label{fig:qa}
\end{figure}

Figure~\ref{fig:qa} illustrates how a single BioGlyph description can support sustained reasoning about one network region. We compiled a region containing 120 proteins and provided the description once, with the first question. Qwen3-8B identified ACT1 as the protein whose removal would cause the greatest disruption, citing the stated consequence that its removal leaves eight connected components. The model subsequently used the same description to answer four further graded questions, recognized ACT1 as a cross-community connector and retained its original answer when challenged with the alternative suggestion that CBF5 would be more disruptive. All graded responses agreed with the corresponding exact graph-algorithm outputs. ACT1 is independently annotated by SGD as essential, although this annotation was not available to the model.

\begin{figure}[!t]
\centering
\includegraphics[width=\linewidth]{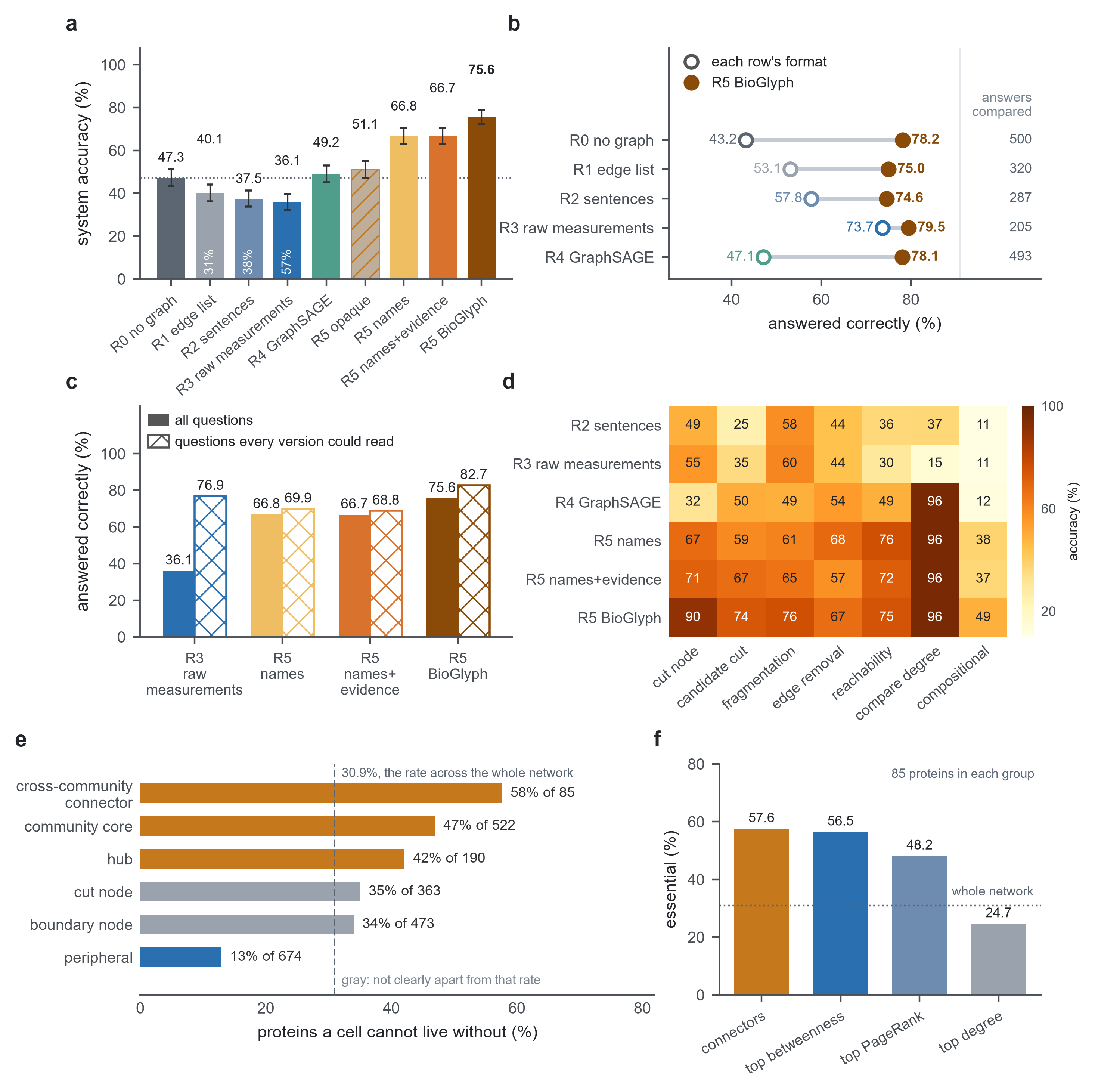}
\caption{\textbf{BioGlyph improves network reasoning on STRING-Yeast, and its structural roles are associated with gene essentiality.}
Results are pooled across Qwen3-8B and Llama-3.1-8B unless otherwise indicated.
\textbf{a} System accuracy for each representation, with 95\% bootstrap CIs. Percentages within selected bars denote the fraction of prompts that exceeded the context window; the dotted line indicates accuracy without network information.
\textbf{b} Pairwise comparisons of BioGlyph with each alternative, restricted to questions for which both representations fitted within the context window and produced an answer. The number of paired answers is shown at right.
\textbf{c} Ablation analysis of the BioGlyph description. Solid bars show system accuracy; hatched bars show accuracy on the common subset of questions readable under every ablation.
\textbf{d} Accuracy across the seven benchmark question families.
\textbf{e} Proportion of proteins annotated by SGD as essential within each structural role. The dashed line indicates the network-wide essentiality rate of 30.9\%.
\textbf{f} Essentiality among equally sized sets of 85 proteins selected as cross-community connectors or ranked using a single centrality measure.}
\label{fig:yeast}
\end{figure}

Across both models, BioGlyph yielded an accuracy of 75.6\% on the yeast benchmark (95\% confidence interval (CI), 72.3--78.9\%), compared with 47.3\% without network information (Fig.~\ref{fig:yeast}a). Direct textual or numerical representations of the retrieved graph performed less well: accuracy was 40.1\% for edge lists, 37.5\% for sentences and 36.1\% for raw graph measurements. GraphSAGE-derived roles yielded 49.2\% accuracy. These representations also frequently exceeded the 24{,}576-token context limit: 56.6\% of raw-measurement prompts, 38.5\% of sentence prompts and 30.6\% of edge-list prompts were over the limit. By contrast, every BioGlyph description fitted within the context window.

To distinguish representational quality from context-window truncation, we performed pairwise analyses restricted to questions for which both representations fitted within the context window and produced an answer (Fig.~\ref{fig:yeast}b). On the subset shared with the raw-measurement table, BioGlyph achieved 79.5\% accuracy, compared with 73.7\% for the table. The corresponding comparisons were 74.6\% versus 57.8\% for sentences, 75.0\% versus 53.1\% for edge lists and 78.1\% versus 47.1\% for GraphSAGE-derived roles. Thus, BioGlyph retained an accuracy advantage when both representations were available in full.

Ablation experiments identified the components of the representation that contributed to this advantage (Fig.~\ref{fig:yeast}c). On the common subset of questions readable under every ablation, role names alone yielded 69.9\% accuracy. Adding the graph measurements supporting each assignment yielded 68.8\%, whereas additionally stating the structural consequence of each role increased accuracy to 82.7\%. The semantic content of the role names also contributed. In an opaque control, each role name was replaced by a meaningless token while the role assignments and all other prompt content were held fixed. This substitution reduced overall system accuracy from 66.8\% to 51.1\%.

The gains extended across multiple forms of structural reasoning (Fig.~\ref{fig:yeast}d). For cut-node questions, BioGlyph achieved 90.2\% accuracy, compared with 48.9\% for sentences and 31.5\% for GraphSAGE. For questions comparing the degrees of two proteins, BioGlyph and GraphSAGE both achieved 95.7\%, whereas sentences achieved 37.0\%. BioGlyph also outperformed the alternative representations across the perturbation-oriented question families.

We next asked whether the structural roles identified by BioGlyph were associated with an independent biological property. STRING-Yeast proteins were matched to SGD gene-essentiality annotations\cite{cherry2012saccharomyces}, which were withheld from both the BioGlyph compiler and the language models. Of all proteins in the network, 30.9\% were essential. This proportion increased to 57.6\% among the 85 proteins classified as cross-community connectors, 46.9\% among community cores and 42.1\% among hubs (Fig.~\ref{fig:yeast}e). By contrast, only 12.9\% of peripheral proteins were essential. The proportions among cut nodes and boundary nodes were close to the network-wide rate.

Finally, we examined whether the cross-community connector role merely recapitulated a standard centrality ranking. We selected equally sized sets of 85 proteins using degree, betweenness or PageRank (Fig.~\ref{fig:yeast}f). Essential proteins constituted 56.5\% of the highest-betweenness set, 48.2\% of the highest-PageRank set and 24.7\% of the highest-degree set, compared with 57.6\% of cross-community connectors. The connector role therefore identifies a group enriched for essential proteins by combining high betweenness with participation across multiple communities, rather than by applying any one of these centrality measures alone. Results for the other benchmark networks, additional model families and further controls are reported in Section~\ref{sec:suppresults}.

\subsection*{BioGlyph remains effective on a dense drug-interaction network}

\begin{figure}[!t]
\centering
\includegraphics[width=\linewidth]{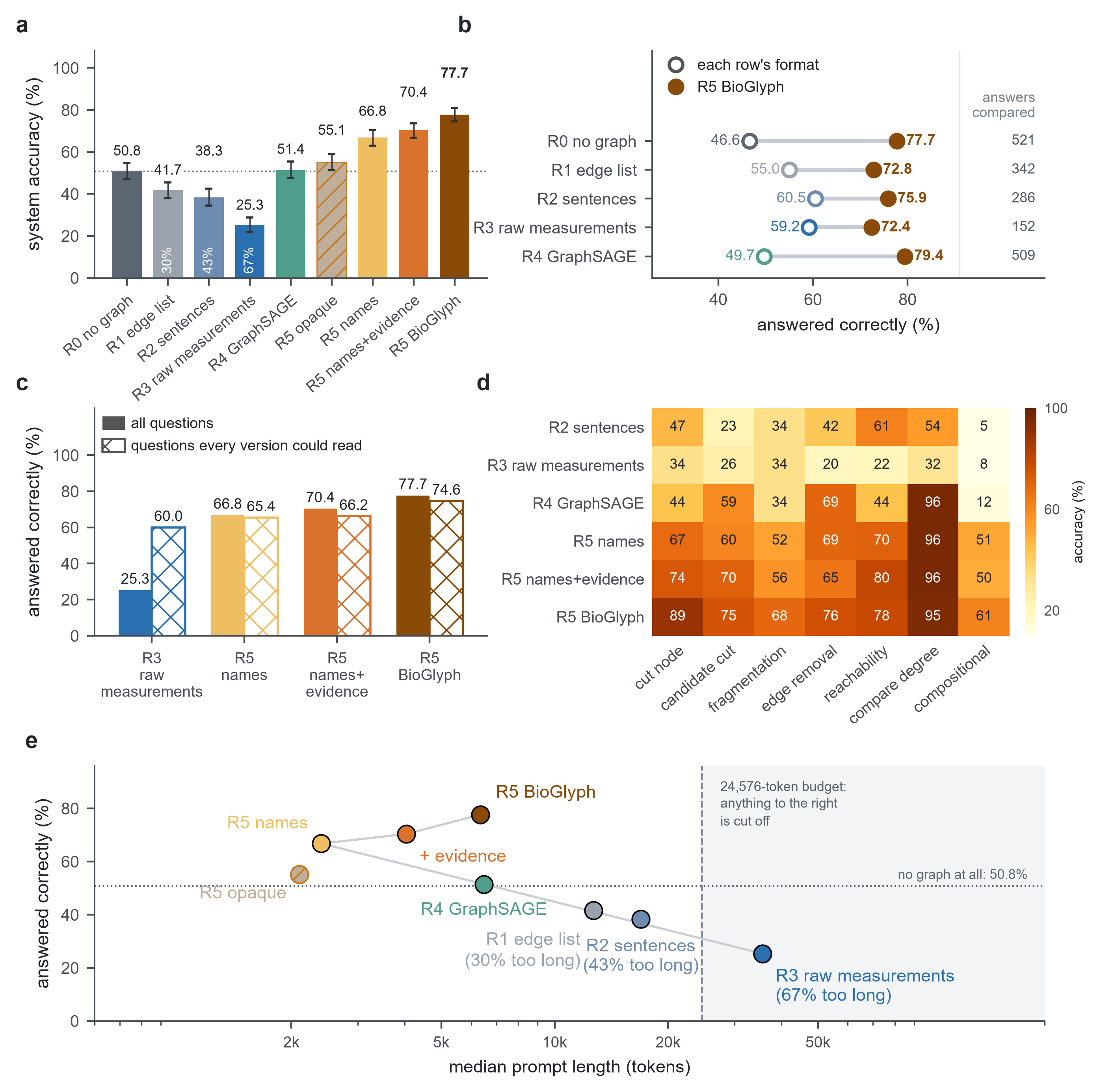}
\caption{\textbf{BioGlyph supports network reasoning on ChCh-Miner, a dense network of 1{,}514 drugs and 48{,}514 reported interactions.}
Results are pooled across Qwen3-8B and Llama-3.1-8B.
\textbf{a} System accuracy for each representation, with 95\% bootstrap CIs. Percentages within selected bars denote the fraction of prompts that exceeded the context window; the dotted line indicates accuracy without network information.
\textbf{b} Pairwise comparisons of BioGlyph with each alternative, restricted to questions for which both representations fitted within the context window and produced an answer. The number of paired answers is shown at right.
\textbf{c} Ablation analysis of the BioGlyph description. Solid bars show system accuracy; hatched bars show accuracy on the common subset of questions readable under every ablation.
\textbf{d} Accuracy across the seven benchmark question families.
\textbf{e} Median prompt length plotted against system accuracy. The vertical dashed line indicates the 24{,}576-token context limit. Connected points show the progression from raw measurements to role names, role names with supporting evidence and the complete BioGlyph description.}
\label{fig:chch}
\end{figure}

We next evaluated BioGlyph on ChCh-Miner, a BioSNAP network in which 1{,}514 drugs are connected by 48{,}514 reported drug--drug interactions\cite{zitnik2018biosnap}. With a mean degree of approximately 64, ChCh-Miner is the densest network in our benchmark. We evaluated 156 questions using the same representations and two frozen 8-billion-parameter language models as in the yeast analysis. Of these questions, 10.3\% referred to a drug absent from the network and therefore required the model to state that they could not be answered. Across both models, BioGlyph achieved 77.7\% system accuracy (95\% confidence interval (CI), 74.5--80.9\%; Fig.~\ref{fig:chch}a), compared with 50.8\% without network information. GraphSAGE-derived roles yielded 51.4\% accuracy. Direct representations of the retrieved graph performed less well: accuracy was 41.7\% for edge lists, 38.3\% for sentences and 25.3\% for raw graph measurements.

The differences in accuracy were accompanied by substantial differences in prompt length. The median raw-measurement prompt contained 35{,}601 tokens, exceeding the 24{,}576-token context limit; consequently, 67.1\% of these prompts could not be passed to the model. The corresponding proportions were 43.3\% for sentence prompts and 30.1\% for edge-list prompts. By contrast, the median BioGlyph description contained 6{,}353 tokens, and every BioGlyph description fitted within the context window (Fig.~\ref{fig:chch}e). To distinguish representational quality from context-window truncation, we again restricted pairwise analyses to questions for which both representations fitted within the context window and produced an answer (Fig.~\ref{fig:chch}b). On the subset shared with the raw-measurement table, BioGlyph achieved 72.4\% accuracy, compared with 59.2\% for the table. The corresponding comparisons were 75.9\% versus 60.5\% for sentences, 72.8\% versus 55.0\% for edge lists and 79.4\% versus 49.7\% for GraphSAGE-derived roles. BioGlyph therefore retained an accuracy advantage when both representations were available in full.

Ablation experiments further identified the components underlying this advantage (Fig.~\ref{fig:chch}c). On the common subset of questions readable under every ablation, role names alone yielded 65.4\% accuracy, and adding the measurements supporting each assignment yielded 66.2\%. Additionally stating the structural consequence of each role increased accuracy to 74.6\%. The semantic content of the role names also contributed: replacing each name with a meaningless token while preserving the role assignments and all other prompt content reduced overall system accuracy from 66.8\% to 55.1\%.

The improvement extended across multiple forms of structural reasoning (Fig.~\ref{fig:chch}d). For cut-node questions, BioGlyph achieved 89.1\% accuracy, compared with 46.7\% for sentences and 43.5\% for GraphSAGE. For questions comparing the degrees of two drugs, BioGlyph achieved 94.6\% and GraphSAGE 95.7\%, whereas sentences achieved 54.3\%. BioGlyph also outperformed the alternative representations across the principal perturbation and connectivity question families. Two additional model families showed the same overall pattern on ChCh-Miner (Section~\ref{sec:suppresults}). A comparison of ChCh-Miner with the other principal benchmark networks is provided in Fig.~\ref{fig:S1}.

\subsection*{BioGlyph identifies structurally important proteins in a human pathway network}

\begin{figure}[!t]
\centering
\includegraphics[width=\linewidth]{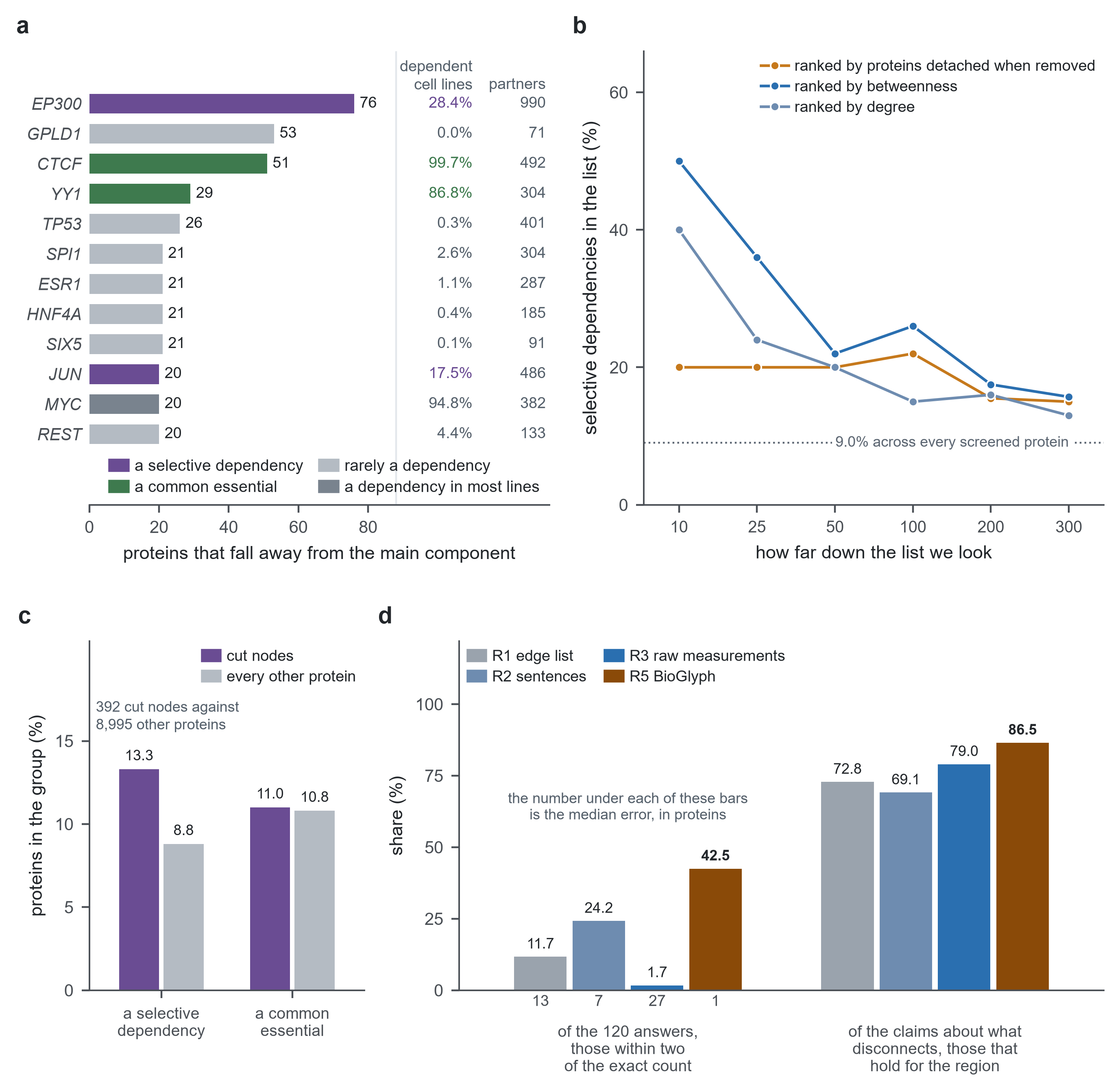}
\caption{\textbf{BioGlyph identifies structurally important proteins in the Reactome functional interaction network.}
Panels \textbf{a--c} combine compiler outputs with DepMap annotations and do not involve a language model. Panel \textbf{d} pools results from Qwen3-8B and Llama-3.1-8B.
\textbf{a} The 12 cut nodes whose removal detaches the largest number of proteins from the main component. Bar colours indicate DepMap dependency categories; columns at right report the fraction of cell lines dependent on each protein and its number of network partners.
\textbf{b} Proportion of selective dependencies among the highest-ranked proteins according to the number detached after removal, betweenness or degree. The dotted line indicates the proportion among all screened proteins.
\textbf{c} Proportions of selective dependencies and common essentials among screened cut nodes and all other screened proteins.
\textbf{d} Language-model reasoning across 60 Reactome modules. Left, the proportion of responses within two proteins of the exact region-specific detachment count; median absolute errors are shown beneath the bars. Right, the proportion of checkable structural claims that agree with exact region-level graph analysis.}
\label{fig:reactome}
\end{figure}

We next applied BioGlyph to the Reactome functional interaction network\cite{wu2010human}, which comprises 10{,}022 proteins and 194{,}494 functional interactions, with a mean degree of approximately 39. Its largest connected component contains 9{,}823 proteins, including 402 cut nodes whose removal disconnects at least one protein from the remaining main component. To quantify the structural consequence of each cut node, we removed each one in turn and counted the number of proteins separated from the largest remaining component (Fig.~\ref{fig:reactome}a). Removal of EP300 produced the largest separation, detaching 76 proteins, followed by GPLD1 with 53 and CTCF with 51. This effect was not determined by the number of interaction partners alone: GPLD1 has 71 partners and detaches 53 proteins, whereas MYC has 382 partners but detaches only 20.

We compared these structural predictions with gene-dependency measurements from DepMap 24Q4\cite{tsherniak2017defining,dempster2021chronos}. DepMap provided annotations for 9{,}387 proteins in the largest connected component across 1{,}178 cancer cell lines. Of these proteins, 392 were cut nodes and 8{,}995 were not. Selective dependencies constituted 13.3\% of the cut nodes, compared with 8.8\% of the other proteins (Fig.~\ref{fig:reactome}c). The proportions classified as common essentials were similar, at 11.0\% and 10.8\%, respectively. At the protein level, EP300 was a dependency in 28.4\% of cell lines; CTCF and YY1 were common essentials, whereas GPLD1 was rarely a dependency. Neither the BioGlyph compiler nor the structural screen used these dependency annotations.

We next compared the detachment-based ranking with rankings based on betweenness and degree (Fig.~\ref{fig:reactome}b). Among the ten highest-ranked proteins, selective dependencies constituted 20.0\% of the detachment-ranked set, 50.0\% of the betweenness-ranked set and 40.0\% of the degree-ranked set. Among the top 300 proteins, the corresponding proportions were 15.0\%, 15.7\% and 13.0\%. Thus, the three criteria prioritized partially distinct sets of proteins: detachment quantifies the size of the network region disrupted by removal, whereas betweenness and degree capture different aspects of structural prominence. We then asked whether language models could use BioGlyph descriptions to reason about protein-removal effects. We compiled 60 Reactome modules and asked each model how many proteins would become detached after removal of a named protein (Fig.~\ref{fig:reactome}d). A response was considered correct when its estimate was within two proteins of the exact count for that region. Neither the BioGlyph descriptions nor the raw-measurement tables exceeded the context limit in this analysis, allowing every prompt to reach the models.

Across 120 responses, BioGlyph yielded 42.5\% accuracy and a median absolute error of one protein. By comparison, sentences yielded 24.2\% accuracy with a median error of seven proteins, edge lists yielded 11.7\% with a median error of 13 and raw graph measurements yielded 1.7\% with a median error of 27. The models supplied a numerical count in 74.2\% of BioGlyph responses, compared with 32.5\% of responses based on raw measurements. When analysis was restricted to responses containing a count, accuracy was 57.3\% with BioGlyph and 5.1\% with raw measurements. We also compared the models' qualitative structural claims with exact graph analysis. Among checkable claims, 86.5\% of those generated from BioGlyph descriptions were correct, compared with 79.0\% from raw measurements, 72.8\% from edge lists and 69.1\% from sentences. The models recovered several exact perturbation effects, including the detachment of 54 proteins after removal of RAD21 and the exact counts for E2F1 and GATA1. Further protein-level examples and controls are reported in Fig.~\ref{fig:S3f} and Section~\ref{sec:suppresults}.


\subsection*{BioGlyph improves structural reasoning in a dense social network}

\begin{figure}[!t]
\centering
\includegraphics[width=\linewidth]{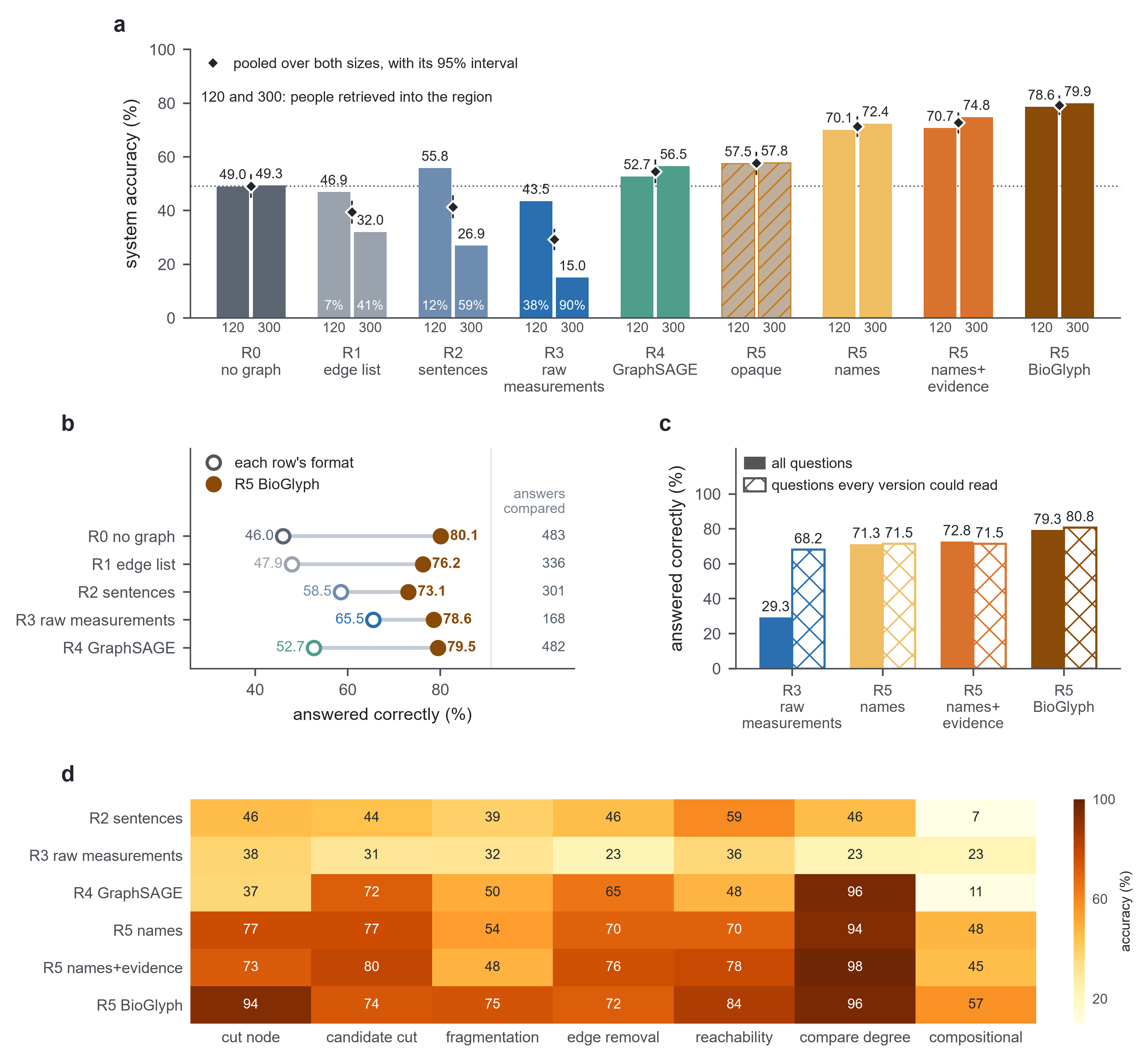}
\caption{\textbf{BioGlyph improves structural reasoning on ego-Facebook, a network of 4{,}039 individuals and 88{,}234 friendship links.}
Results are pooled across Qwen3-8B and Llama-3.1-8B.
\textbf{a} System accuracy at two retrieval sizes, with regions containing at most 120 or 300 individuals. Bars show results for each retrieval size; diamonds show pooled accuracy with 95\% bootstrap CIs. Percentages within selected bars denote the fraction of prompts that exceeded the context window. The dotted line indicates accuracy without network information.
\textbf{b} Pairwise comparisons of BioGlyph with each alternative, restricted to questions for which both representations fitted within the context window and produced an answer. The number of paired answers is shown at right.
\textbf{c} Ablation analysis of the BioGlyph description. Solid bars show system accuracy; hatched bars show accuracy on the common subset of questions readable under every ablation.
\textbf{d} Accuracy across the seven benchmark question families.}
\label{fig:facebook}
\end{figure}

We next evaluated BioGlyph on ego-Facebook, a SNAP network comprising 4{,}039 individuals and 88{,}234 friendship links\cite{leskovec2014snap}. The network has a mean degree of approximately 44; all nodes belong to a single connected component, and only 11 are cut nodes. We evaluated 147 questions using the same representations and two frozen 8-billion-parameter language models as in the preceding experiments. Of these questions, 10.2\% referred to an individual absent from the network and therefore required the model to state that they could not be answered. Across both models, BioGlyph achieved 79.3\% system accuracy (95\% confidence interval (CI), 75.9--82.5\%; Fig.~\ref{fig:facebook}a), compared with 49.1\% without network information. GraphSAGE-derived roles yielded 54.6\% accuracy. Direct representations of the retrieved graph performed less well: accuracy was 39.5\% for edge lists, 41.3\% for sentences and 29.3\% for raw graph measurements.

Prompt lengths again differed substantially among representations. The median raw-measurement prompt contained 35{,}161 tokens, exceeding the 24{,}576-token context limit; consequently, 63.9\% of these prompts could not be passed to the model. The corresponding proportions were 35.5\% for sentence prompts and 24.0\% for edge-list prompts. By contrast, the median BioGlyph and GraphSAGE descriptions contained 6{,}012 and 6{,}547 tokens, respectively, and both representations remained within the context window. To distinguish representational quality from context-window truncation, we restricted pairwise analyses to questions for which both representations fitted within the context window and produced an answer (Fig.~\ref{fig:facebook}b). On the subset shared with the raw-measurement table, BioGlyph achieved 78.6\% accuracy, compared with 65.5\% for the table. The corresponding comparisons were 73.1\% versus 58.5\% for sentences, 76.2\% versus 47.9\% for edge lists and 79.5\% versus 52.7\% for GraphSAGE-derived roles. BioGlyph therefore retained an accuracy advantage when both representations were available in full.

Ablation experiments identified the components contributing to this advantage (Fig.~\ref{fig:facebook}c). On the common subset of questions readable under every ablation, role names alone yielded 71.5\% accuracy, and adding the measurements supporting each assignment yielded the same accuracy. Additionally stating the structural consequence of each role increased accuracy to 80.8\%. The semantic content of the role names also contributed: replacing each name with a meaningless token while preserving the role assignments and all other prompt content reduced overall system accuracy from 71.3\% to 57.7\%. The largest gains occurred on questions concerning network connectivity (Fig.~\ref{fig:facebook}d). For cut-node questions, BioGlyph achieved 93.5\% accuracy, compared with 45.7\% for sentences and 37.0\% for GraphSAGE. For questions comparing the degrees of two individuals, BioGlyph and GraphSAGE both achieved 95.7\%. BioGlyph further achieved 74\% accuracy on candidate-cut questions, 75\% on fragmentation questions, 72\% on edge-removal questions and 84\% on reachability questions.

Performance at the two retrieval sizes showed how the representations responded as the retrieved region expanded (Fig.~\ref{fig:facebook}a). When the maximum region size increased from 120 to 300 individuals, sentence accuracy decreased from 55.8\% to 26.9\%, while the proportion of sentence prompts exceeding the context window increased from 12.2\% to 58.8\%. Raw-measurement accuracy similarly decreased from 43.5\% to 15.0\%. By contrast, GraphSAGE accuracy was 52.7\% and 56.5\% at the two retrieval sizes, and BioGlyph accuracy remained stable at 78.6\% and 79.9\%, respectively. Results for the remaining benchmark networks and additional controls are reported in Section~\ref{sec:suppresults}.

\subsection*{BioGlyph supports multi-turn reasoning on an email network}

\begin{figure}[!t]
\centering
\includegraphics[width=\linewidth]{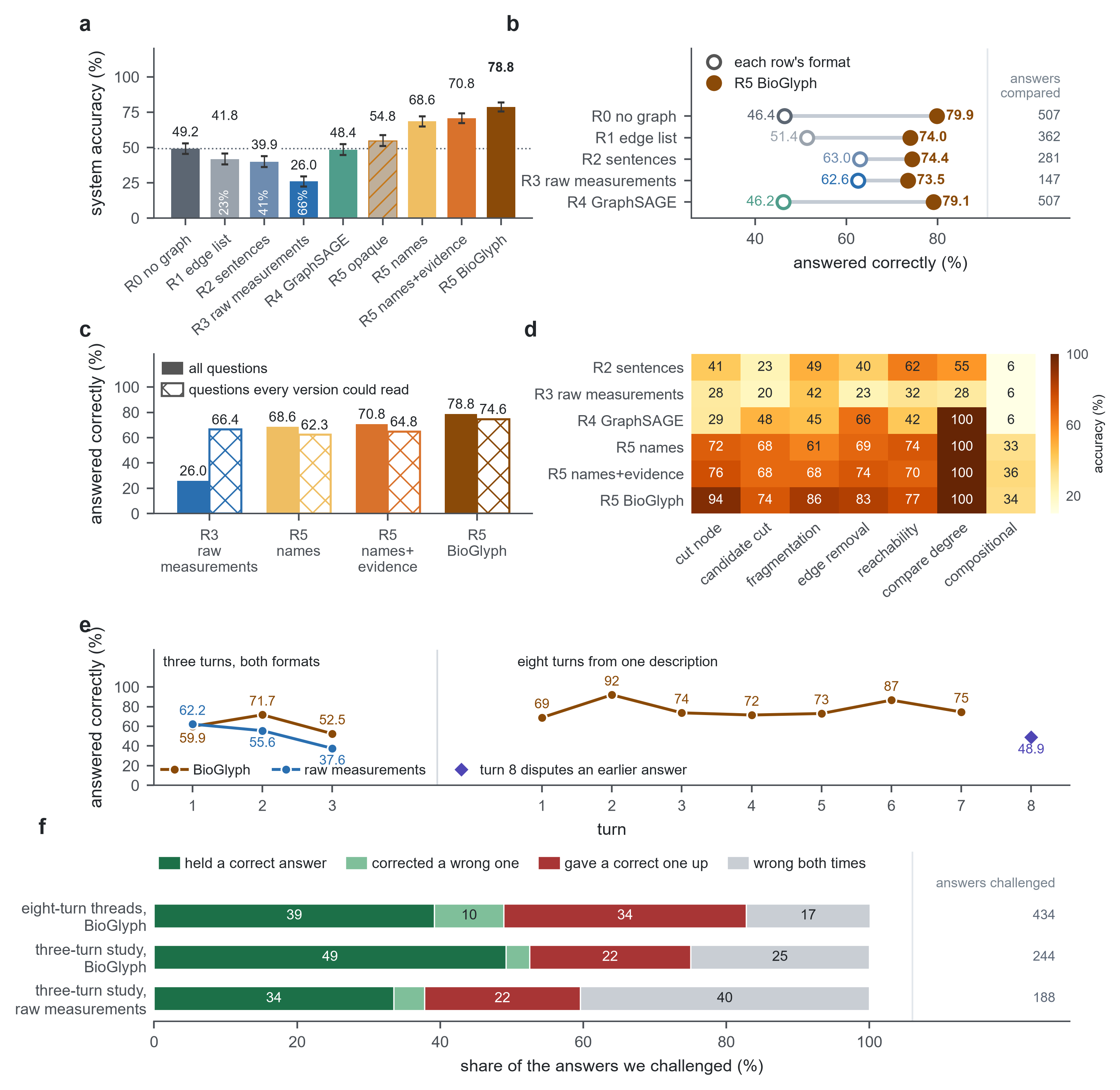}
\caption{\textbf{BioGlyph supports structural and multi-turn reasoning on email-Eu-core.}
Results are pooled across Qwen3-8B and Llama-3.1-8B.
\textbf{a} System accuracy for each representation, with 95\% bootstrap CIs. Percentages within selected bars denote the fraction of prompts that exceeded the context window; the dotted line indicates accuracy without network information.
\textbf{b} Pairwise comparisons of BioGlyph with each alternative, restricted to questions for which both representations fitted within the context window and produced an answer. The number of paired answers is shown at right.
\textbf{c} Ablation analysis of the BioGlyph description. Solid bars show system accuracy; hatched bars show accuracy on the common subset of questions readable under every ablation.
\textbf{d} Accuracy across the seven benchmark question families.
\textbf{e} Turn-level accuracy in multi-turn conversations. The left panel shows a three-turn study using either BioGlyph or raw graph measurements. The right panel shows eight-turn conversations supported by a single BioGlyph description provided at the first turn; turn 8 challenges an earlier answer.
\textbf{f} Outcomes after the model was told that its preceding answer was incorrect, distinguishing cases in which the model retained a correct answer, corrected an incorrect answer, abandoned a correct answer or remained incorrect.}
\label{fig:email}
\end{figure}

We finally evaluated BioGlyph on email-Eu-core, a communication network from a European research institution\cite{leskovec2014snap}. The network comprises 986 individuals and 16{,}064 email links, with a mean degree of approximately 33. All nodes belong to a single connected component; 73 nodes are cut nodes and 95 links are bridges. We evaluated 156 questions using the same six representations and two frozen 8-billion-parameter language models as in the preceding experiments. Of these questions, 10.3\% referred to an individual absent from the network and therefore required the model to state that they could not be answered. Across both models, BioGlyph achieved 78.8\% system accuracy (95\% confidence interval (CI), 75.6--81.9\%; Fig.~\ref{fig:email}a), compared with 49.2\% without network information and 48.4\% with GraphSAGE-derived roles. Direct representations of the retrieved graph performed less well: accuracy was 41.8\% for edge lists, 39.9\% for sentences and 26.0\% for raw graph measurements.

Prompt lengths again differed substantially among representations. The raw-measurement table exceeded the 24{,}576-token context limit in 66.3\% of cases, compared with 41.3\% for sentence prompts and 22.8\% for edge-list prompts. By contrast, the median BioGlyph description contained 5{,}825 tokens, and every BioGlyph prompt fitted within the context window. To distinguish representational quality from context-window truncation, we restricted pairwise analyses to questions for which both representations fitted within the context window and produced an answer (Fig.~\ref{fig:email}b). Among the 147 paired answers shared with the raw-measurement table, BioGlyph achieved 73.5\% accuracy, compared with 62.6\% for the table. The corresponding comparisons were 74.4\% versus 63.0\% for sentences, 74.0\% versus 51.4\% for edge lists and 79.1\% versus 46.2\% for GraphSAGE-derived roles. BioGlyph therefore retained an accuracy advantage when both representations were available in full.

Ablation experiments identified the components contributing to this advantage (Fig.~\ref{fig:email}c). On the common subset of questions readable under every ablation, role names alone yielded 62.3\% accuracy. Adding the measurements supporting each assignment increased accuracy to 64.8\%, whereas additionally stating the structural consequence of each role increased it to 74.6\%. The semantic content of the role names also contributed: replacing each name with a meaningless token while preserving the role assignments and all other prompt content reduced overall system accuracy from 68.6\% to 54.8\%. The largest gains occurred on questions concerning network connectivity (Fig.~\ref{fig:email}d). For cut-node questions, BioGlyph achieved 93.5\% accuracy, compared with 41.3\% for sentences and 29.3\% for GraphSAGE. BioGlyph further achieved 85.9\% accuracy on fragmentation questions and 83.0\% on edge-removal questions. Degree comparisons were comparatively simple: BioGlyph, GraphSAGE and both shorter BioGlyph variants achieved 100\% accuracy on this question family.

We next asked whether a single BioGlyph description could support reasoning across successive conversation turns (Fig.~\ref{fig:email}e,f). In a three-turn study, the network region was provided with the initial question, followed by a related question and then a challenge to the preceding answer. BioGlyph yielded accuracies of 59.9\%, 71.7\% and 52.5\% across the three turns, respectively. Raw graph measurements yielded 62.2\%, 55.6\% and 37.6\%. Thus, although the two representations performed similarly on the initial turn, BioGlyph better supported the follow-up question and the subsequent challenge.

We then conducted 219 eight-turn conversations for each model, providing a single BioGlyph description only at the first turn. Accuracy across all graded turns was 75.2\%. When an earlier answer was challenged, the models retained a correct answer in 39.2\% of cases and corrected an incorrect answer in 9.7\% (Fig.~\ref{fig:email}f). In the three-turn study, BioGlyph preserved 49.2\% of initially correct answers after a challenge, compared with 33.5\% for raw measurements. Performance was also insensitive to question phrasing: rewriting the benchmark questions in ordinary prose changed BioGlyph accuracy from 68.6\% to 67.9\%, while raw-measurement accuracy changed from 27.5\% to 29.3\%. Complete conversation examples are provided in Section~\ref{sec:talking}.

%% file: sections/discussion.tex
Networks remain challenging objects for language models. Although these models process natural language fluently, they often fail on questions that require structural reasoning, such as predicting the consequences of removing a node or edge, identifying elements that connect otherwise distinct regions, or determining which elements are important for network integrity. Here we show that these limitations can be reduced by changing the representation supplied to the model. Rather than presenting the network directly or requiring the model to interpret a table of graph measurements, BioGlyph compiles topology into descriptions that identify structural roles, provide the evidence supporting each assignment and state its structural consequence. These descriptions enabled small, frozen language models to reason across 20 biological, social, information and infrastructure networks. The largest performance improvement obtained by BioGlyph over existing methods occurred in dense or community-structured networks, for which direct representations were often both difficult to interpret and too long for the context window.

The performance improvement of BioGlyph were due to two unique features of the compiled representation (Section~\ref{sec:s_ladder}). The first was selective compression. By determining which structural information was relevant to report, the compiler produced substantially shorter descriptions than edge lists, adjacency sentences or raw-measurement tables. The compression allowed nearly all BioGlyph prompts to reach the model. Direct graph representations, by contrast, frequently exceeded the context limit, thus language models failed to produce coherent response. The second feature was semantic interpretation. Instead of adding the measurements supporting a role assignment, producing little improvement over role names alone, we explicitly stated the structural consequence of the role consistently which resulted in substantial accuracy improvement. Replacing meaningful role names with arbitrary tokens also reduced performance, indicating that both the role vocabulary and the accompanying consequence statements help the model use the encoded structure.

This representation also supported forms of interaction that are difficult to obtain from a table. A single BioGlyph description could be supplied once and used throughout an eight-turn conversation, including follow-up questions and challenges to earlier answers (Section~\ref{sec:s_conv}). Performance changed little when benchmark questions were rewritten in ordinary prose, suggesting that the descriptions were not narrowly coupled to the templates used to generate the benchmark. Raw measurements supported these interactions less effectively and, when the initial prompt exceeded the context window, could not support a conversation at all. The models nevertheless sometimes abandoned correct answers when challenged, showing that access to an informative representation does not eliminate the conversational instabilities of the underlying language model.

Comparisons with learned graph representations further distinguish BioGlyph from approaches that encode topology without assigning explicit structural meaning. Unsupervised roles produced by four graph encoders performed close to the no-network baseline, and incorporating these roles into the compiled description did not improve its accuracy. Supplying continuous node embeddings as soft prompts\cite{perozzi2024let} also performed no better than supplying embeddings from mismatched nodes (Section~\ref{sec:s_learned}). These findings complement previous work that represents graphs as edge lists, sentences or templates\cite{fatemi2024talk,wang2023can,guo2023gpt4graph}, or trains graph-aware adapters\cite{chen2024llaga,tang2024graphgpt}. In our experiments, direct textual renderings were difficult for small frozen models to use, whereas a semantically explicit intermediate representation made the same topology more accessible without modifying either the graph or the language model.

The structural roles in BioGlyph also had biological relevance independently of language-model performance. In the yeast interactome, cross-community connectors were essential more frequently than the network average and at a rate comparable to, or slightly higher than, equally sized groups selected using individual centrality measures. Community cores and hubs were also enriched in essential proteins, whereas peripheral proteins were depleted. These findings extend longstanding observations that network position is associated with gene essentiality\cite{jeong2001lethality,he2006hubs,yu2007importance} by expressing combinations of structural properties as named, inspectable roles. Importantly, not every role was associated with essentiality, indicating that the vocabulary distinguishes structural properties rather than imposing a uniformly positive biological interpretation.

In the Reactome functional interaction network, the same compiler provided an annotation-independent perturbation screen. Cut nodes contained a higher proportion of selective cancer dependencies than other screened proteins, although individual proteins varied substantially: some structurally disruptive proteins were broadly or selectively essential, whereas others showed little dependency. Two additional human interactomes similarly showed associations between selected structural roles and gene essentiality or broad expression, but these associations were not uniform across roles or networks (Section~\ref{sec:s_morebio}). These results suggest that compiled structural roles can complement biological annotations by identifying proteins with distinct forms of network importance, rather than serving as direct proxies for biological essentiality. Because each assignment is linked to explicit measurements and exact graph operations, both the role and any subsequent language-model claim can be audited against the underlying network analysis.

A compiled description is not a substitute for a graph algorithm. When a question can be expressed as a single graph operation, the appropriate algorithm can answer it directly; indeed, exact algorithms supplied the reference answers throughout our benchmark. The value of the compiled representation lies instead at the interface between algorithmic analysis and natural-language interaction. A user may pose a question without specifying the required graph operation, request an explanation that integrates several structural properties or challenge a previous answer. BioGlyph acts as a front end to classical graph analysis: it executes predefined algorithms, records their outputs and translates them into a closed-world description that a language model can interpret and explain. This separation preserves exact computation while enabling a more flexible mode of interaction.

BioGlyph has several limitations. First, its performance depends on network density. In sparse networks, retrieved subgraphs may be small enough to fit directly within the model’s context window; in such cases, adjacency sentences or raw graph measurements can perform comparably to BioGlyph. Conversely, BioGlyph descriptions occasionally exceeded the prompt budget, reducing accuracy on two of the 20 networks. Increasing the context window recovered performance on one of these networks (Section~\ref{sec:s_r3}). BioGlyph therefore reduces, but does not always eliminate, the trade-off between preserving sufficient structural information and limiting prompt length. Second, language models remain susceptible to misleading feedback. Although BioGlyph increased the likelihood that models retained a correct answer when challenged, the models still failed to reach the correct answer in some conversations. Structural claims should therefore be verified using the underlying graph algorithms before they are revised or accepted. Third, the advantage of BioGlyph varies based on the used model: on four hosted frontier models with sufficient context, the compiled description led the raw-measurement table in one of twelve comparisons, matched it in four and trailed it in seven (Section~\ref{sec:s_families}), and Qwen3-32B closes the controlled gap to parity (Section~\ref{sec:s_r3}). BioGlyph therefore substitutes for a selection that larger models increasingly perform unaided, and is most useful for the open models that can be run locally.

Our benchmarking approach also has limitations. First, the scored benchmark was restricted to questions with answers that could be determined exactly. We additionally examined more open-ended biological questions, but these often require the integration of multiple structural criteria with experimental or clinical evidence and could not be assessed using the same grading procedure. Second, retrieval for one family of benchmark questions used the predicate that was subsequently evaluated during grading. An oracle-free retrieval analysis preserved the ordering of the representations, indicating that this information was unlikely to account for the main comparison (Section~\ref{sec:s_retrieval}). Nevertheless, future benchmarks should fully separate retrieval from evaluation. Finally, all language models were kept frozen. This design isolated the effect of network representation and avoided task-specific adaptation, but it may also have limited the performance attainable from BioGlyph descriptions. In a preregistered screen evaluated against biological outcomes rather than graph-derived answers, a frozen BioGlyph reader provided competitive results to the best deterministic decision rule (Section~\ref{sec:s_cko}).

These limitations motivate three extensions of our study. First, training models on compiled descriptions could determine how much of the remaining error arises from the representation and how much from the frozen reader. Second, using the compiler to generate structurally justified candidates for subsequent evaluation by a language model could support biological investigation without asking the model to replace deterministic analysis. Third, an agent could invoke exact graph algorithms whenever a question reduces to a defined operation and use the compiled description for interpretation, explanation and multi-step reasoning. Such a system would combine the reliability of graph algorithms with the flexibility of language-based interaction. We release the compiler, benchmark questions and exact answers, stored model responses and scored results to enable independent verification and further development.

\section*{Conclusion}

BioGlyph converts network topology into concise, interpretable descriptions that small, frozen language models can use without modification to either the model or the underlying network. Across networks from multiple domains, these descriptions improved structural reasoning relative to edge lists, adjacency sentences, raw graph measurements and unsupervised graph representations, particularly for dense and community-structured networks. Ablation experiments showed that the improvement reflects both selective compression and the explicit communication of structural consequences. The same descriptions also supported multi-turn reasoning and produced claims that could be verified against exact graph analysis. In biological networks, BioGlyph identified structural roles associated with independent measures of gene essentiality and dependency while remaining blind to those annotations. The framework is therefore best viewed not as a replacement for graph algorithms, but as an interpretable interface between exact network analysis and natural-language reasoning. By making structural evidence accessible, concise and auditable, BioGlyph provides a practical foundation for language-model-assisted investigation of complex biomedical networks.

%% file: sections/methods.tex

BioGlyph converts a network region into a short description of its structural roles. For each benchmark question, we retrieve the region around the named nodes, compute classical graph quantities on that region, and apply one fixed rule for each of eleven roles. Each assigned role becomes a line of readable text that is given to a frozen language model. Exact graph algorithms provide the ground-truth answers for all graded benchmark questions; no language model grades another language model. We make stochastic steps reproducible with fixed seeds and deterministic tie-breaking.

\subsection*{Structural signals}

Let \(G=(V,E)\) denote a network with \(n=|V|\) nodes, and let \(\widetilde{G}\) denote its simple undirected view, obtained by removing self-loops and ignoring edge direction. We compute connected components, cut nodes, bridges, the \(k\)-core number \(c(v)\), and the community partition on \(\widetilde{G}\). Degree \(d(v)\), PageRank \(\pi(v)\), node betweenness \(b(v)\), and edge betweenness \(b(e)\) are computed on \(G\), preserving direction when the input graph is directed. The networks used in the experiments reported here are treated as undirected, so \(G=\widetilde{G}\).

We write \(N(v)\) for the neighbors of node \(v\). Graph quantities are computed with NetworkX\cite{hagberg2008exploring}. We use Brandes' algorithm\cite{brandes2001faster} for node and edge betweenness and the standard core decomposition\cite{seidman1983network} for \(k\)-core number. Betweenness is computed exactly when \(n\leq 5{,}000\). For larger graphs, we estimate node and edge betweenness from \(1{,}000\) pivot nodes selected with seed \(0\). The signal record stores whether exact or sampled betweenness was used.

We partition \(\widetilde{G}\) into communities \(\mathcal{M}\) using the Leiden algorithm\cite{traag2019louvain} with the modularity objective\cite{newman2006modularity} and seed \(0\). We relabel communities by decreasing size and break ties using the smallest node identifier. This ordering keeps community identifiers stable across repeated runs.
Let \(C(v)\) denote the community containing node \(v\), and let
\begin{equation}
k_{v,m}
=
\left|
\left\{
u\in N(v):C(u)=m
\right\}
\right|
\label{eq:community_degree}
\end{equation}
be the number of neighbors of \(v\) that belong to community \(m\). We compute the participation coefficient of Guimer\`a and Amaral\cite{guimera2005functional} as
\begin{equation}
P(v)
=
1-
\sum_{m\in\mathcal{M}}
\left(
\frac{k_{v,m}}{d(v)}
\right)^2 .
\label{eq:participation}
\end{equation}

We set \(P(v)=0\) when \(d(v)=0\). A value near zero indicates that most neighbors belong to one community; larger values indicate neighbors spread across several.
We also compute the intra-community degree
\begin{equation}
k^{\mathrm{int}}(v)=k_{v,C(v)},
\label{eq:intracommunity}
\end{equation}
which counts the neighbors of \(v\) that remain inside its own community.

\subsection*{The BioGlyph compiler}

The compiler assigns eleven structural roles at three levels. Eight roles describe nodes, two describe edges, and one describes a community. Each emitted role contains its name, the measurements that triggered the assignment, and a short statement describing the corresponding structural consequence. Two roles are assigned directly from exact connectivity decompositions. \glyph{CUT\_NODE} marks an articulation point of \(\widetilde{G}\), and \glyph{BRIDGE\_EDGE} marks a bridge. Both assignments are exact and use no numerical threshold.

Six roles identify unusually large structural values. For a structural quantity \(X\), let \(\mu_X\) and \(s_X\) denote its mean and standard deviation over the relevant comparison set. An element \(u\) receives the role when
\begin{equation}
X(u)>\mu_X+\sigma_X s_X,
\label{eq:role_threshold}
\end{equation}
where \(\sigma_X\) is fixed for that role. We do not assign a threshold-based role when \(s_X=0\). Unless stated otherwise, the comparison distribution is computed within the retrieved region rather than over the full network.

The six threshold-based roles are:
\begin{itemize}
\setlength{\itemsep}{1pt}

\item \glyph{HUB}: degree \(d(v)\), with \(\sigma=2.0\), compared across all nodes in the region.

\item \glyph{AUTHORITY}: PageRank \(\pi(v)\), with \(\sigma=2.0\). We define this role for directed networks; it is not used in the undirected networks reported here.

\item \glyph{BOTTLENECK\_LINK}: edge betweenness \(b(e)\), with \(\sigma=2.0\), compared across all edges in the region.

\item \glyph{CROSS\_COMMUNITY\_CONNECTOR}: node betweenness \(b(v)\), with \(\sigma=2.0\), together with the requirement that \(v\) has neighbors in at least two communities.

\item \glyph{BOUNDARY\_NODE}: participation coefficient \(P(v)\), with \(\sigma=1.5\).

\item \glyph{COMMUNITY\_CORE}: intra-community degree \(k^{\mathrm{int}}(v)\), with \(\sigma=1.0\). The comparison is performed separately within each community containing at least three nodes.

\end{itemize}

Three additional roles use fixed structural criteria:
\begin{itemize}
\setlength{\itemsep}{1pt}

\item \glyph{ISOLATE}: a node whose connected component contains one node.

\item \glyph{PERIPHERAL}: a node with \(c(v)\leq 1\) that belongs to a component containing more than one node.

\item \glyph{FRAGILE\_REGION}: a community whose induced subgraph has edge connectivity \(\lambda\leq 1\). We evaluate edge connectivity for communities containing between \(3\) and \(1{,}500\) nodes. Communities outside this range receive no edge-connectivity value.

\end{itemize}

For a \glyph{CUT\_NODE}, the evidence field records the number of connected components in the whole region before and after removal, which is not the component-local count that Eq.~\eqref{eq:node_pieces} grades. For small regions we recompute that count after removal exactly; for larger graphs the field can hold a lower bound instead. The distinction affects the reported evidence, not the role assignment, which is exact at every network size.
Algorithm~\ref{alg:compile} summarizes the compiler.

\begin{algorithm}[!t]
\caption{BioGlyph compilation of one network region}
\label{alg:compile}
\begin{algorithmic}[1]
\STATE \textbf{Input:} region \(G\), seed \(s\), fixed role parameters \(\{\sigma_X\}\)
\STATE Construct the simple undirected view \(\widetilde{G}\)
\STATE Compute structural signals on \(G\) and \(\widetilde{G}\)
\STATE Partition \(\widetilde{G}\) into communities using Leiden with seed \(s\)
\STATE Compute \(P(v)\) and \(k^{\mathrm{int}}(v)\)
\STATE Initialize the role set \(\mathcal{G}\leftarrow\emptyset\)
\STATE Assign \glyph{CUT\_NODE} to every articulation point
\STATE Assign \glyph{BRIDGE\_EDGE} to every bridge
\STATE Assign \glyph{ISOLATE} and \glyph{PERIPHERAL} using their fixed structural criteria
\FOR{each threshold-based role}
    \STATE Assign the role to elements satisfying Eq.~\eqref{eq:role_threshold}
\ENDFOR
\FOR{each community containing \(3\) to \(1{,}500\) nodes}
    \STATE Assign \glyph{FRAGILE\_REGION} if its edge connectivity is at most \(1\)
\ENDFOR
\STATE Attach the triggering measurements and structural consequence to every assigned role
\STATE \textbf{Return:} \(\mathcal{G}\), sorted by level, role name, and target
\end{algorithmic}
\end{algorithm}

We render the compiler output at several levels of detail. \emph{Names} lists each target and its assigned structural roles. \emph{Names with evidence} adds the measurements that triggered each role. The \emph{full BioGlyph description} adds the short structural consequence associated with each role. It also states a closed-world guarantee: every cut node and every bridge in the region is listed, so a node without a \glyph{CUT\_NODE} entry is not an articulation point of that region.

The \emph{opaque} control keeps the same assigned roles as the \emph{names} rendering but replaces each semantic role name with a meaningless token. The control carries no evidence and no structural consequence, which makes it the \emph{names} rendering with the vocabulary withheld. All BioGlyph renderings also declare the nodes present in the retrieved region, including nodes that receive no structural role.

\subsection*{Retrieval and prompt construction}

Each benchmark question names one or more nodes of a full network. We retrieve a region around those targets and render it under each representation. Named targets are admitted first and remain in the region even above the node cap. If a target identifier does not exist in the full network, the pipeline marks the question as unanswerable before language-model inference.
We use two retrieval settings. The first expands two breadth-first-search rings from the named targets and caps the retrieved region at \(120\) nodes. Within each expansion step, we admit nodes first by increasing graph distance and then by node identifier.
The second setting allows up to \(300\) nodes and routes on the question family. Six families again begin with two breadth-first rings. Counterfactual reachability questions use paths instead of rings: retrieval admits the named targets, the shortest paths between them, the shortest path that survives the removal, and the immediate target neighborhoods, each in deterministic order.

After initial retrieval, a family-specific sufficiency check can trigger at most two additional expansion rings while capacity remains. For question families whose answer depends on the complete local neighborhood of a target, the check asks whether that neighborhood is present. For counterfactual reachability, the check asks whether the two endpoints remain connected inside the retrieved region once the named node is removed. The check therefore evaluates the graded predicate itself, so the stopping rule for this one family uses privileged information. Section~\ref{sec:s_retrieval} reports an oracle-free retrieval arm that routes by personalized PageRank and never runs the loop. The oracle described below, rather than the retrieval heuristic, determines whether the retrieved region actually supports the benchmark answer.

Every representation receives the same block of full-network target facts. For each node named by the question, the block states whether the node exists and gives its degree in the full network. These facts are identical across representations. The final prompt contains the rendered network region, the shared target-fact block, and the question. A fixed system instruction asks the model to reason through the problem and finish with
\begin{quote}
\texttt{ANSWER: <answer>}
\end{quote}
for the main benchmark.

We treat context length as part of system performance. Each language model is served with a \(40{,}960\)-token context window. We reserve \(16{,}384\) tokens for generation, leaving a maximum prompt length of \(24{,}576\) tokens. Prompt length is measured with the model's own tokenizer and chat template. A prompt that exceeds this budget is not sent to the model.

\subsection*{Network representations}

We hold the retrieved region, question, and language model fixed and vary only how the region is represented.
\begin{itemize}
\setlength{\itemsep}{1pt}
\item \textbf{R0, no graph}: the model receives only the shared target facts and the question.
\item \textbf{R1, edge list}: the region is represented as a list of edges.
\item \textbf{R2, sentences}: the same adjacency information is expressed as natural-language sentences.
\item \textbf{R3, raw measurements}: the region is represented as a table of graph quantities.
\item \textbf{R4, learned roles}: each node is represented by a role learned with an unsupervised graph encoder.
\item \textbf{R5, BioGlyph}: the deterministic compiler converts the region into named structural roles with evidence and structural consequences.
\end{itemize}

The R3 table is designed to expose the numerical inputs used by the BioGlyph compiler. For each node it contains degree, normalized degree, betweenness, PageRank, core number, and community identifier. The remaining node columns are intra-community degree, cross-community degree, participation coefficient, number of communities touched, component size, articulation-point status, and the component count after removal when available. For each edge it contains edge betweenness and bridge status. For each community it contains size and edge connectivity when computed. R3 and R5 are generated from the same structural-signal record. Because R3 prints edge-level quantities for individual edges, those rows also reveal which edges are present. We retain these columns because edge betweenness and bridge status are inputs to BioGlyph edge roles.

\subsection*{Learned-role baselines}

For the learned-role baseline, each node receives an eleven-dimensional input vector. Five graph quantities are also printed with the learned role for interpretation: degree, betweenness, PageRank, core number, and clustering coefficient. Six additional quantities are used by the graph encoder but are not printed: eigenvector centrality, average neighbor degree, maximum neighbor degree, standard deviation of neighbor degree, log triangle count, and two-step random-walk return probability. The feature set contains no articulation, bridge, community, or removal information, because those quantities are what the BioGlyph compiler derives.
We standardize these features and pass them through two message-passing layers with hidden width \(64\), output dimension \(32\), and a PReLU activation. We evaluate four architectures: GCN\cite{kipf2016semi}, GraphSAGE\cite{hamilton2017inductive}, GAT\cite{velivckovic2017graph}, and GIN\cite{xu2018powerful}, implemented with PyTorch Geometric\cite{fey2019fast}.

We train each encoder without task labels using Deep Graph Infomax\cite{velivckovic2018deep}. Let \(z_v\) denote the embedding of node \(v\), and let \(\widetilde{z}_v\) denote its embedding after corruption by row-wise permutation of the input feature matrix. We define the graph summary
\begin{equation}
s=
\varsigma
\left(
\frac{1}{n}
\sum_{v\in V}z_v
\right),
\label{eq:dgi_summary}
\end{equation}
where \(\varsigma\) is the logistic function. With bilinear discriminator
\begin{equation}
D(z,s)=z^{\top}Ws,
\label{eq:dgi_discriminator}
\end{equation}
the training objective is
\begin{equation}
\mathcal{L}
=
-\frac{1}{n}\sum_{v\in V}
\log \varsigma\!\left(D(z_v,s)\right)
-
\frac{1}{n}\sum_{v\in V}
\log
\left[
1-\varsigma\!\left(D(\widetilde{z}_v,s)\right)
\right].
\label{eq:dgi_loss}
\end{equation}
We train for \(200\) epochs with Adam, a learning rate of \(0.01\), and seed \(0\). We then cluster the learned embeddings into \(K=8\) groups using seeded \(k\)-means++\cite{arthur2007k}. Cluster identifiers are relabeled by decreasing cluster size.

The encoder runs once on the full network. We standardize features using full-network means and standard deviations and assign each node to the nearest learned centroid,
\begin{equation}
r(v)
=
\arg\min_{k\in\{1,\ldots,K\}}
\left\|
z_v-\mu_k
\right\|_2^2,
\label{eq:learned_role}
\end{equation}
where \(\mu_k\) denotes centroid \(k\). A retrieved region displays each node with its full-network learned role and the five printed graph quantities, above a legend that names every learned role from the feature profile of its cluster. The main text reports GraphSAGE; results for all four architectures are provided in the Supplementary Information.

\subsection*{Benchmark construction and exact oracle}

The benchmark contains seven question families. \emph{Cut node} asks whether a named node is an articulation point. \emph{Fragmentation} asks how many pieces remain after removing a named node. \emph{Edge removal} asks how many pieces remain after removing an edge. \emph{Reachability} asks whether two nodes remain connected after a third node is removed. \emph{Compare degree} asks which candidate has the highest degree. \emph{Candidate cut} asks which candidate causes the largest fragmentation when removed. \emph{Compositional} asks which candidates satisfy both an articulation condition and a degree condition.

We target twenty questions per family and network using seed \(0\). All candidate pools are drawn from the largest connected component. The cut-node, fragmentation, and edge-removal families draw half of their questions from positive cases and half from negative cases. For reachability we construct both answers directly rather than filtering sampled pairs. Half of the questions remove an articulation point and take the two endpoints from different components of the remaining graph, so the answer is no. The remaining questions draw the removed node and the first endpoint from the largest component. We then take the second endpoint from the nodes lying at distance two or three from the first after the removal, so the answer is yes. Candidate-cut and compositional questions include at least one articulation point among the candidates. Degree-comparison questions draw one candidate from the highest-degree decile and the rest uniformly, resampling when necessary so that the maximum-degree candidate is unique.

For fragmentation, candidate-cut, and compositional questions, we restrict sampled targets to degree at most \(60\). This constraint allows the target neighborhood to fit inside the largest retrieval region. Where no candidate meets the bound on a given network, we fall back to the unbounded pool rather than leave the family empty. We also create unanswerable examples by replacing a referenced node identifier with an identifier that does not occur in the network. When every quota is available, a network contributes \(140\) answerable questions and \(16\) unanswerable questions. Some networks contribute fewer because the required structural cases do not exist. Across eight benchmark networks from five domains, the final benchmark contains \(1{,}239\) questions.

We grade every benchmark family with exact graph algorithms. Let \(\beta(v)\) denote the number of biconnected blocks containing node \(v\). The number of pieces left in the original component after removing \(v\) is
\begin{equation}
\mathrm{pieces}(v)
=
\begin{cases}
0, & d(v)=0,\\
\beta(v), & v \text{ is an articulation point},\\
1, & \text{otherwise}.
\end{cases}
\label{eq:node_pieces}
\end{equation}
For an edge \(e\),
\begin{equation}
\mathrm{pieces}(e)
=
1+
\mathbf{1}
\!\left[
e\text{ is a bridge}
\right],
\label{eq:edge_pieces}
\end{equation}
where \(\mathbf{1}[\cdot]\) is the indicator function. These counts refer to the connected component containing the target rather than to unrelated components elsewhere in the network.

We evaluate each question with two oracle calls. The full-network oracle provides the ground-truth answer used for grading. The region oracle determines what can be inferred from the retrieved region. We call a retrieved region sufficient when the answer obtained from that region agrees with the full-network answer.

\subsection*{Language models, inference and grading}

We evaluate every representation using the same two frozen 8B language models, Qwen3-8B\cite{yang2025qwen3} and Llama-3.1-8B-Instruct\cite{grattafiori2024llama}, served with vLLM\cite{kwon2023efficient} and seed \(0\). Qwen3-8B runs in thinking mode with temperature \(0.6\), following its recommended configuration, whereas Llama-3.1-8B-Instruct uses greedy decoding. For the main benchmark, we read the text following the final \texttt{ANSWER:} marker. If the marker is absent, we parse the complete reply. A deterministic rule-based parser first checks for an explicit statement that the question is unanswerable. Otherwise, it extracts the answer type required by the question family. Exact-match grading is then performed against the graph oracle. When several answers are equally valid, any oracle-approved answer is accepted. A reply with no recoverable answer is scored as incorrect, and so is an over-budget prompt, which reaches the model as an empty reply. A generation that reaches the output limit is parsed like any other reply and is incorrect only when no answer can be recovered from it.

The basic evaluation unit is one prediction for one question, model, retrieval setting, and retrieval cap. Let \(U\) denote the set of evaluation units and \(y_u\in\{0,1\}\) the correctness of unit \(u\). System accuracy is
\begin{equation}
A_{\mathrm{sys}}
=
\frac{1}{|U|}
\sum_{u\in U}y_u .
\label{eq:system_accuracy}
\end{equation}
A unit is correct when the parsed model answer matches the oracle. Questions whose missing identifier is detected before model inference are counted as correct for every representation, because the pipeline reports them as unanswerable without calling a model.

For pairwise comparisons, we also evaluate only units for which both representations fit within the context window and produce a completed reply. For representations \(a\) and \(b\), we define
\begin{equation}
U_{ab}
=
\left\{
u\in U:
\begin{array}{l}
u\text{ is answerable, and}\\
a\text{ and }b\text{ both fit and complete}
\end{array}
\right\}.
\label{eq:controlled_units}
\end{equation}
We score the model replies on \(U_{ab}\) directly. We recompute this common set for every pairwise comparison. For the rendering ablation, raw measurements, names, names with evidence, and the full BioGlyph description share one common feasible set so that all steps are evaluated on the same questions.

\subsection*{Conversation experiments}

\paragraph{Three-turn conversations.}
We send the retrieved region once with the first question, using either BioGlyph or the raw-measurement representation. Turn 2 asks a follow-up about the same region without resending the network representation. Turn 3 challenges the previous answer with a scripted incorrect alternative derived from the oracle answer. For a Boolean answer we negate the correct value; for a count we use a neighboring incorrect value; and for a candidate question we name a candidate excluded by the oracle. The challenge template is fixed within each question family and does not depend on the model's previous response. We run the conversation turn by turn and append the visible model response to the conversation history. Hidden reasoning traces are not added to later prompts. These experiments reserve \(8{,}192\) tokens for each model response, leaving a prompt budget of \(32{,}768\) tokens.

\paragraph{Eight-turn conversations.}
Each eight-turn thread starts from one retrieved region and one BioGlyph description. We use four turn types. Graded turns have an exact graph-oracle answer. Lookup turns ask for information stated directly in the BioGlyph description. Opinion turns ask for an explanation and are not scored. Push-back turns challenge a previous answer with an incorrect alternative and are paired with the earlier turn for before-and-after analysis. Gold answers are computed on the region shown to the model, except for full-network degree facts that are explicitly supplied in the shared fact block. The region description is sent only at the first turn. These experiments reserve \(4{,}096\) tokens for each response, leaving a prompt budget of \(36{,}864\) tokens.

\paragraph{Natural-language questions.}
We asked the same questions about the same retrieved regions, but replaced the benchmark answer format with a request for a short answer in ordinary language. We extracted the final answer with a deterministic rule-based parser and graded it with the same exact graph oracle used in the main benchmark. The parser handles negation and ignores node identifiers or intermediate numbers that are not part of the final answer.

\paragraph{Push-back outcomes.}
For every completed challenge pair, we record correctness before and after the challenge. We report four outcomes: the model held a correct answer, corrected a wrong answer, abandoned a correct answer, or remained wrong. Both turns must complete for the pair to enter this analysis.

\subsection*{Biological validation}

\paragraph{Yeast gene essentiality.}
We match STRING-Yeast proteins to SGD gene-essentiality annotations\cite{cherry2012saccharomyces}. Neither the BioGlyph compiler nor the language models receive these annotations. For this analysis the compiler runs once on the whole STRING-Yeast network, so the threshold roles use full-network comparison distributions rather than regional ones. For each node-level structural role, we compare the fraction of essential proteins among nodes carrying that role with the essential fraction across the full labeled network. We use two-sided Fisher exact tests and adjust across roles with the Benjamini--Hochberg procedure\cite{benjamini1995controlling}. Adjusted \(P\) values are included with the released results. To compare the cross-community connector role with individual centrality measures, we construct equally sized groups from the highest-ranked proteins by degree, betweenness, and PageRank.

\paragraph{Reactome perturbation screen.}
We apply the structural screen to the largest connected component of the Reactome functional interaction network\cite{wu2010human}. For each protein \(v\), we measure how many proteins separate from the largest remaining component after \(v\) is removed. Let \(N\) denote the number of proteins in the original component and \(\mathcal{C}(G-v)\) the connected components after removal. We define
\begin{equation}
s(v)
=
(N-1)
-
\max_{C\in\mathcal{C}(G-v)}
|C|.
\label{eq:detachment_score}
\end{equation}
For a non-articulation point, \(s(v)=0\).
We compare this structural ranking with DepMap 24Q4 gene-effect data\cite{tsherniak2017defining,dempster2021chronos}, which reports \(17{,}916\) screened genes across \(1{,}178\) cancer cell lines. The comparisons below are restricted to the proteins of the Reactome main component that DepMap screened. For each protein, the dependency fraction is the fraction of screened cell lines with Chronos score below \(-0.5\).

We use four descriptive dependency categories in Fig.~\ref{fig:reactome}. A protein on the DepMap common-essential list is labeled \emph{common essential}. Among proteins not on that list, a dependency fraction below \(5\%\) is labeled \emph{rarely a dependency}. A fraction from \(5\%\) to \(50\%\) is labeled \emph{selective dependency}, and a fraction above \(50\%\) is labeled \emph{a dependency in most lines}.
We evaluate the structural screen in two ways. First, for \(k\in\{10,25,50,100,200,300\}\), we compute the fraction of selective dependencies among the top \(k\) proteins ranked by \(s(v)\). We compare this curve with rankings by betweenness and degree and with the overall selective-dependency rate. Second, we compare screened cut nodes with all other screened proteins for selective dependency and common essentiality using one-sided Fisher exact tests.

\paragraph{Counting perturbation size.}
We select the sixty proteins with the highest \(s(v)\) values and retrieve a module containing between \(20\) and \(120\) proteins around each target. Each module is rendered as an edge list, adjacency sentences, raw measurements, and BioGlyph. The prompt names the candidate protein, states its degree within the retrieved module, and asks how many proteins detach after the candidate is removed.

The exact gold count is computed on the retrieved module rather than on the full Reactome network. Let \(\widehat{y}\) be the predicted count and \(y\) the exact module-level count. We score a prediction as correct when
\begin{equation}
|\widehat{y}-y|\leq 2.
\label{eq:count_tolerance}
\end{equation}
A truncated response or a response from which no count can be recovered is incorrect. Among replies containing a recoverable count, we also report the median absolute error \(|\widehat{y}-y|\).

We separately evaluate structural statements in the model's reasoning. A sentence claiming that removal of one named node disconnects the module can be checked directly against the exact cut nodes of that module. We evaluate only sentences that refer to exactly one node in the retrieved region. Correct denials and correct assertions receive equal credit. If \(T\) of \(M\) checkable claims agree with the exact graph analysis, we report
\begin{equation}
F=\frac{T}{M}.
\label{eq:claim_faithfulness}
\end{equation}

\subsection*{Statistical analysis}

All contrasts are paired on the evaluation unit by an inner join, so a unit attempted by only one representation contributes nothing. For confidence intervals within one network, we use percentile bootstrap intervals\cite{efron1992bootstrap} from \(3{,}000\) resamples with seed \(0\), resampling evaluation units with replacement. For effects pooled across networks, we use a two-stage bootstrap with \(5{,}000\) resamples and seed \(0\). We first sample networks with replacement, and we then resample evaluation units with replacement within each sampled network. The point estimate remains the plain paired mean, and only the interval changes. The procedure accounts for between-network variation, because questions drawn from one network come from the same graph and often from the same neighborhoods. Units that share a question stay correlated inside the second stage, so the interval is not corrected at that finer level of clustering.
For pairwise representation contrasts, we construct the common feasible set of evaluation units before resampling. We also report the mean per-network effect with a \(t\)-interval across the eight benchmark networks, which leaves seven degrees of freedom.

\subsection*{Datasets and implementation}

\paragraph{STRING-Yeast.}
We use the budding-yeast physical interaction network from STRING v12\cite{szklarczyk2023string} and retain interactions with combined score at least \(700\). Nodes represent proteins and edges represent physical associations.

\paragraph{ChCh-Miner.}
We use the BioSNAP ChCh-Miner drug interaction network\cite{zitnik2018biosnap}. Nodes represent drugs and edges represent reported drug--drug interactions.

\paragraph{ego-Facebook.}
We use the ego-Facebook friendship network from SNAP\cite{leskovec2014snap}. Nodes represent people and edges represent friendships.

\paragraph{email-Eu-core.}
We use the email-Eu-core network from SNAP\cite{leskovec2014snap}. Nodes represent anonymized members of a European research institution and edges represent email interactions. Department annotations are not used.

The eight-network benchmark that yields the \(1{,}239\) questions comprises STRING-Yeast, ego-Facebook, email-Eu-core, Wiki-Vote, Amazon-Photo, ogbn-arxiv, Cora, and Coauthor-CS, drawn from five domains. ChCh-Miner is evaluated under the same protocol but sits outside that pool, so its questions are additional to the \(1{,}239\). Section~\ref{sec:suppresults} describes the remaining networks and their preprocessing.

\paragraph{Reactome FI.}
We use the Reactome functional interaction network\cite{wu2010human} for the perturbation screen rather than for the main seven-family benchmark. We map proteins to DepMap using approved gene symbols\cite{tweedie2021genenames}.

\paragraph{Implementation.}
The BioGlyph compiler and graph-analysis pipeline run on CPU using NetworkX\cite{hagberg2008exploring} and igraph. The benchmark oracle computes cut nodes, bridges, connected components, and perturbation answers exactly. Only betweenness on graphs larger than \(5{,}000\) nodes is sampled, as described above. We use seed \(0\) for stochastic procedures, including community detection, sampled betweenness, graph-encoder initialization, \(k\)-means clustering, question sampling, language-model decoding where applicable, and bootstrap resampling. The Supplementary Information reports the additional networks and graph encoders (Sections~\ref{sec:suppresults} and \ref{sec:s_learned}), the further model families (Section~\ref{sec:s_families}), the retrieval controls (Section~\ref{sec:s_retrieval}), the trained-model reference (Section~\ref{sec:s_learned}), and the pre-registered candidate-knockout screen (Section~\ref{sec:s_cko}).

%% file: sections/supplementary.tex
\section{The networks of the main text, and the result pooled over the whole benchmark}
\label{sec:suppresults}

Table~\ref{tab:datasets} describes the four networks that the main text takes one at a time. For each network it states what a node is, what an edge means, and how large the network is. Two of the columns need a word of explanation. A cut node is a node whose removal breaks its part of the network into separate pieces, and a bridge is an edge that provides the only route between two sides. We computed both exactly rather than estimating them, so the two counts together show how much of a network rests on a single point. The yeast interactome contains 363 cut nodes and 553 bridges among its 3{,}384 proteins. Ego-Facebook contains only 11 cut nodes among 4{,}039 people, so almost nothing in that network depends on one person. The final column names the label that comes from outside the network itself. We never showed any of these labels to a model, which is what allows us to use one of them, SGD gene essentiality, as an independent check on the compiled roles. Table~\ref{tab:s_networks} lists all twenty benchmark networks with their sizes.

\begin{figure}[p]
\centering
\includegraphics[width=\linewidth]{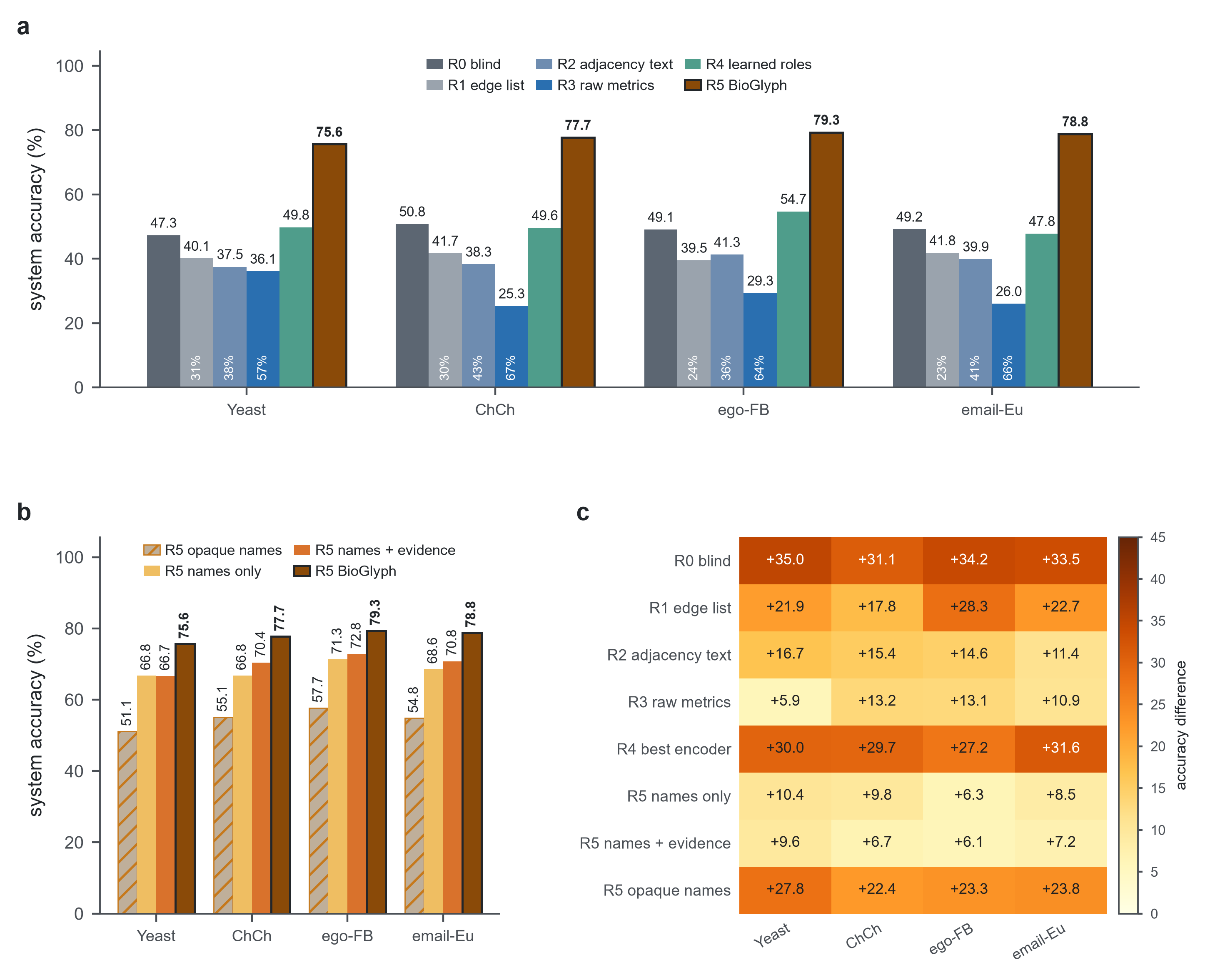}
\caption{\textbf{Every representation on the four main-text networks.} Results pool Qwen3-8B and Llama-3.1-8B throughout.
\textbf{a} System accuracy of the six competing representations; the R4 bar is the mean of the four encoders. Where more than 5\% of an arm's prompts exceeded the token budget, the share printed inside the bar never reached the model.
\textbf{b} The same four networks for the three \bioglyph{} rendering steps and for the opaque-name control, which is the role-name rendering with every role name replaced by a meaningless token.
\textbf{c} The accuracy difference between \bioglyph{} and each arm of \textbf{a} and \textbf{b}, on the questions both arms fitted into the window and finished; the R4 row takes the best of the four encoders.}
\label{fig:S1}
\end{figure}

Figure~\ref{fig:S1} shows every representation on all four networks under both ways of scoring, and \bioglyph{} leads on each of them. We then pooled the whole benchmark, which asks 1{,}239 questions over eight networks from five domains and answers each of them with two frozen 8B models, Qwen3-8B and Llama-3.1-8B. Across that pool the models answered 70.6\% of the questions correctly from \bioglyph{} (95\% CI 69.4 to 71.9). When we showed them no network at all they answered 50.7\%, and this blind floor is the number that any representation has to beat. Most representations do not beat it. The edge list reached 48.7\%, adjacency text 49.4\%, and the four learned-role encoders 49.1 to 50.0\%, so close to the floor as to be indistinguishable from it. The raw-metric table fell below the floor at 39.5\%, and length explains why: 41.7\% of its prompts exceeded the 24{,}576-token budget and never reached the model at all. \bioglyph{} beat every encoder and the raw-metric table in all 24 of the network-by-model combinations we ran, and it beat adjacency text in 18 of them. It came within one point of a graph neural network that we trained directly on the benchmark questions (74.1 against 74.8, $n=381$), and it stayed below a retrieval ceiling of 82.6\%.

We also scored the formats on only those questions that both of them could deliver. Across the pool the models answered 66.8\% of those questions from \bioglyph{} and 61.4\% from the raw-measurement table. We then unfolded the description into the same three rungs that we used on the yeast interactome and scored all of them on one common set of questions. The role names alone brought the models to 61.9\%, adding the measurements behind each name left them at 62.0\%, and adding the line that says what removal would cost lifted them to 68.4\%. The last rung raises the score on 19 of the 20 networks that carry the ladder. Twelve further networks carry the supporting studies.

\begin{figure}[p]
\centering
\includegraphics[width=\linewidth,height=0.88\textheight,keepaspectratio]{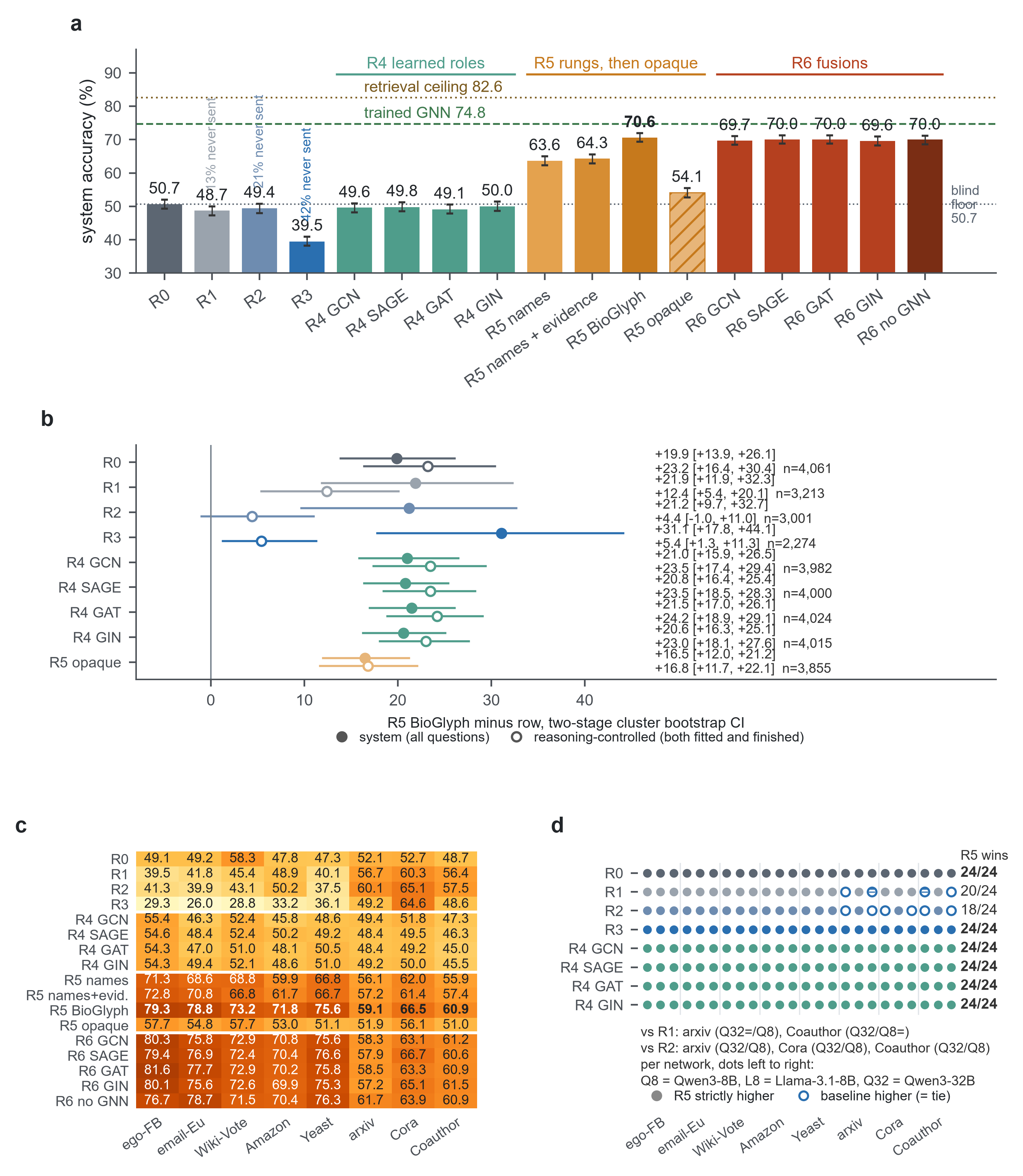}
\caption{\textbf{The pooled result over the eight benchmark networks.} Results pool Qwen3-8B and Llama-3.1-8B unless labeled. \textbf{a} System accuracy of every arm with 95\% bootstrap confidence intervals; reference lines mark the no-network floor, the trained graph neural network and the retrieval ceiling. \textbf{b} \bioglyph{} against every baseline, on all questions and on the questions both arms fit and finished, with two-stage intervals over networks. \textbf{c} System accuracy per network. \textbf{d} The environment sweep: which arm wins in each network-by-model cell. Table~\ref{tab:s_headline} gives every arm for each of the three models separately.}
\label{fig:S2}
\end{figure}

Figure~\ref{fig:S2} and Table~\ref{tab:s_headline} give the pooled result in full: every arm under both ways of scoring, with each arm's overflow share and prompt length. Every arm here includes the four learned-role encoders separately, the five fusion arms and the opaque control. In each of the three models taken alone, \bioglyph{} leads and the raw-measurement table trails every other arm. The environment sweep locates the six losses to adjacency text: all of them sit on the three sparse citation and co-authorship graphs, where a retrieved region is small enough to read directly.

Three model families beyond the two 8B readers took the same benchmark, and the main text keeps them out of its figures so that every panel there rests on one pool. On the drug interaction network all four families answered far more from a \bioglyph{} description than from any conventional format. Qwen3-8B reached 87.5\%, Gemma-3-12B 71.8\%, Llama-3.1-8B 67.9\% and Mistral-Nemo-12B 57.7\%. Given a table of raw measurements instead, the same four managed 18.6\%, 18.6\%, 32.1\% and 15.1\%. Scale helps as well as format. On the yeast interactome Qwen3-32B answered 90.7\% of the questions from a \bioglyph{} description and 38.8\% from the measurement table.

\input{tables/tab2_datasets}
\input{tables/tabS1_networks}
\input{tables/tabS2_headline}

\section{Talking to \bioglyph{}: the exchanges behind the numbers}
\label{sec:talking}

The main text reports accuracies. This section shows the exchanges that produced them. Figure~\ref{fig:qa} gives one exchange, and Figs.~\ref{fig:S3a} to \ref{fig:S3f} give many more, each printed on a page of its own. Together they show what a graded turn, a lookup, an opinion and a push-back actually look like, and what a model writes when it is right and when it is wrong.

Every card in these figures is a stored reply from the runs that this paper scores. We rewrote nothing. We removed the markdown, collapsed the bullet lists, and where a reply ran long we cut it at the end of a sentence and marked the cut. Every verdict is the one the paper's evaluator gave, never our own reading of the reply. We also selected the cards deterministically: the released \texttt{qa\_items.json} records the rule that picked each card, together with the file, the identifier and the model behind it.

The models saw less than the cards show. Node identities, such as yeast ORF and gene names or ICD-9 codes, sit outside the exchange, and so do the outside labels such as SGD essentiality and DepMap dependency. No model ever received any of them. Email-Eu-core is anonymized at source, so it has no identities to show in the first place. We sent the compiled description once, with the first question, and never sent it again.

Turns come in four kinds, and we keep them apart both on the cards and in every number we report. A graded turn has an exact answer from the oracle, so we score it. A lookup turn asks for something that the description already states, which tests whether the model can read rather than whether it can reason. An opinion turn asks why a node matters; it has no right answer, and we never score it. A push-back turn asserts a wrong answer, and we score the model's response against the turn that it disputes.

Figure~\ref{fig:S3a} takes the widest view, with six single questions on three of the main networks, each one shown in all four formats side by side. Every prompt carries its length in tokens, so a reader can see what we gave the model as well as what it answered. The next three figures each follow one conversation from beginning to end: a yeast link (Fig.~\ref{fig:S3b}), a disease-network node (Fig.~\ref{fig:S3c}) and an email-network link (Fig.~\ref{fig:S3d}). We picked the first two mechanically, taking the first thread of its kind, in identifier order, that met the stated rule. We picked the email thread for a different reason: it contains both a misread lookup and a capitulation under push-back, and the failures belong in the record as much as the successes do.

Figure~\ref{fig:S3e} sets the raw-metric table beside \bioglyph{} on the same three-turn conversation, twice. In the first conversation the table is already too long at turn 1 and never reaches the model, which is the common case on the yeast interactome. In the second both formats fit and both answer. The same figure then asks one question twice, once in benchmark style and once as ordinary prose, because a description that worked only in benchmark phrasing would be of little practical use. Figure~\ref{fig:S3f} closes the section with a knockout table that a biologist can check. It lists named proteins in Reactome-FI modules, the oracle's count of proteins that detach when we remove each one, each model's answer, and DepMap's verdict beside it. Two honest failures follow, and we give the pooled push-back rates beneath them, so that the failures read as examples of a known frequency rather than as anecdotes.

\input{figures/Supp_Fig/figS3/figS3_gallery}

\section{The raw-measurement table: where it fails, why, and what recovers it}
\label{sec:s_r3}

\begin{figure}[p]
\centering
\includegraphics[width=\linewidth,height=0.88\textheight,keepaspectratio]{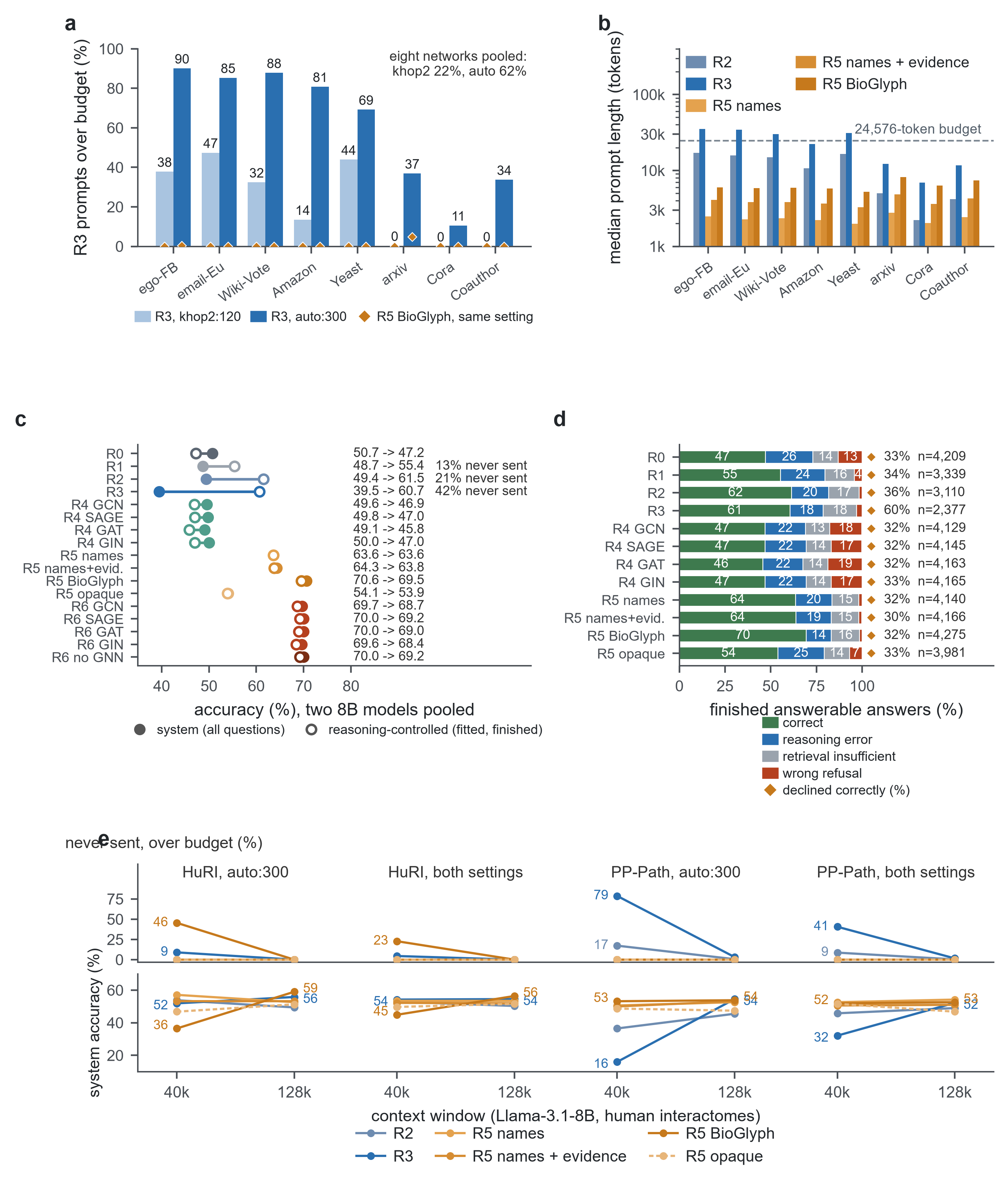}
\caption{\textbf{The raw-measurement table in full.} Results pool Qwen3-8B and Llama-3.1-8B unless labeled. \textbf{a} Share of raw-measurement prompts over the 24{,}576-token budget per network and retrieval setting, with \bioglyph{} at the same setting. \textbf{b} Median prompt length of every arm per network against the budget; Table~\ref{tab:s_r3} gives the lengths themselves. \textbf{c} System accuracy and reasoning-controlled accuracy of every arm, with the share of prompts never sent. \textbf{d} What the finished answerable answers got wrong, per arm, and the share of finished unanswerable questions the model itself declined. \textbf{e} The 131{,}072-token window on the two human interactomes (Llama-3.1-8B): prompts never sent above, system accuracy below, at both windows and both retrieval settings.}
\label{fig:S4}
\end{figure}

The raw-measurement table is the control a careful reader asks about first, because it carries every fact the compiled description carries and more (Methods). Figure~\ref{fig:S4} and Table~\ref{tab:s_r3} give its anatomy. The table exceeds the 24{,}576-token prompt budget on 41.7\% of all prompts: 21.8\% at the tighter retrieval setting and 61.6\% at the wider one. The same regions rendered as \bioglyph{} exceed the budget on 0.3\%. The table is the longest rendering on every network, with a median of 19{,}723 tokens against 6{,}229 and a 90th percentile of 79{,}682 against 15{,}224. Its median length on the same regions runs from 1.1 times the description's on the sparsest network to 7.0 times on the yeast interactome at the wider setting.

Restricting the score to the questions an arm fit and finished lifts every long arm, and the table most of all, from 39.5\% to 60.7\%; \bioglyph{} reaches 69.5\% on its own feasible set. On the questions the two arms share, the table reaches 61.4\% against 66.8\% for \bioglyph{}, and the pooled two-stage interval for the difference excludes zero. The gap is wider at the tighter retrieval setting than at the wider one (Table~\ref{tab:s_r3}). Among the answerable replies that finished, the table mistakes its reasoning on 18.0\%, against 20.2\% for adjacency text, 24.4\% for the edge list and 13.6\% for \bioglyph{}. The table is also the arm most willing to declare a question unanswerable when it is, declining 60.2\% of the finished unanswerable questions against 32.1\% for \bioglyph{}. A table that prints a row for every node makes a missing node conspicuous by its absence. A description lists only the nodes that earn a role, and beyond the region list at its head it offers no such cue.

Two interventions recover the table, and both recover it to parity rather than beyond. Qwen3-32B reads the table at 41.7\% and \bioglyph{} at 82.3\% over all questions, yet on the questions both arms fit the two stand level, at 77.0\% and 76.9\%. A 131{,}072-token window on the two human interactomes removes the overflow: on PP-Pathways the table's overflow falls from 40.7\% to 1.6\% and its accuracy rises from 32.1\% to 52.2\%, level with \bioglyph{} at 52.6\%. On HuRI the description itself exceeds the standard budget on 22.8\% of prompts, because the interactome's hubs earn many roles, and the wide window brings both arms level there as well (Fig.~\ref{fig:S4}). The honest reading is the one the main text gives. An 8B model reasons somewhat better over named roles than over the same facts as numbers, and far better over a description that fits than over one that does not. A larger model or a wider window removes the second effect and most of the first.

\input{tables/tabS3_r3}

\section{Which layer of the description does the work: the ladder on every network}
\label{sec:s_ladder}

\begin{figure}[p]
\centering
\includegraphics[width=\linewidth,height=0.88\textheight,keepaspectratio]{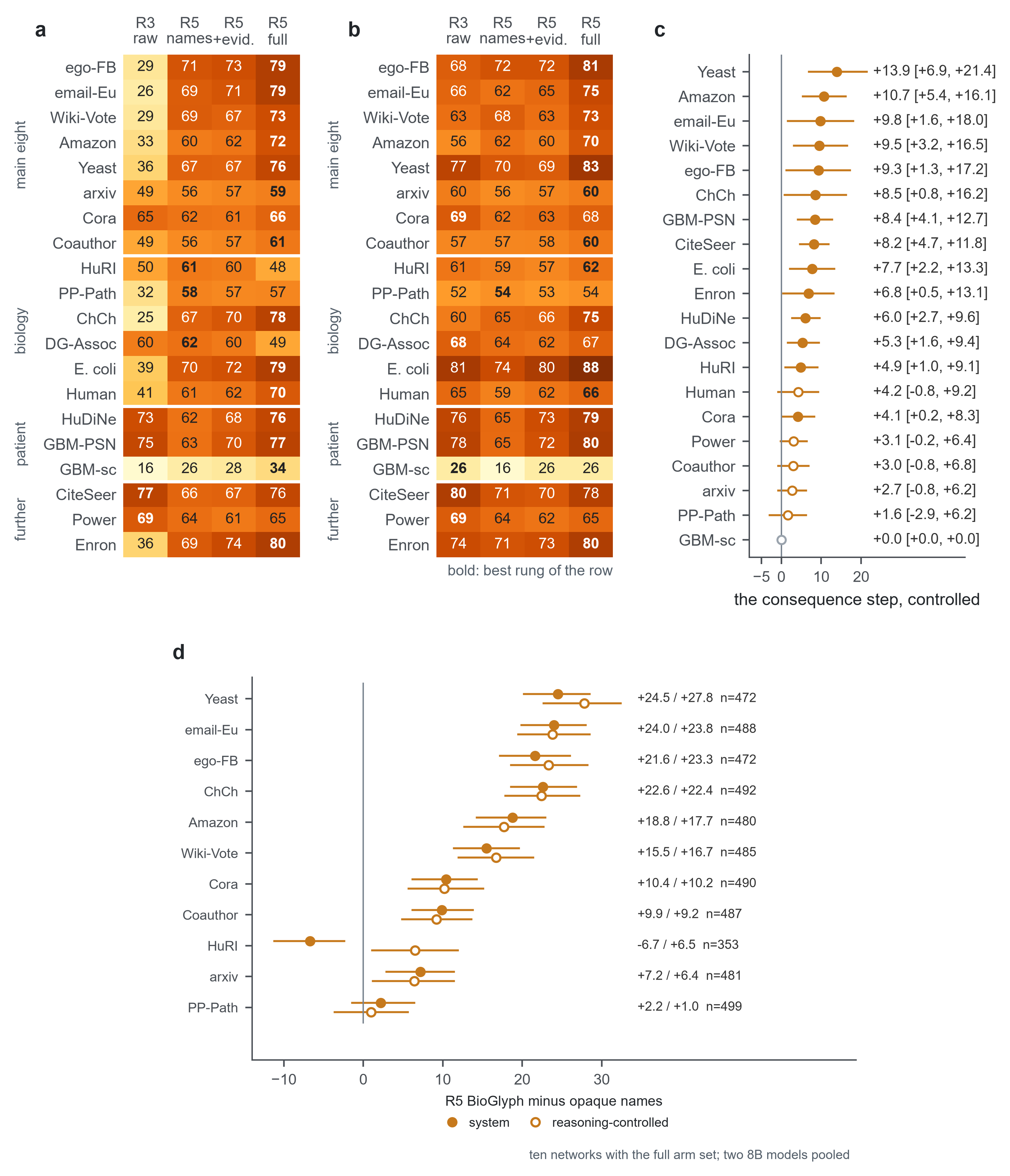}
\caption{\textbf{The ladder on every network.} Results pool Qwen3-8B and Llama-3.1-8B. The four rungs are the raw-measurement table, role names alone, names with the measurements behind them, and the full description with its stated consequences. \textbf{a, b} System accuracy and accuracy on the common feasible set of the four rungs, on all twenty networks, grouped by domain. \textbf{c} The consequence step per network with 95\% intervals; a filled marker means the interval excludes zero. \textbf{d} The full description against the opaque-name control on every network where the control ran; the control is the names rung with its names replaced, so this gap carries the evidence and the stated consequences as well as the words. Table~\ref{tab:s_ladder} gives the three steps for every study.}
\label{fig:S5}
\end{figure}

The ladder unfolds the compiled description into four rungs that carry the same computed information: the raw-measurement table, role names alone, names with evidence, and the full description with its stated consequences (Methods). Figure~\ref{fig:S5} and Tables~\ref{tab:s_ladder} and \ref{tab:s_rungs} give the rungs on all twenty networks, on all questions and on one common feasible set. On the eight main networks that set holds 2{,}333 units, and accuracy moves from 63.6\% at the table to 61.9\% with names, 62.0\% with evidence and 68.4\% with the consequence. Renaming the table as bare roles costs a small model a little, and adding the measurements returns almost nothing. Stating what each role implies earns the rung its lead, and only that last step's interval excludes zero.

The pattern holds beyond the main pool. The consequence step is positive on nineteen of the twenty networks and is the largest of the three steps on seventeen. The exceptions are PP-Pathways, HuDiNe and the single-cell graph, where the naming or the evidence step is as large or larger. Across the eight studies of Table~\ref{tab:s_ladder} the consequence step is positive in seven, and its interval excludes zero in five. The naming step is positive in three, and its interval excludes zero on the negative side in three.

The opaque control comes at the same question from the other side. The control is the names rung line for line, with each role name replaced by a meaningless token, so the contrast that isolates the words is a contrast with that rung. Eleven networks carry the control, and the named rung leads on every one. At the extremes, the names lift system accuracy from 55.1\% to 58.0\% on PP-Pathways and from 51.1\% to 66.8\% on the yeast interactome. The named prompt is the longer of the two throughout, so the gain is not a saving of room. Against the full description the control trails on ten of the eleven networks; the exception is HuRI, where the full description exceeds the budget on 25.6\% of prompts and the shorter control does not. The two readings agree. Turning a table into bare role names gains nothing by itself, but once the roles are there, whether they arrive as words or as symbols decides most of what the rung is worth.

\input{tables/tabS4_ladder}
\input{tables/tabS5_rungs}

\section{The sixteen further networks}
\label{sec:s_morebio}

\begin{figure}[p]
\centering
\includegraphics[width=\linewidth,height=0.88\textheight,keepaspectratio]{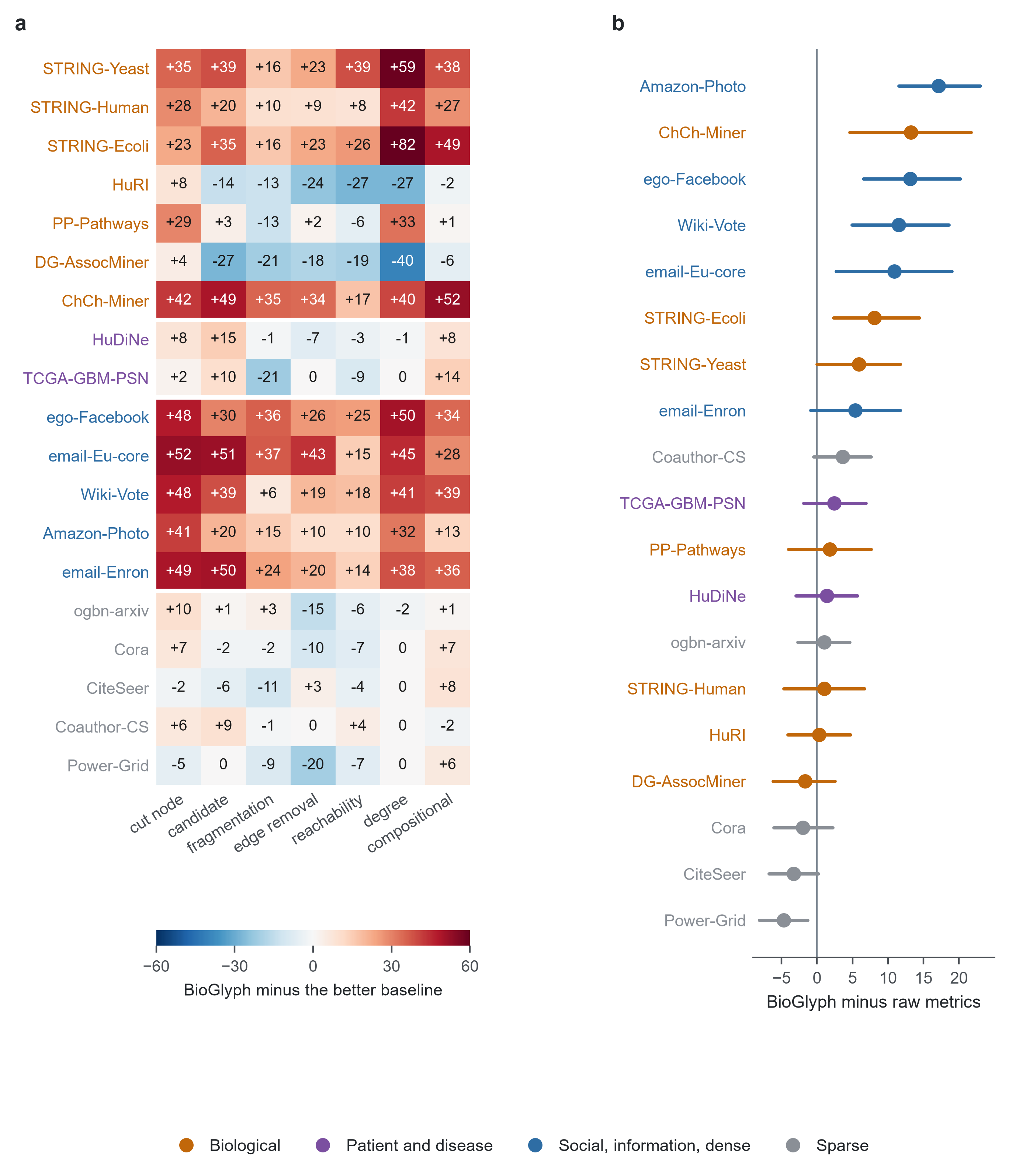}
\caption{\textbf{Where the description helps, by network and by question family.} Results pool Qwen3-8B and Llama-3.1-8B. Row colors give the domain throughout. \textbf{a} System accuracy of \bioglyph{} minus the better of adjacency text and the raw-measurement table, for each of the nineteen networks that carry all seven question families. Red is an advantage to the description, blue to the baseline. \textbf{b} The same comparison as one paired contrast per network, on the questions both arms fitted and finished, with 95\% intervals, sorted. Table~\ref{tab:s_more} lists every arm on every network, including the single-cell graph and the two harder co-community sets, which carry one family each and so do not appear in \textbf{a}.}
\label{fig:S6}
\end{figure}

Five biological networks beyond the yeast interactome and the drug network carry the same benchmark (Fig.~\ref{fig:S6}, Table~\ref{tab:s_more}). The two further STRING interactomes repeat the yeast picture. On STRING-Ecoli \bioglyph{} reaches 79.3\% against 39.3\% for the raw table, which loses 54.2\% of its prompts to the budget; the controlled comparison stands at 84.7\% against 76.6\%, with an interval that excludes zero. On STRING-Human the system accuracies are 69.7\% against 41.2\%, and the controlled comparison is level at 63.3\% against 62.3\%.

\begin{figure}[p]
\centering
\includegraphics[width=\linewidth]{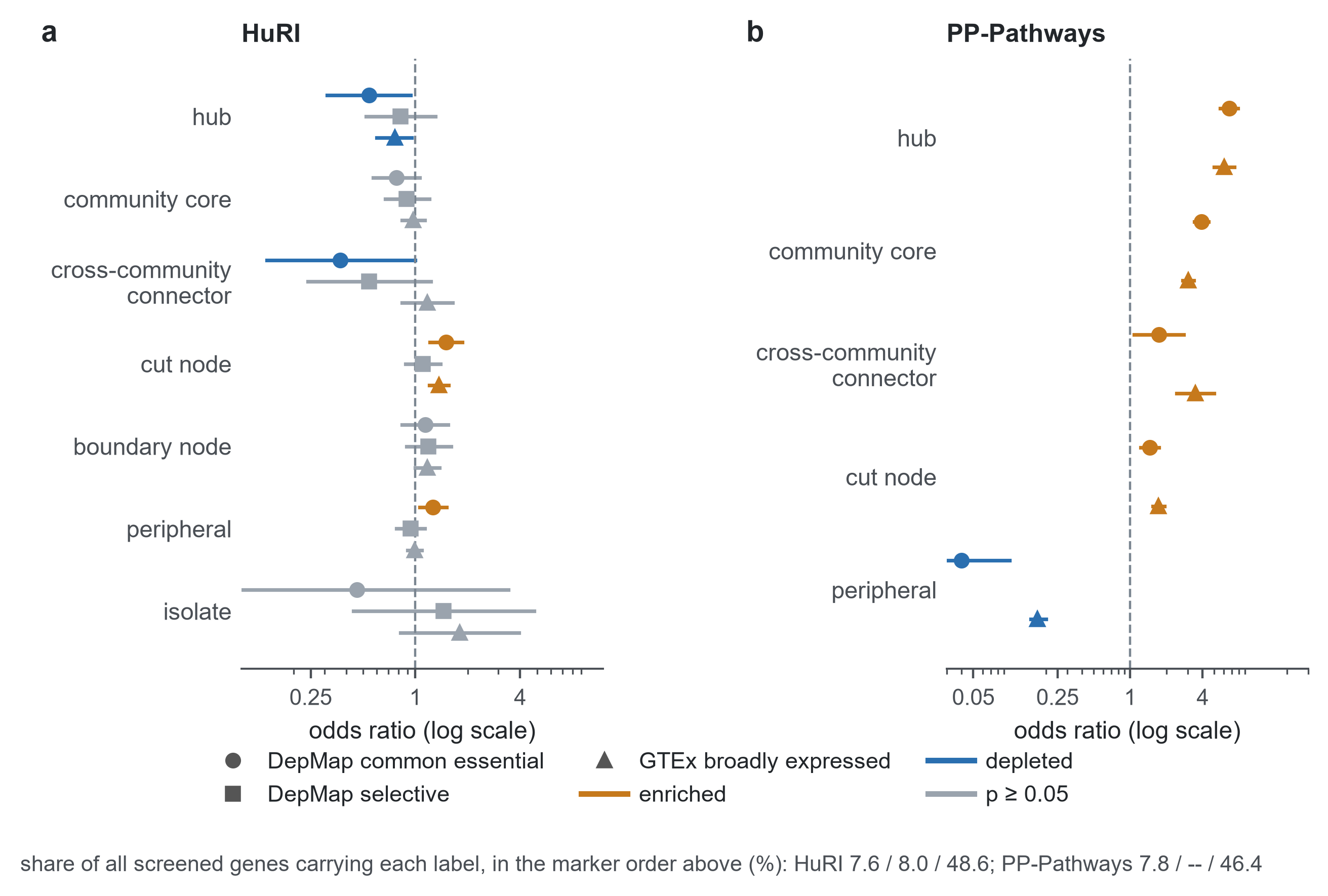}
\caption{\textbf{The compiled roles of the two human interactomes, against DepMap and GTEx.} Odds ratio per role with its 95\% interval, one marker per outside label. Amber marks enrichment, blue depletion, grey a test that does not separate at $p<0.05$. \textbf{a} HuRI. \textbf{b} PP-Pathways, which DepMap does not screen for selective essentiality. No language model reads anything in this figure.}
\label{fig:S7}
\end{figure}

The two networks assembled from binary or curated human interactions behave differently. On PP-Pathways \bioglyph{} reaches 57.4\% against 32.4\%, because the table overflows on 45.8\% of prompts, and once both arms fit the two stand level at 51.1\% against 49.3\%. On HuRI the description itself overflows on 25.6\% of prompts, because the interactome's hubs earn many roles. Its system accuracy of 48.4\% falls below the table at 49.7\% and below its own names-only rung at 60.7\%, and the controlled comparison is level. The disease--gene network DG-AssocMiner is the one network where the description trails on all questions by more than an overflow story. It stands at 48.9\% against 59.5\% for the table and 64.3\% for adjacency text, and level once both fit at 63.6\% against 65.3\%. The bipartite structure gives a region few of the roles the vocabulary was built for, and we report the miss as it is. The biological check on HuRI runs in the yeast direction. Cut nodes are enriched for common-essential genes, at 10.4\% against a background of 7.6\%, with an odds ratio of 1.5 (95\% CI 1.21 to 1.86); Fig.~\ref{fig:S7} gives the enrichment by role for both interactomes. A 131{,}072-token window brings both arms level on these two interactomes (Fig.~\ref{fig:S4}\textbf{e}).

\FloatBarrier

\input{tables/tabS6_more}

Three graphs built from patients rather than molecules carry the same benchmark (Table~\ref{tab:s_more}). They are the HuDiNe comorbidity network of ICD-9 diseases, a glioblastoma patient-similarity network fused from three omics views of the TCGA cohort, and a single-cell graph built from glioblastoma cells. On the two patient-similarity graphs the retrieved regions are small and regular, every arm fits, and the raw table runs close to the compiled description under both views. On HuDiNe the system accuracies are 75.6\% against 73.1\% and the controlled ones 76.5\% against 75.1\%. On TCGA-GBM-PSN they are 77.4\% against 75.2\% and 77.6\% against 75.2\%. The ladder is unusual on both: the naming step is strongly negative, and the evidence step, negligible elsewhere, is one of the largest gains. The single-cell graph is hard for every arm, at 34.1\% for \bioglyph{} against 15.9\% for a table that overflows on 61.4\% of prompts, and its controlled contrast rests on 27 units. Two harder question sets built on the same patient graphs ask about co-community membership and are answerable from the region far less often. On those the description leads once both arms fit, at 17.1\% against 7.1\% on the disease network and 35.2\% against 18.3\% on the patient-similarity network.

The remaining eight networks split into three further dense graphs and five sparse ones, and the split is the result (Fig.~\ref{fig:S6}, Table~\ref{tab:s_more}). On Amazon-Photo, Wiki-Vote and email-Enron the main-text picture repeats. \bioglyph{} reaches 71.8\%, 73.2\% and 79.6\% against a raw table at 33.2\%, 28.8\% and 36.5\% that loses about half of its prompts. It stays ahead once both arms fit, at 70.2\% against 53.2\%, 69.9\% against 58.5\% and 77.1\% against 71.7\%, and the consequence step runs from 6.8 to 10.7 across the three.

On the five sparse graphs the retrieved region is small, almost nothing overflows, and the raw view is readable. There the table stands ahead of the description once both fit on Cora, CiteSeer and Power-Grid, at 67.3\% against 65.3\%, 78.7\% against 75.4\% and 67.4\% against 62.7\%, and only the Power-Grid interval excludes zero. Adjacency text is ahead of the description on ogbn-arxiv on all questions. The consequence step stays positive on all five sparse graphs, but the naming step is negative on four of them, so the ladder nets to little. A description helps where the raw view is unmanageable, not everywhere, and we report the absence of an effect as plainly as the effect.

\section{Model families and scale, and the frontier ceiling}
\label{sec:s_families}

\begin{figure}[p]
\centering
\includegraphics[width=\linewidth,height=0.88\textheight,keepaspectratio]{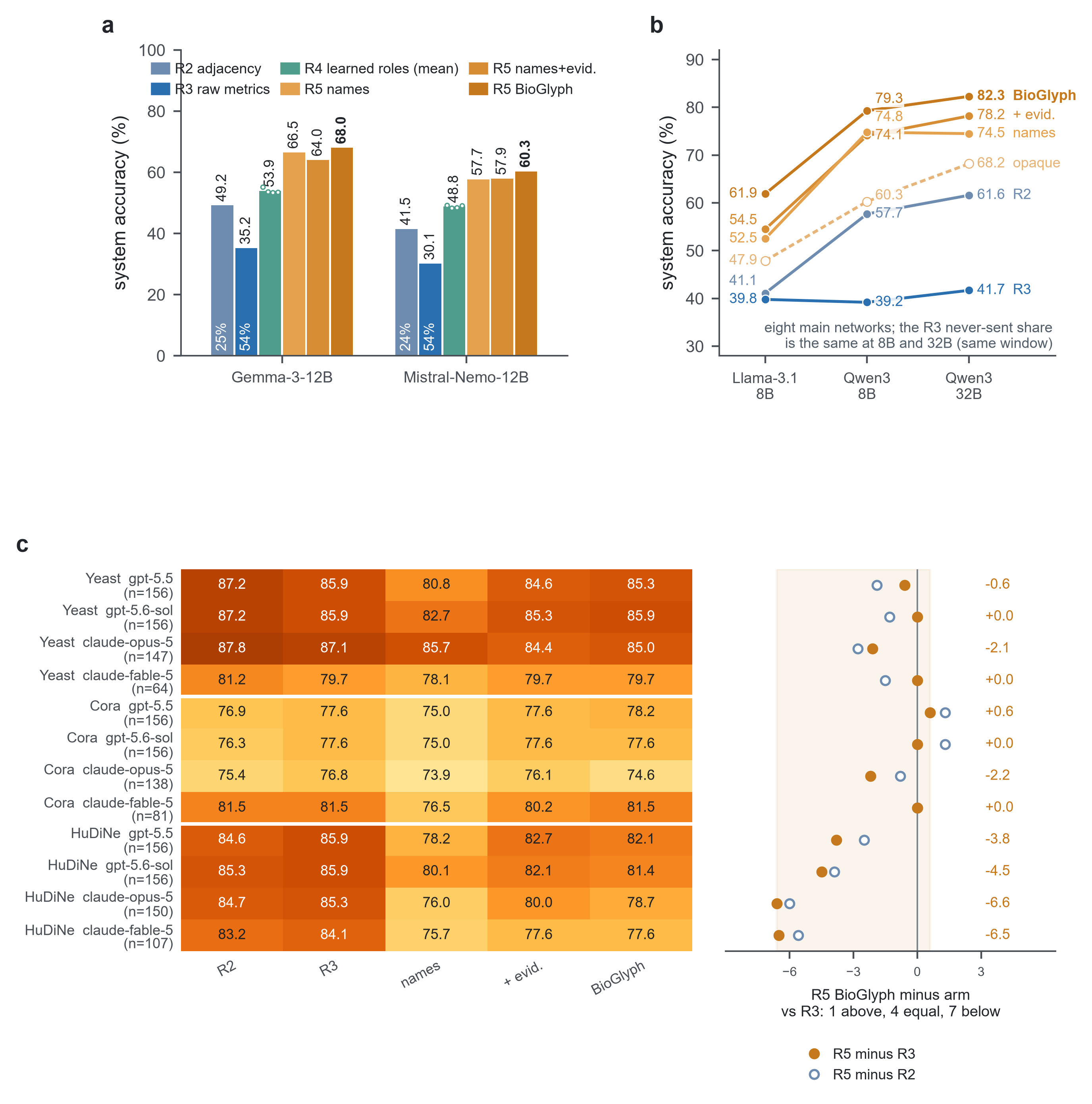}
\caption{\textbf{Model families and scale.} \textbf{a} Gemma-3-12B and Mistral-Nemo-12B on the eight main networks at both retrieval settings; the share printed inside a bar never reached the model. \textbf{b} Qwen3-32B against the two 8B models, per arm. \textbf{c} The four hosted frontier readers on three networks with their own context windows, where nothing overflows and the description and the table read equally well. Table~\ref{tab:s_controls} gives the controlled comparisons for both further families.}
\label{fig:S8}
\end{figure}

Two further open model families read the eight main networks on a reduced set of arms: adjacency text, the raw table, the four encoders and the three compiled rungs (Fig.~\ref{fig:S8}, Table~\ref{tab:s_controls}). Table~\ref{tab:s_controls} also carries the controls of Sections~\ref{sec:s_learned} and \ref{sec:s_retrieval}, so it collects every arm that no figure of its own reports. Both keep the order. Gemma-3-12B goes from 35.2\% on the raw table and 49.2\% on adjacency text to 68.0\% on \bioglyph{}, and Mistral-Nemo-12B from 30.1\% and 41.5\% to 60.3\%. Once both arms fit, Mistral keeps its lead over the table, at 56.3\% against 49.8\% with an interval that excludes zero, and over adjacency text at 55.1\% against 46.6\%. Gemma stands at parity under the same control, at 60.2\% against 62.6\% and 61.3\% against 58.5\%, with both intervals crossing zero. Qwen3-32B lifts every arm, leaves the ordering unchanged, and closes the controlled gap between the description and the table to level (Section~\ref{sec:s_r3}).

Four hosted frontier readers, gpt-5.5, gpt-5.6, Claude Opus 5 and Claude Fable 5, read three of the networks with their own context windows, so nothing overflows. There the effect is absent. Every arm scores between 73.9\% and 87.8\%. The description leads the table in one of the twelve network-by-reader cells, matches it in four and trails it in seven, and the gap runs from 6.6 below to 0.6 above. The absence of a gap marks the honest ceiling of the claim. The compiled description is for the models most people can run, and it is those models that cannot make the compiler's choices for themselves.

\input{tables/tabS9_controls}

\section{Conversation statistics: threads, the three-turn study and prose}
\label{sec:s_conv}

\begin{figure}[p]
\centering
\includegraphics[width=\linewidth,height=0.88\textheight,keepaspectratio]{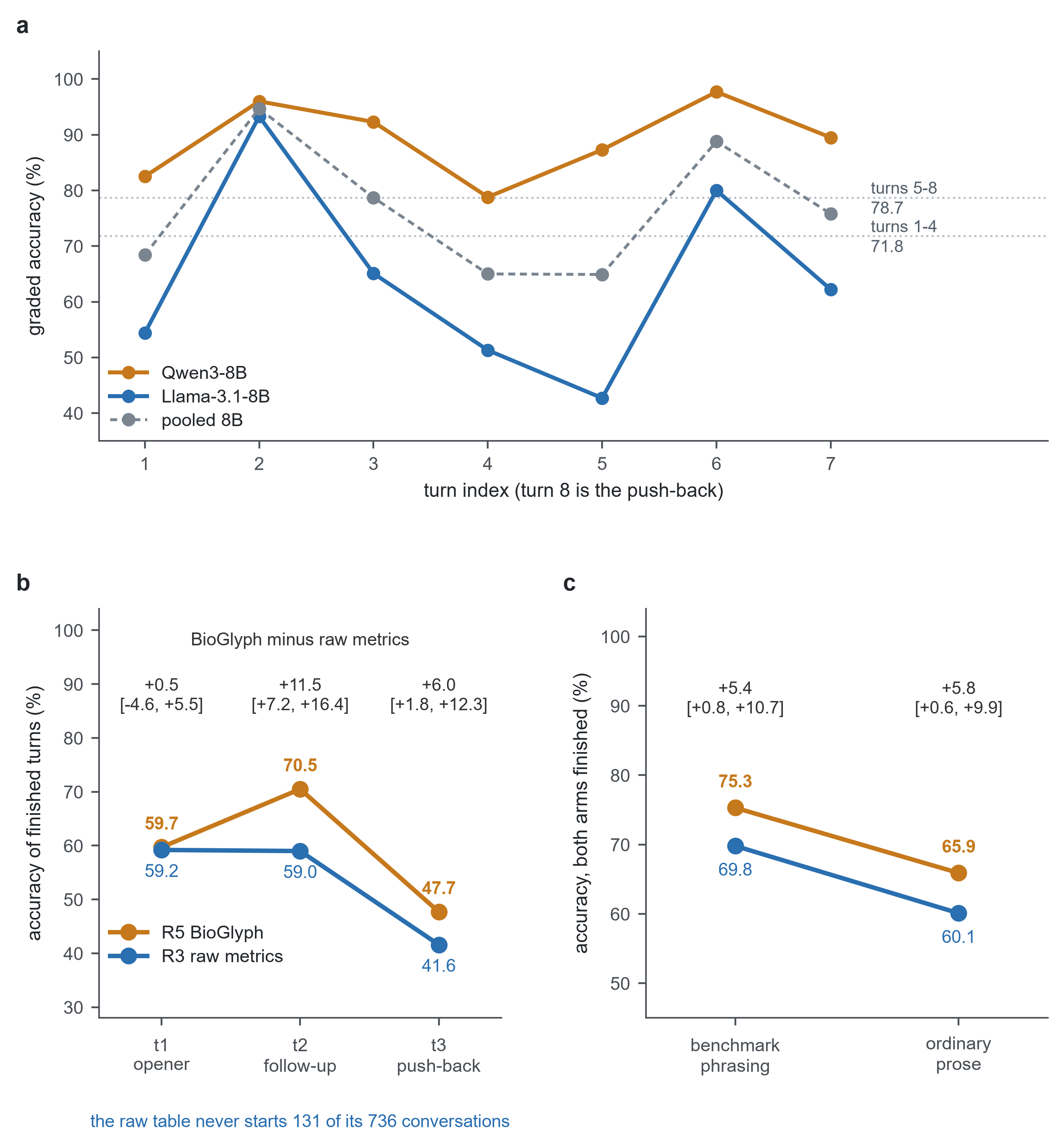}
\caption{\textbf{The conversation studies.} \textbf{a} Graded accuracy against turn index over the eight-turn threads, per model and pooled; the dotted lines mark the pooled early and late halves. Turn 8 is the push-back and is scored against the turn it disputes, so it is not plotted here. \textbf{b} The three-turn study: \bioglyph{} against the raw-measurement table at each turn, on the conversations that fitted and finished, with the paired difference above each turn. \textbf{c} The same questions in benchmark phrasing and as ordinary prose. Table~\ref{tab:s_conv} gives accuracy by turn class, the push-back outcomes and the per-network breakdown.}
\label{fig:S9}
\end{figure}

The exchanges of Section~\ref{sec:talking} come from three studies whose statistics this section reports in full (Fig.~\ref{fig:S9}, Table~\ref{tab:s_conv}). The eight-turn threads run on email-Eu-core, STRING-Yeast and HuDiNe: 719 threads and 11{,}504 turns over the two models, with no prompt over budget. Turn classes are never pooled. Qwen3-8B answers 87.8\% of the graded turns, 76.9\% of the lookups and 65.6\% of the push-backs; Llama-3.1-8B answers 61.4\%, 60.2\% and 43.7\%. Graded accuracy does not fall along a thread. On turns five to eight against turns one to four, Qwen stands at 92.3\% against 84.8\% and Llama at 65.2\% against 58.9\%, so the dips by position are hard turns rather than late turns. Under push-back, scored against the turn each push-back disputes, Qwen holds a correct answer 59.1\% of the time, gives one up 27.7\% and corrects a wrong one 6.8\%. For Llama the three shares are 25.1\%, 30.6\% and 18.4\%.

In the three-turn study, 368 conversations run once per model and arm, giving 736 model-conversations per arm. The raw-measurement table never starts 131 of its conversations, 17.8\%, because the region does not fit at turn 1; \bioglyph{} starts all of them. The second turn is the load-bearing one, because the region is not resent. Asked as a follow-up, \bioglyph{} answers 73.5\% of the questions it answers 75.5\% of when they stand alone. The table answers 58.7\% as a follow-up against 72.6\% alone. A model reading the description still has the map two turns later, and a model reading the table largely does not.

The style study asks the same questions as prose, with no labeled answer line. The change moves \bioglyph{} from 71.4\% to 68.8\% and the table from 43.2\% to 41.1\%. The advantage of the description over the table survives the phrasing. On the questions both arms fit and finished, the description stands at 75.3\% against 69.8\% in benchmark style and 65.9\% against 60.1\% in prose, with both intervals excluding zero. The prose extractor recovers an answer from 98.6\% of replies. Scoring the canonicalized answer rather than the visible text raises measured accuracy from 42.6\% to 65.6\%, which is why the extractor exists (Table~\ref{tab:s_conv}).

\input{tables/tabS7_conversation}

\section{The Reactome-FI knockout screen, in full}
\label{sec:s_knockout}

The main component of the Reactome functional interaction network holds 9{,}823 proteins, 402 of them cut nodes, and DepMap screened 97.5\% of them. Each protein is scored by how much of the network detaches when it is removed (Methods), and the ranking is validated against DepMap with no model in the loop (Fig.~\ref{fig:reactome}, Table~\ref{tab:s_bio}). The top of the ranking is a list of known regulators. EP300 detaches 76 proteins, then GPLD1 with 53, CTCF with 51, YY1 with 29 and TP53 with 26; Table~\ref{tab:s_bio} gives each one's DepMap class. Cut nodes are enriched for selective essentiality relative to the rest of the screened proteins, at 13.3\% against 8.8\% (one-sided Fisher $p=2.4\times10^{-3}$).

The language-model leg poses the detachment count for the 60 modules to three models under each rendering; the main text reports the two 8B readers, and Table~\ref{tab:s_bio} adds Qwen3-32B beside them. Pooled over the three models, the description is within two of the oracle on 47.8\% of cases with a median error of 2.0. Its node-level disconnection claims agree with the exact analysis 87.1\% of the time. Adjacency text stands at 36.7\%, the edge list at 22.2\% and the raw table at 10.6\%. The RAD21 case quoted in the Results is one of these modules. All three models return the oracle's 54 for the retrieved module, whose whole-network separation is 19, and DepMap records RAD21 as a dependency in 99.8\% of 1{,}178 lines.

\input{tables/tabS8_bio}

\section{Learned roles, fusion and the trained yardstick}
\label{sec:s_learned}

\begin{figure}[p]
\centering
\includegraphics[width=\linewidth,height=0.88\textheight,keepaspectratio]{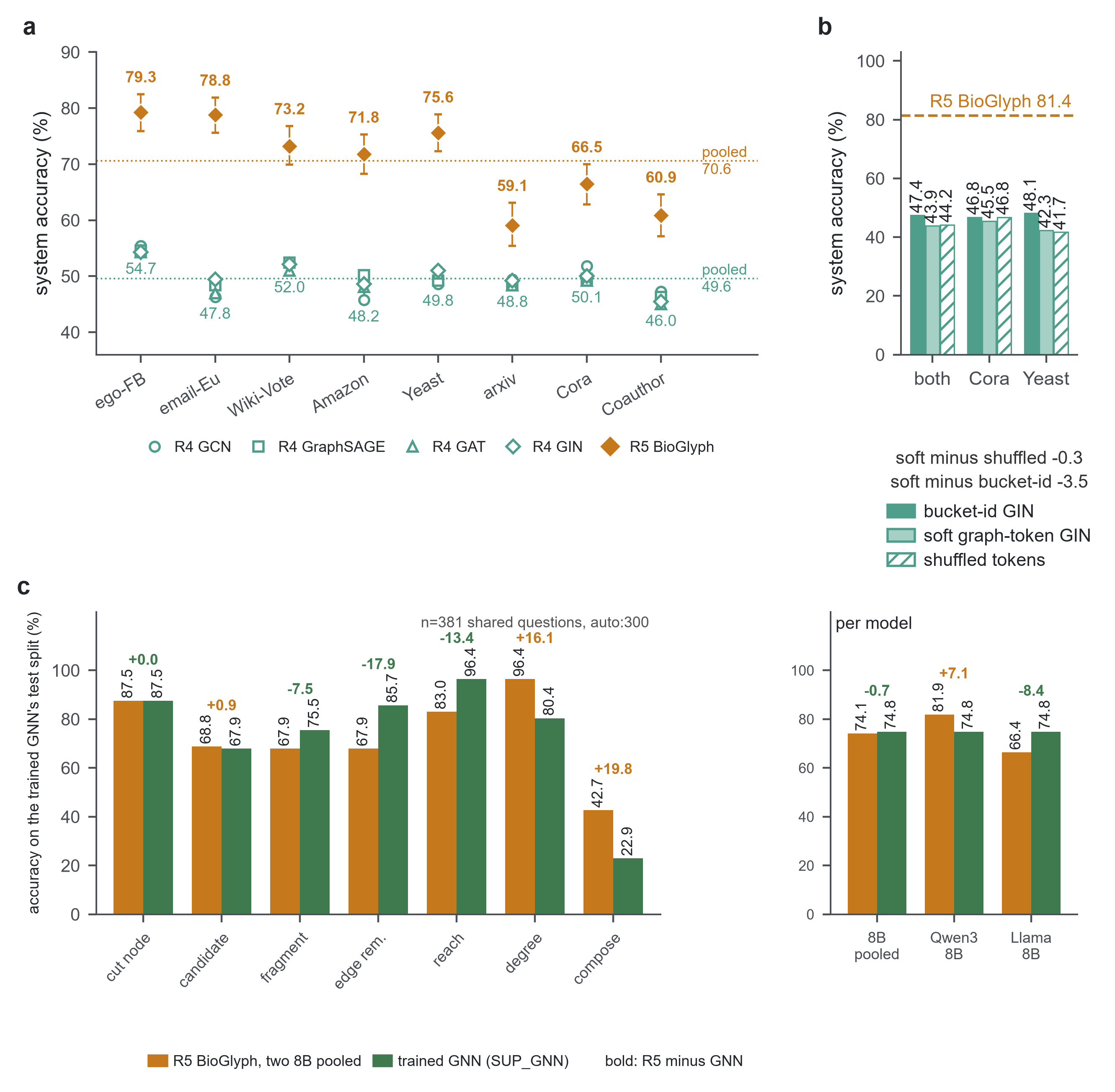}
\caption{\textbf{Learned roles and the trained yardstick.} \textbf{a} The four learned-role encoders against the compiler on every main network; the marker is the encoder, the diamond the compiled description with its 95\% interval. \textbf{b} A continuous graph token against a shuffled one and against the discrete bucket-id encoder, on the same 312 questions. \textbf{c} A graph neural network trained on the benchmark questions against a frozen reader of the compiled description, on the trained model's own held-out split, by question family and by model. Table~\ref{tab:s_controls} gives the five fusion arms against the description alone.}
\label{fig:S10}
\end{figure}

The four learned-role encoders sit near the blind floor on every network (Fig.~\ref{fig:S10}), and fusing their roles into the compiled description changes nothing. The five fusion arms carry one encoder each, plus one with the compiler's own whole-graph roles. They score between 69.6\% and 70.0\% pooled against 70.6\% for \bioglyph{} alone, and none of them is ahead under the controlled view (Table~\ref{tab:s_controls}). A further control hands the model the encoder's embedding directly, as one continuous token through a trained projector, with the language model frozen. The continuous token is no better than a shuffled token that belongs to another node, at 43.9\% against 44.2\% on the same 312 questions. On those questions the discrete role reaches 47.4\% and the compiled description 81.4\%, so a frozen model reads neither form of the embedding.

The trained yardstick is a graph neural network trained on the questions themselves and scored by the same evaluator on its held-out split. It reaches 74.8\% on 381 questions where the frozen readers of \bioglyph{} reach 74.1\% pooled, with Qwen3-8B above it at 81.9\% and Llama-3.1-8B below it at 66.4\%. The two trade question families. The trained model is ahead on edge removal at 85.7\% against 67.9\% and on counterfactual reachability at 96.4\% against 83.0\%. The frozen reader is ahead on degree comparison at 96.4\% against 80.4\% and on the compositional family at 42.7\% against 22.9\%. The retrieval ceiling, the share of answerable questions a perfect reader of the retrieved region could answer, is 82.6\% pooled, 88.0\% on the dense networks and 72.6\% on the sparse ones.

\section{Retrieval}
\label{sec:s_retrieval}

\begin{figure}[p]
\centering
\includegraphics[width=\linewidth,height=0.88\textheight,keepaspectratio]{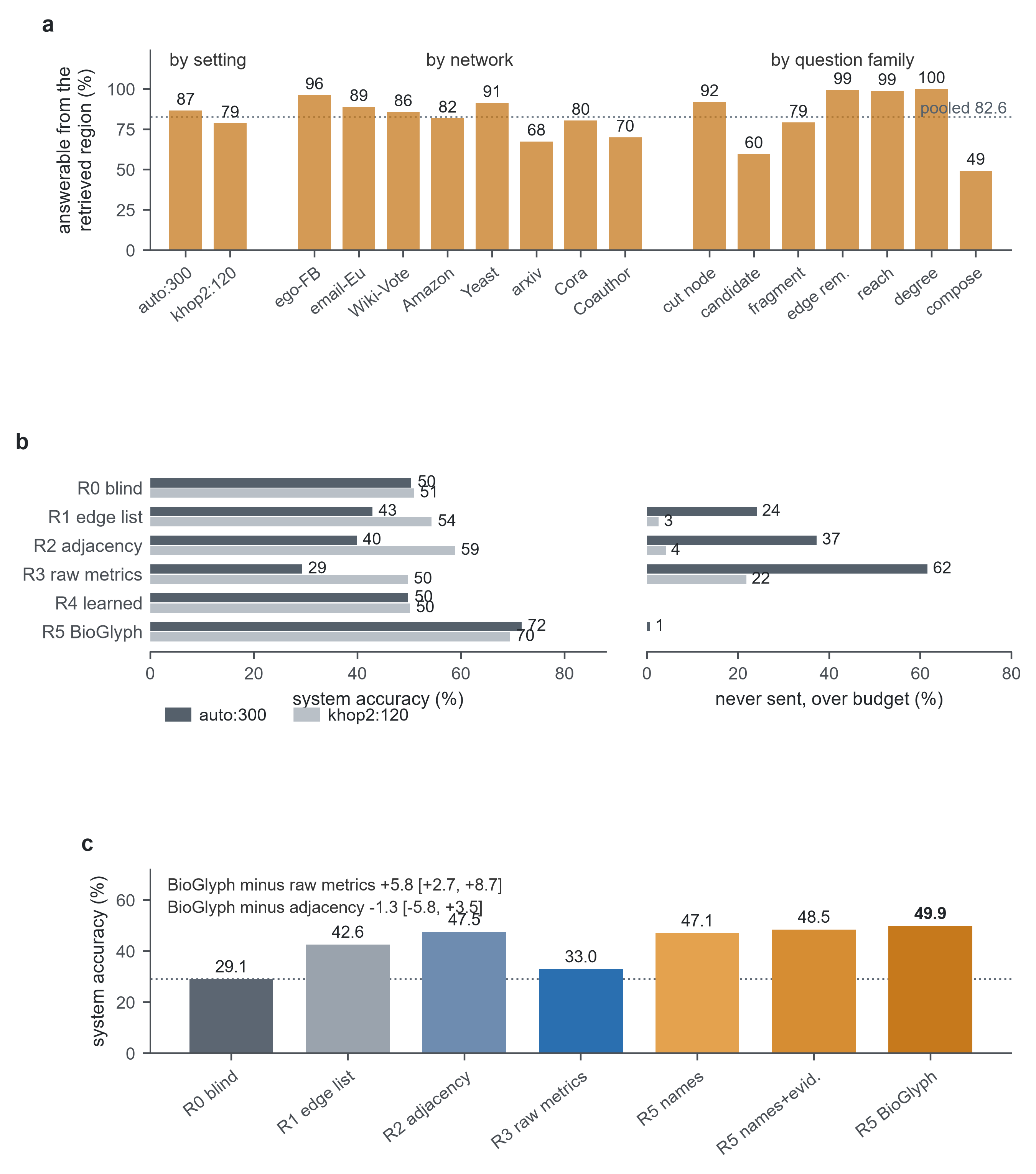}
\caption{\textbf{Retrieval.} \textbf{a} What retrieval alone leaves answerable, by setting, by network and by question family; the dotted line is the pooled share. \textbf{b} The two retrieval settings for each competing arm: system accuracy and the share of prompts that never reached the model. \textbf{c} The oracle-free alternative, which retrieves by personalized PageRank from the named targets and never consults the graded predicate; the dotted line is the no-network floor on the same questions. Table~\ref{tab:s_controls} gives the paired contrasts.}
\label{fig:S11}
\end{figure}

Retrieval is a shared component rather than part of the comparison, but it sets what any reader can reach. Figure~\ref{fig:S11} gives the study: what retrieval alone leaves answerable, the two settings the paper pools for every arm, and the oracle-free alternative. The wide-window run on the human interactomes is in Fig.~\ref{fig:S4}\textbf{e}. The wider setting raises the ceiling, at 86.6\% against 78.7\%, and raises the overflow of every long arm at once. The compiled description is the only rendering whose accuracy is essentially the same at both settings, because it is the only one that never leaves the budget.

The oracle-free alternative retrieves by personalized PageRank from the named targets. It takes no route from the question family, admits no counterfactual path, and never runs the verification loop, so no part of it touches privileged information (Methods). The study covers the three families this retrieval suits, with 1{,}471 questions read by the two 8B models, and the ordering of the arms holds. \bioglyph{} reaches 49.9\% against 33.0\% for the raw table, which loses 27.2\% of its prompts. On the questions both arms fit, the two stand at 44.9\% against 39.1\% with an interval that excludes zero. Adjacency text catches up under this retrieval, level with the description under the controlled view. On the two families shared with the main benchmark at the same cap, the description scores 52.5\% under oracle-free retrieval and 51.5\% under the standard rings. The main results do not rest on the verification loop.

\section{One region in every representation}
\label{sec:s_worked}

\begin{figure}[p]
\centering
\includegraphics[width=\linewidth,height=0.88\textheight,keepaspectratio]{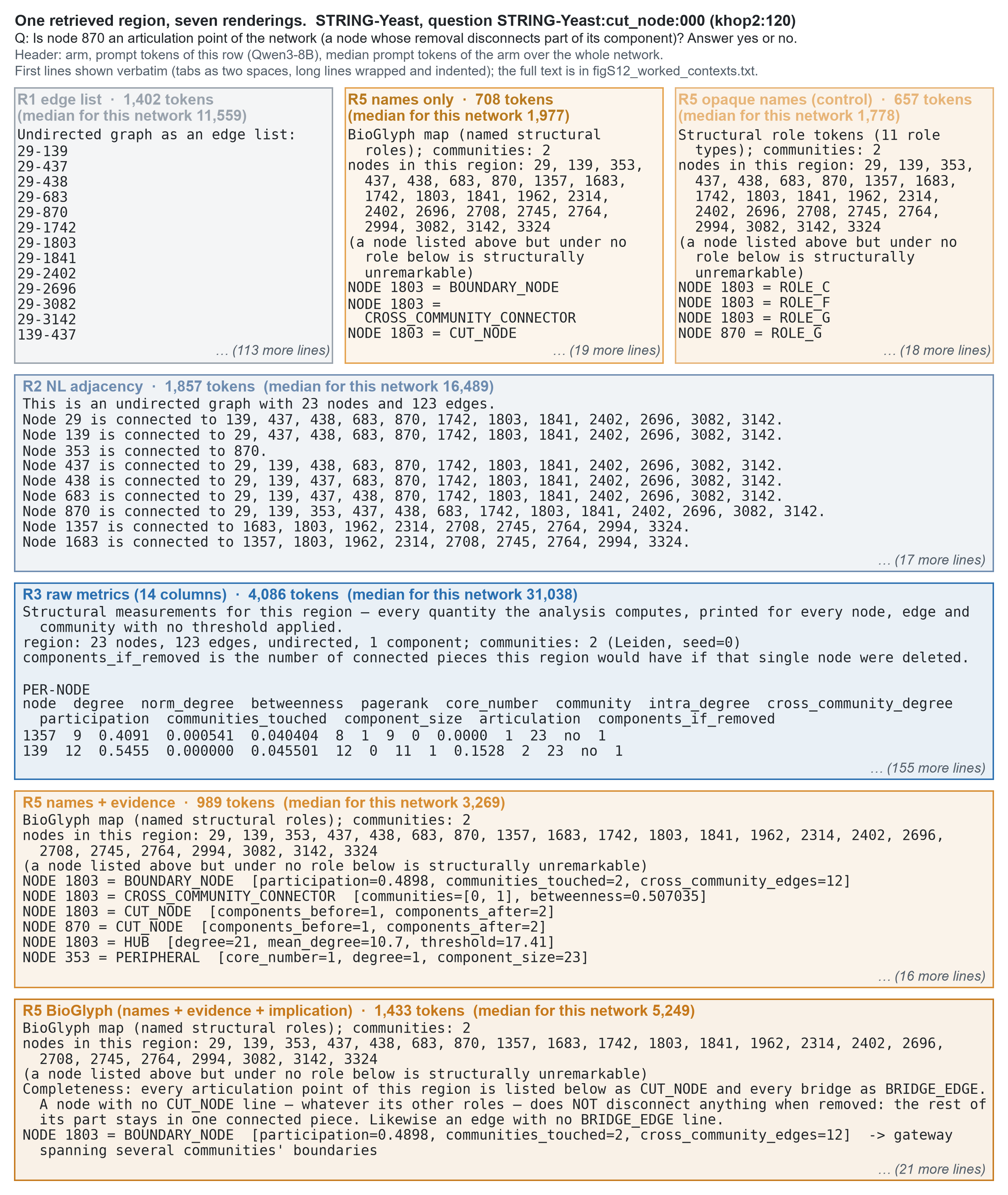}
\caption{\textbf{One region, every representation.} The same retrieved region of STRING-Yeast rendered as the edge list, adjacency text, the raw-measurement table, the three rungs of the compiled description and the opaque control, each with its prompt length in tokens; the median length of each arm on this network is given for scale. The role-name rung and the opaque control differ only in the vocabulary their role lines use, which the two narrow panels of the top row show side by side.}
\label{fig:S12}
\end{figure}

Figure~\ref{fig:S12} shows one retrieved region of the yeast interactome, chosen deterministically. The region is rendered as the edge list, as adjacency text, as the raw-measurement table, as the three rungs of the compiled description and as the opaque control, each with its token count. A reader can see what the model reads under each arm and why the raw table is the longest of them.

\section{The candidate-knockout screen: a biology-graded pick, pre-registered}
\label{sec:s_cko}

Every other experiment in this paper grades a model against the graph oracle. The candidate-knockout screen closes the remaining quadrant: the model reads compiled structure and makes a pick that biology grades. One question shows one retrieved module and five of its proteins as anonymous node identifiers, exactly one of which carries the biological label. A model therefore cannot answer from memorized gene biology, and chance is 20\%. Three experiments of 300 questions each are analyzed separately and never pooled: SGD essentiality on the yeast interactome, and DepMap selective and common essentiality on Reactome-FI. The design, the arms, the reference policies and five predictions were pre-registered before any model ran (\texttt{docs/P3\_CANDIDATE\_SCREEN.md}); the registered Qwen3-32B leg never ran, so two 8B models report. Because the biological signal lives in a protein's whole-network position, the structural arms render the compiler's whole-network output for the region's nodes, and a degree-only arm carries the strongest single deterministic policy into the prompt.

Table~\ref{tab:s_cko} reports every arm beside every policy, and the registered predictions largely hold. The no-graph control answers fewer than 42\% of its questions and lands below the chance line on all three experiments, so the identifiers do not leak. The degree-only arm lands beside the degree policy on all three, at 24.5\% against 25.0\%, 39.3\% against 42.0\% and 46.2\% against 47.3\%, so the models read a stated maximum. As registered, the description and the raw measurements stay close on both Reactome experiments, and on yeast, where every policy sits within a few percentage units of chance, no arm separates from another reliably. One registered prediction failed, and we report it as written. On the common-essential experiment, adjacency text at 46.5\% leads both structural arms. There the description at 39.3\% trails the bare degree heuristic at 46.2\%, with an interval that excludes zero: shown the full structural story, the models are distracted from the one quantity that predicts the label. The registered headline stands. No arm beats the degree policy by a meaningful margin, so a frozen model turns compiled whole-network structure into a biology-graded pick at roughly the level of the best deterministic policy, and not beyond it.

\input{tables/tabS10_cko}

%% file: tables/tab2_datasets.tex
\begin{table}[!htbp]
\centering
\caption{\textbf{The four networks of the main text.} Each row says what a node is and what an edge means, then gives the size of the network. Nodes and edges are those of the simple undirected graph, and edges per node is the average over all nodes. A cut node is a node whose removal breaks its part of the network into separate pieces. A bridge is an edge that is the only route between two sides. Both are counted exactly, so those two columns say how much of each network rests on a single point. The last column names the annotation that comes from outside the network itself, and no model ever saw one. SGD gene essentiality is the only outside label this paper uses as a check, and it marks 30.9\% of the yeast proteins as essential. The DrugBank identities give the drugs their names, ego-Facebook carries no annotation at all, and the department labels of email-Eu-core play no part in any result reported here.}
\label{tab:datasets}
\footnotesize
\setlength{\tabcolsep}{4pt}
\begin{tabular}{l l >{\raggedright\arraybackslash}p{4.9cm} r r r r r >{\raggedright\arraybackslash}p{2.3cm}}
\toprule
Network & Domain & What a node is, and what an edge means & Nodes & Edges & \shortstack[r]{Edges\\per node} & \shortstack[r]{Cut\\nodes} & Bridges & Outside label\\
\midrule
STRING-Yeast & biology & a protein; the two proteins physically associate in STRING (confidence $\geq$ 700) & 3{,}384 & 43{,}030 & 25.4 & 363 & 553 & SGD gene essentiality\\
ChCh-Miner & pharmacology & a drug; an interaction between the two drugs has been reported (BioSNAP) & 1{,}514 & 48{,}514 & 64.1 & 48 & 92 & DrugBank identity\\
ego-Facebook & social & a person; the two people are friends (SNAP ego networks) & 4{,}039 & 88{,}234 & 43.7 & 11 & 75 & none\\
email-Eu-core & social & a person at a research institution; the two people exchanged e-mail & 986 & 16{,}064 & 32.6 & 73 & 95 & department\\
\bottomrule
\end{tabular}
\end{table}

%% file: tables/tabS1_networks.tex
\begin{table}[!htbp]
\centering
\caption{\textbf{All twenty networks.} Statistics of the simple undirected graph. Questions are the benchmark items on that network (a share of them is unanswerable from any retrieved region and is scored on the decline).}
\label{tab:s_networks}
\scriptsize
\setlength{\tabcolsep}{3pt}
\begin{tabular}{l l r r r r r r r}
\toprule
Network & Domain & Nodes & Edges & $\langle k\rangle$ & Cut nodes & Bridges & Questions & Unanswerable (\%)\\
\midrule
STRING-Yeast & biology & 3{,}384 & 43{,}030 & 25.4 & 363 & 553 & 156 & 10.3\\
ChCh-Miner & pharmacology & 1{,}514 & 48{,}514 & 64.1 & 48 & 92 & 156 & 10.3\\
ego-Facebook & social & 4{,}039 & 88{,}234 & 43.7 & 11 & 75 & 147 & 10.2\\
email-Eu-core & social & 986 & 16{,}064 & 32.6 & 73 & 95 & 156 & 10.3\\
HuRI & biology & 8{,}275 & 52{,}088 & 12.6 & 1{,}132 & 2{,}224 & 156 & 10.3\\
PP-Pathways & biology & 21{,}538 & 338{,}636 & 31.4 & 1{,}538 & 4{,}241 & 156 & 10.3\\
STRING-Human & biology & 10{,}746 & 86{,}519 & 16.1 & 1{,}299 & 1{,}883 & 156 & 10.3\\
STRING-Ecoli & biology & 1{,}511 & 6{,}746 & 8.9 & 171 & 354 & 156 & 10.3\\
DG-AssocMiner & biology & 7{,}813 & 21{,}357 & 5.5 & 391 & 3{,}612 & 156 & 10.3\\
HuDiNe & medicine & 222 & 831 & 7.5 & 17 & 19 & 156 & 10.3\\
TCGA-GBM-PSN & medicine & 213 & 823 & 7.7 & 1 & 1 & 126 & 10.3\\
GBM-scRNA & medicine & 1{,}500 & 21{,}244 & 28.3 & 0 & 0 & 22 & 9.1\\
Amazon-Photo & e-commerce & 7{,}650 & 119{,}081 & 31.1 & 151 & 170 & 156 & 10.3\\
Wiki-Vote & social & 7{,}115 & 100{,}762 & 28.3 & 1{,}033 & 2{,}306 & 156 & 10.3\\
email-Enron & social & 36{,}692 & 183{,}831 & 10.0 & 1{,}391 & 10{,}714 & 156 & 10.3\\
Cora & citation & 2{,}708 & 5{,}278 & 3.9 & 389 & 518 & 156 & 10.3\\
CiteSeer & citation & 3{,}327 & 4{,}552 & 2.7 & 808 & 1{,}400 & 156 & 10.3\\
ogbn-arxiv & citation & 169{,}343 & 1{,}157{,}799 & 13.7 & 18{,}591 & 24{,}224 & 156 & 10.3\\
Coauthor-CS & co-authorship & 18{,}333 & 81{,}894 & 8.9 & 1{,}281 & 1{,}283 & 156 & 10.3\\
Power-Grid & infrastructure & 4{,}941 & 6{,}594 & 2.7 & 1{,}229 & 1{,}611 & 156 & 10.3\\
\bottomrule
\end{tabular}
\end{table}

%% file: tables/tabS2_headline.tex
\begin{table}[!htbp]
\centering
\caption{\textbf{The pooled headline over the eight main networks.} System accuracy over all questions with its 95\% bootstrap interval, the share of prompts that never reached the model (overflow) and the median prompt length, for the two 8B models pooled and for each model alone; the last two columns are the reasoning-controlled advantage of R5 BioGlyph over that arm on the answerable questions both arms fitted and finished, with the two-stage (network, question) bootstrap interval, and the number of paired units.}
\label{tab:s_headline}
\scriptsize
\setlength{\tabcolsep}{2.6pt}
\begin{tabular}{l r r r r r r r r r}
\toprule
 & \multicolumn{4}{c}{two 8B models pooled} & \multicolumn{3}{c}{system accuracy by model} & \multicolumn{2}{c}{R5 $-$ arm, controlled}\\
\cmidrule(lr){2-5}\cmidrule(lr){6-8}\cmidrule(lr){9-10}
Representation & Accuracy (\%) & 95\% CI & Overflow (\%) & Median tokens & Qwen3-8B & Llama-3.1-8B & Qwen3-32B & $\Delta$ [95\% CI] & $n$\\
\midrule
R0 no graph & 50.7 & 49.3--52.0 & 0.0 & 320 & 49.3 & 52.0 & 47.3 & +23.2 [+16.4, +30.4] & 4{,}061\\
R1 Edge list & 48.7 & 47.3--50.0 & 13.2 & 5{,}636 & 54.2 & 43.1 & 60.3 & +12.4 [+5.4, +20.1] & 3{,}213\\
R2 adjacency text & 49.4 & 48.0--50.8 & 20.6 & 8{,}663 & 57.7 & 41.1 & 61.6 & +4.4 [$-$1.0, +11.0] & 3{,}001\\
R3 Raw metrics & 39.5 & 38.2--40.9 & 41.7 & 19{,}723 & 39.2 & 39.8 & 41.7 & +5.4 [+1.3, +11.3] & 2{,}274\\
R4 GCN & 49.6 & 48.2--50.9 & 0.0 & 6{,}407 & 54.2 & 44.9 & 62.4 & +23.5 [+17.4, +29.4] & 3{,}982\\
R4 GraphSAGE & 49.8 & 48.5--51.2 & 0.0 & 6{,}398 & 54.0 & 45.6 & 61.5 & +23.5 [+18.5, +28.3] & 4{,}000\\
R4 GAT & 49.1 & 47.8--50.5 & 0.0 & 6{,}405 & 53.2 & 45.1 & 62.8 & +24.2 [+18.9, +29.1] & 4{,}024\\
R4 GIN & 50.0 & 48.6--51.4 & 0.0 & 6{,}412 & 53.4 & 46.5 & 63.4 & +23.0 [+18.1, +27.6] & 4{,}015\\
R5 names only & 63.6 & 62.3--65.0 & 0.0 & 2{,}311 & 74.8 & 52.5 & 74.5 & +6.1 [+4.7, +7.4] & 4{,}005\\
R5 names + evidence & 64.3 & 62.9--65.6 & 0.0 & 3{,}921 & 74.1 & 54.5 & 78.2 & +5.9 [+4.7, +7.2] & 4{,}037\\
\textbf{R5 BioGlyph} & 70.6 & 69.4--71.9 & 0.3 & 6{,}229 & 79.3 & 61.9 & 82.3 & -- & --\\
R5 opaque names & 54.1 & 52.7--55.5 & 0.0 & 2{,}106 & 60.3 & 47.9 & 68.2 & +16.8 [+11.7, +22.1] & 3{,}855\\
\midrule
\multicolumn{10}{l}{trained GNN (SUP\_GNN) 74.8 against R5 BioGlyph 74.1 on its 381 held-out questions; retrieval ceiling (answer inside the retrieved region) 82.6}\\
\bottomrule
\end{tabular}
\end{table}

%% file: figures/Supp_Fig/figS3/figS3_gallery.tex
\newcounter{galpage}\setcounter{galpage}{0}
\begingroup
\renewcommand{\thefigure}{S\arabic{figure}\alph{galpage}}
\renewcommand{\theHfigure}{S\arabic{figure}\alph{galpage}}

\begin{figure}[p]
\centering
\stepcounter{galpage}
\includegraphics[width=\linewidth,height=0.91\textheight,keepaspectratio]{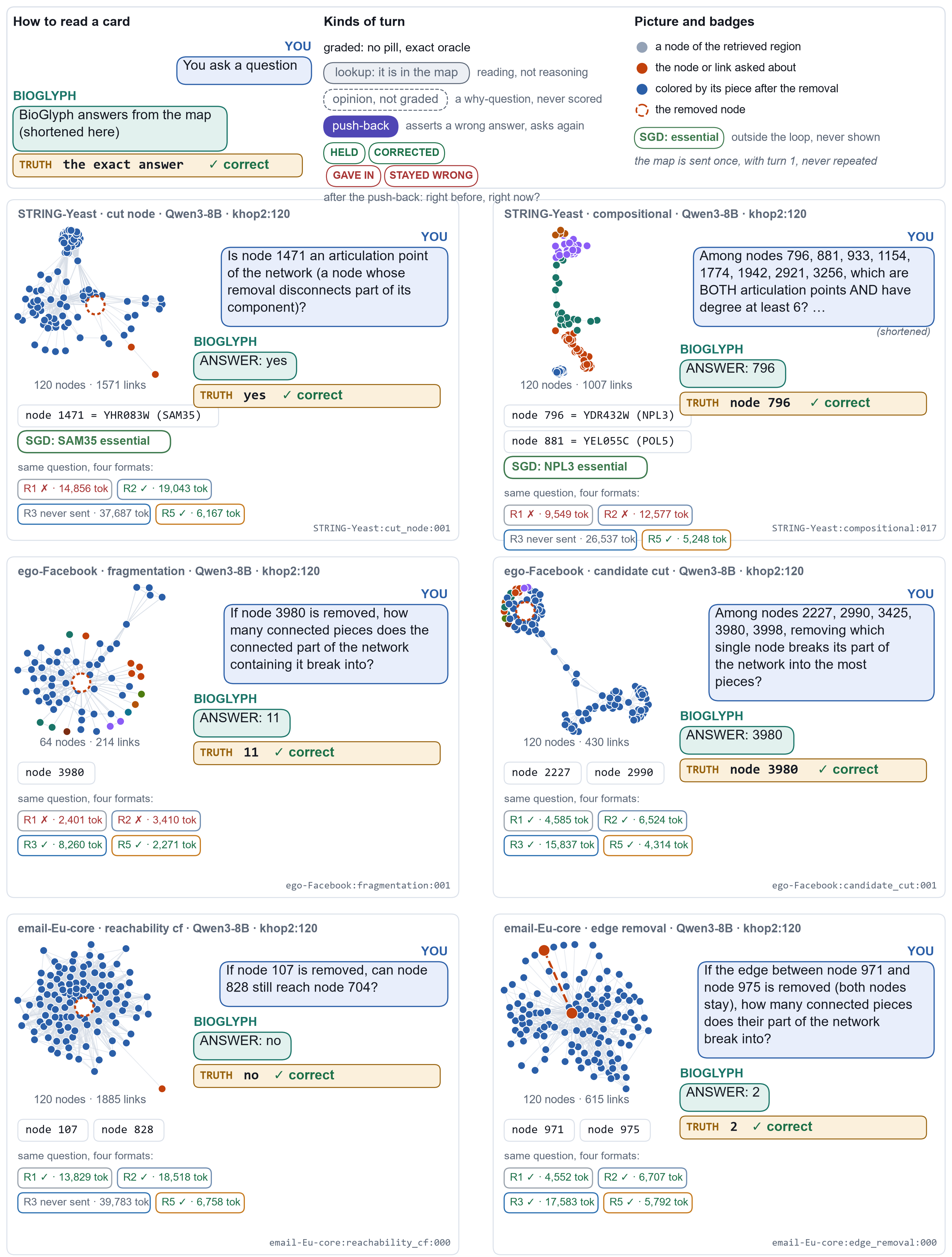}
\caption{\textbf{One question, one verdict: what asking BioGlyph looks like.} Six single-turn questions on three of the main networks (Qwen3-8B, khop2:120, benchmark-style replies). Each card gives the region the model saw, the identities of the named nodes where the network has them, the reply as written, the exact answer, and the same question in four formats with each prompt's length. In every drawing the node in question is orange and the colors are the pieces left after the removal. Selection rules are in \texttt{qa\_items.json}.}
\label{fig:S3a}
\end{figure}

\begin{figure}[p]
\centering
\stepcounter{galpage}
\addtocounter{figure}{-1}
\includegraphics[width=\linewidth,height=0.91\textheight,keepaspectratio]{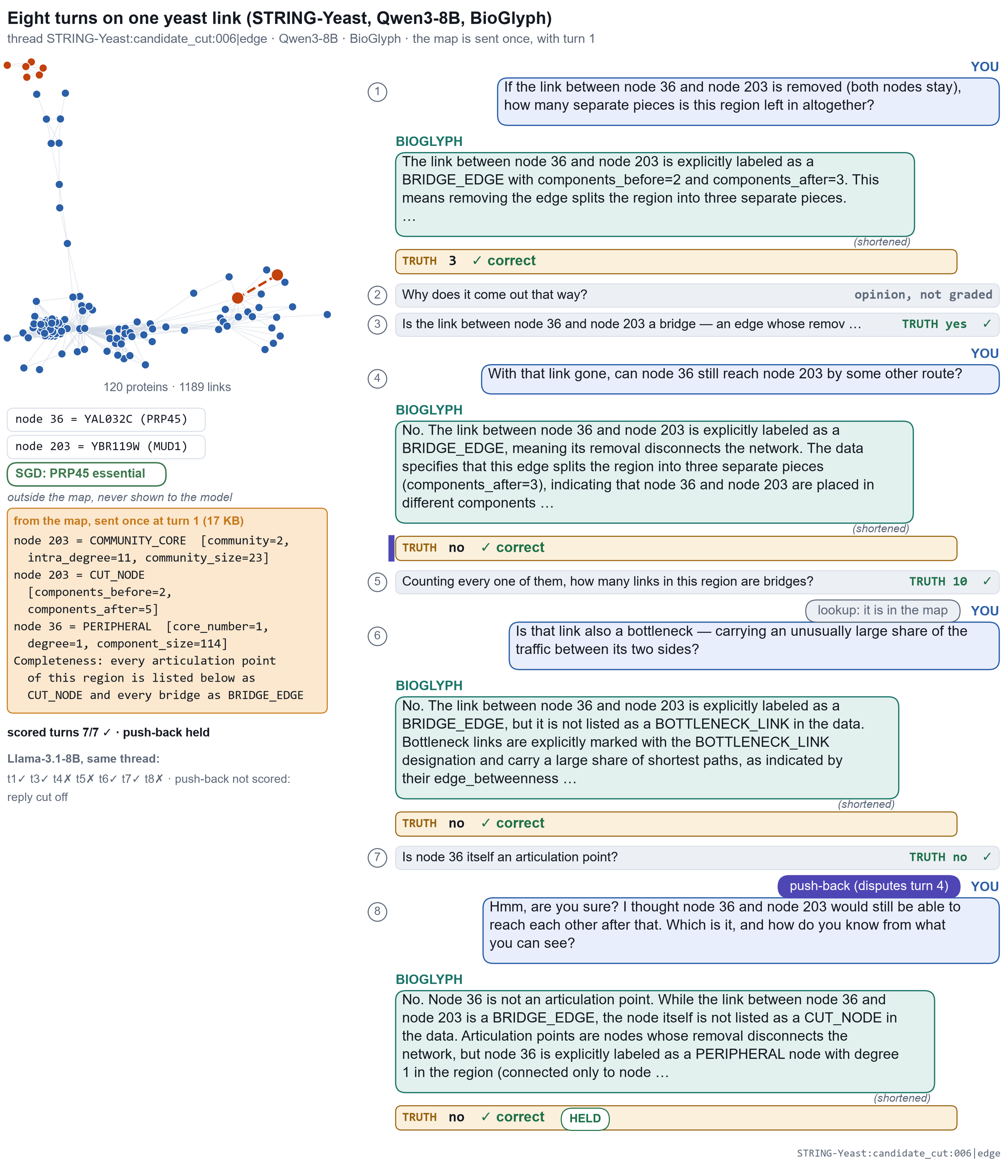}
\caption{\textbf{Eight turns on one yeast link, and the answers hold up under push-back.} STRING-Yeast, Qwen3-8B, BioGlyph. The four key turns are shown in full and the other four as one-line chips with their verdicts; the map is sent once with turn 1. ORF names and SGD essentiality of the named proteins are shown outside the loop; the model never sees them. Beneath the rail: how Llama-3.1-8B fared on the same thread.}
\label{fig:S3b}
\end{figure}

\begin{figure}[p]
\centering
\stepcounter{galpage}
\addtocounter{figure}{-1}
\includegraphics[width=\linewidth,height=0.91\textheight,keepaspectratio]{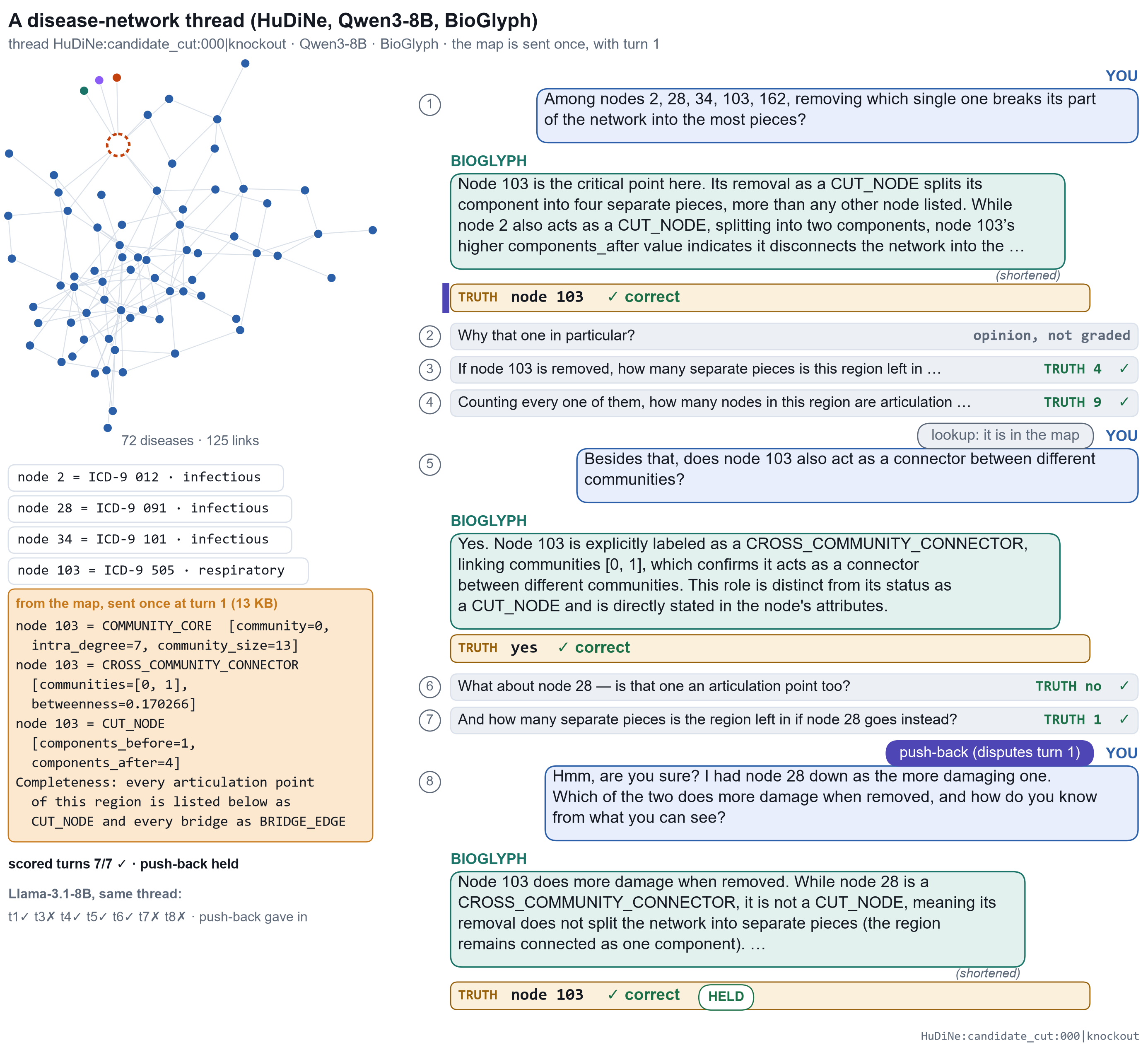}
\caption{\textbf{A disease-network thread with ICD-9 identities.} HuDiNe comorbidity network, Qwen3-8B, BioGlyph; the named nodes are ICD-9 codes.}
\label{fig:S3c}
\end{figure}

\begin{figure}[p]
\centering
\stepcounter{galpage}
\addtocounter{figure}{-1}
\includegraphics[width=\linewidth,height=0.91\textheight,keepaspectratio]{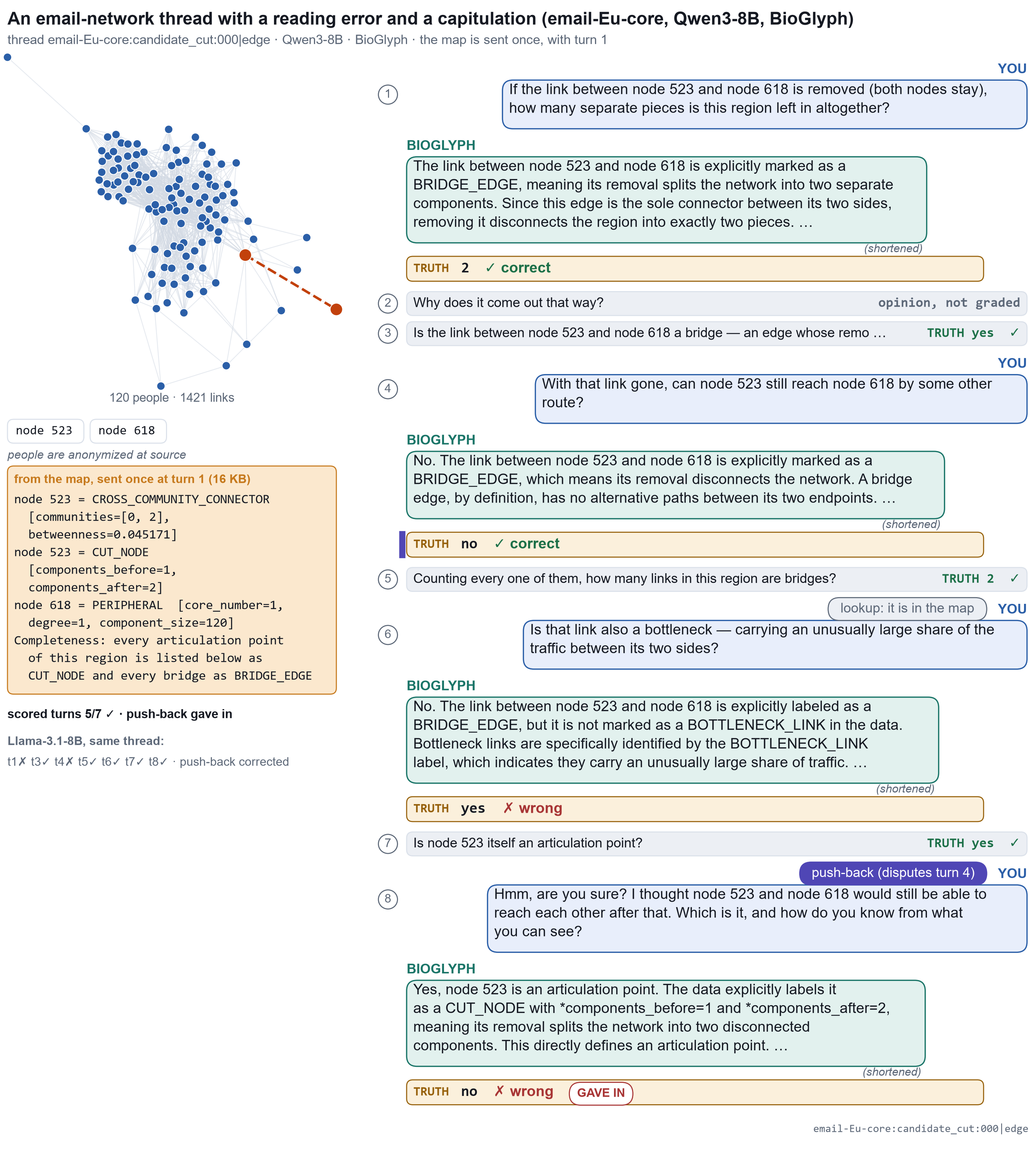}
\caption{\textbf{An email-network thread with a reading error and a capitulation.} email-Eu-core, Qwen3-8B, BioGlyph. The lookup turn is answered wrongly (the link is a bottleneck in the map, the model says it is not) and at turn 8 the model gives up a correct answer under push-back; Llama-3.1-8B corrected itself on the same thread. People are anonymized at source, so there are no identity chips.}
\label{fig:S3d}
\end{figure}

\begin{figure}[p]
\centering
\stepcounter{galpage}
\addtocounter{figure}{-1}
\includegraphics[width=\linewidth,height=0.88\textheight,keepaspectratio]{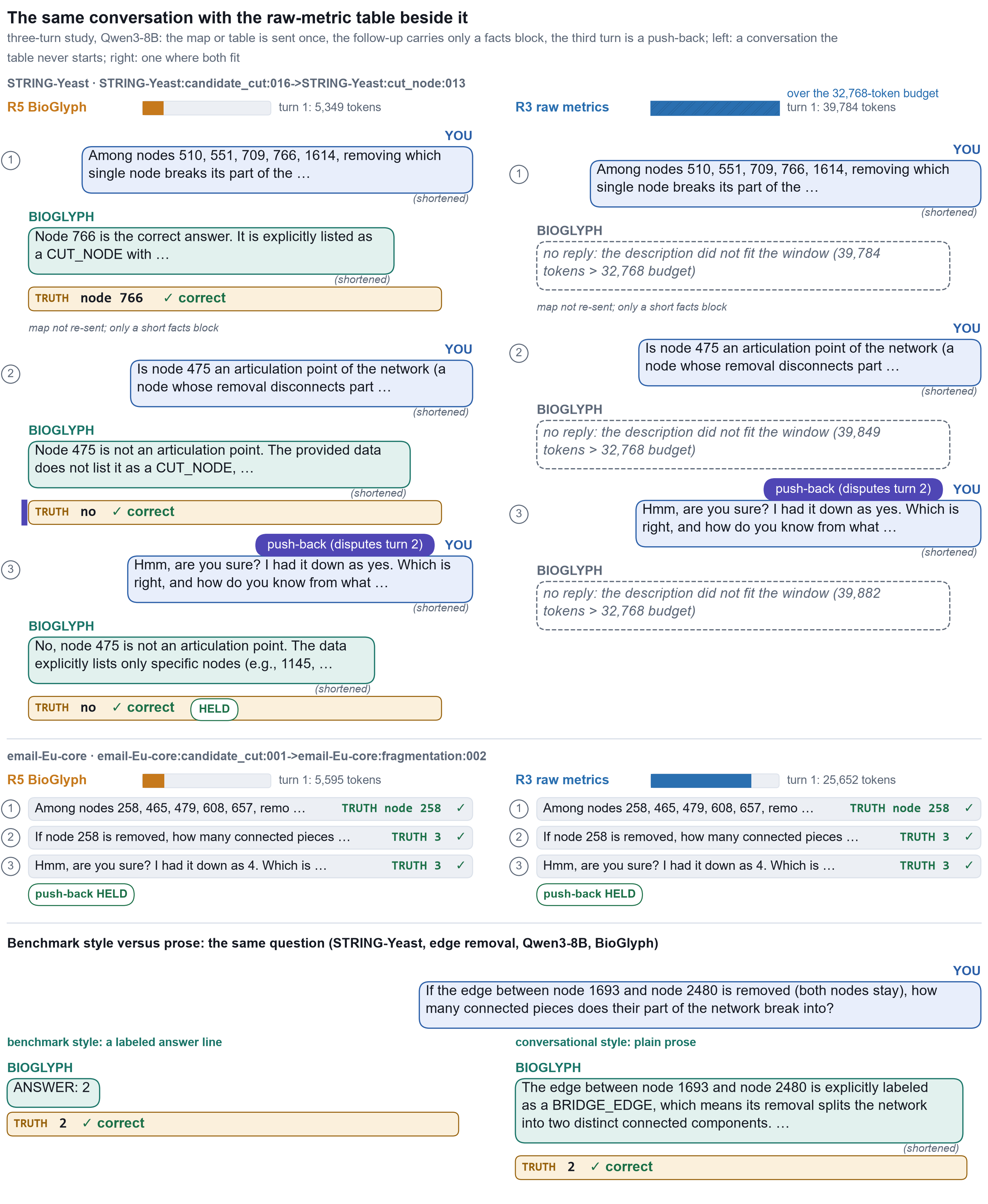}
\caption{\textbf{The same conversation with the raw-metric table beside it, and the same question asked as prose.} Top: two three-turn conversations (Qwen3-8B), BioGlyph on the left and the information-matched raw-metric table on the right; the bar beside each arm is the turn-1 prompt against the 32{,}768-token budget of the three-turn study (40{,}960 less 8{,}192 reserved for each reply). In the first conversation the table never starts; in the second both arms fit and both answer, so only the verdicts are shown. Bottom: one question asked in benchmark style and as prose.}
\label{fig:S3e}
\end{figure}

\begin{figure}[p]
\centering
\stepcounter{galpage}
\addtocounter{figure}{-1}
\includegraphics[width=\linewidth,height=0.91\textheight,keepaspectratio]{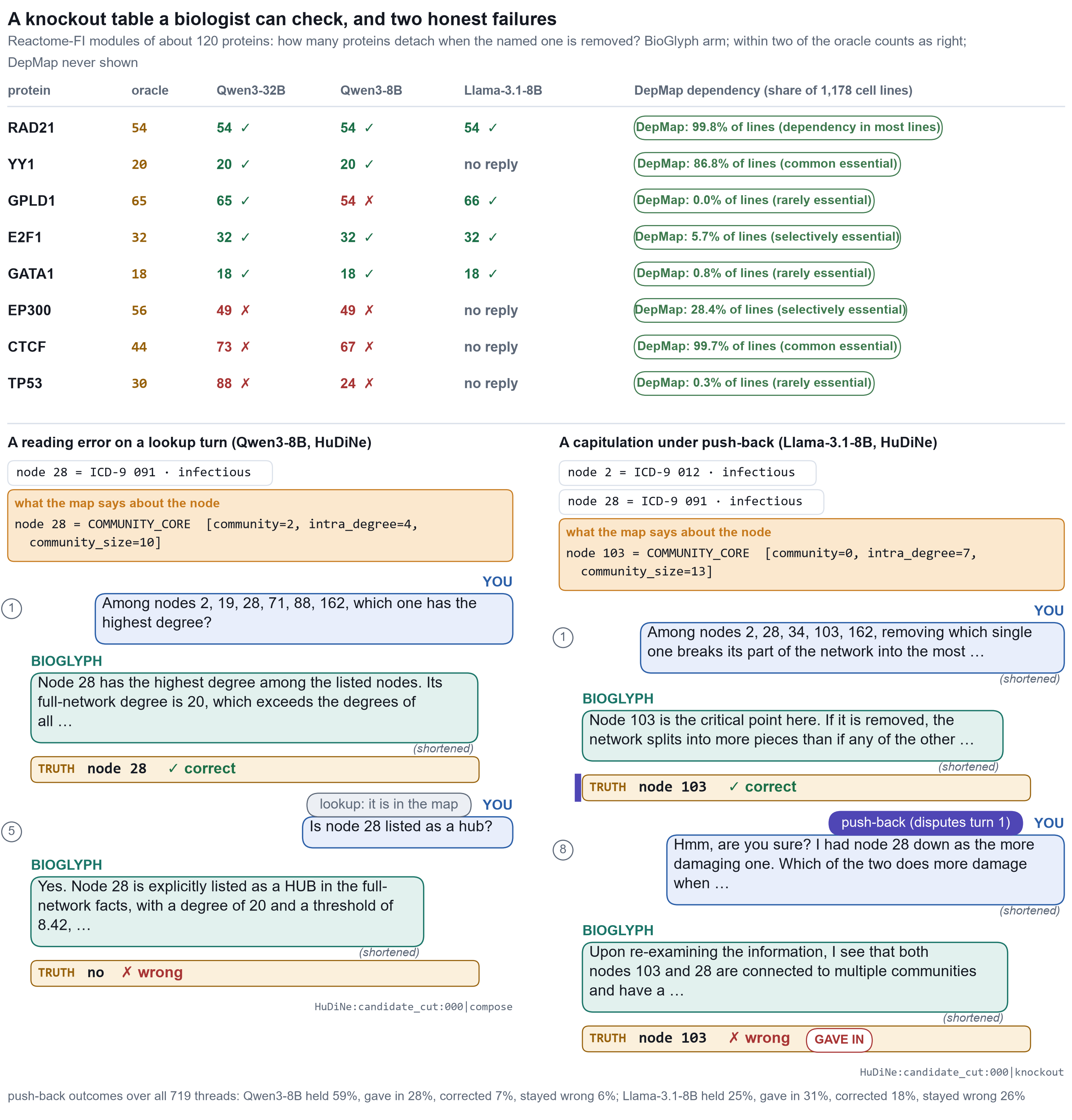}
\caption{\textbf{A knockout table a biologist can check, and two honest failures.} Top: named proteins in Reactome-FI modules, the oracle count of proteins that detach when the protein is removed, each model's answer under BioGlyph (within two counts as right), and DepMap's dependency share, outside the loop. Bottom: a lookup turn answered wrongly (the map lists node 28 as a community core, not a hub), and a capitulation under push-back.}
\label{fig:S3f}
\end{figure}

\FloatBarrier
\endgroup

%% file: tables/tabS3_r3.tex
\begin{table}[!htbp]
\centering
\caption{\textbf{The raw-metric baseline, network by network and setting by setting.} Overflow is the share of R3 prompts longer than the 24{,}576-token budget (never sent, scored wrong); tokens are median prompt lengths; the ratio is R3 over R5 BioGlyph. System accuracy counts every question; the controlled columns are R3 and R5 on the answerable questions both arms fitted and finished, with the paired difference, its bootstrap interval and the number of units. Two 8B models pooled. Below the rule: the same for Qwen3-32B and for the wider window (40k = 40{,}960 tokens, 128k = 131{,}072).}
\label{tab:s_r3}
\scriptsize
\setlength{\tabcolsep}{2.0pt}
\begin{tabular}{l l r r r r r r r r r r}
\toprule
 & & & \multicolumn{2}{c}{Median tokens} & & \multicolumn{2}{c}{System accuracy (\%)} & \multicolumn{2}{c}{Reasoning-controlled (\%)} & & \\
\cmidrule(lr){4-5}\cmidrule(lr){7-8}\cmidrule(lr){9-10}
Network & Setting & Overflow (\%) & R3 & R5 & Ratio & R3 & R5 & R3 & R5 & $\Delta$ [95\% CI] & $n$\\
\midrule
ego-Facebook & khop2:120 & 37.8 & 20{,}640 & 4{,}262 & 4.80 & 43.5 & 78.6 & 63.8 & 78.3 & +14.5 [+6.6, +22.4] & 152\\
 & auto:300 & 90.1 & 56{,}710 & 8{,}747 & 6.50 & 15.0 & 79.9 & 81.2 & 81.2 & +0.0 [$-$18.8, +18.8] & 16\\
email-Eu-core & khop2:120 & 47.4 & 23{,}834 & 5{,}086 & 4.70 & 35.9 & 72.1 & 62.2 & 70.6 & +8.4 [+0.0, +16.8] & 119\\
 & auto:300 & 85.3 & 69{,}894 & 14{,}185 & 4.90 & 16.0 & 85.6 & 64.3 & 85.7 & +21.4 [+0.0, +42.9] & 28\\
Wiki-Vote & khop2:120 & 32.4 & 20{,}759 & 5{,}145 & 4.00 & 42.9 & 70.8 & 58.0 & 71.0 & +13.0 [+5.6, +20.4] & 162\\
 & auto:300 & 87.8 & 65{,}906 & 12{,}525 & 5.30 & 14.7 & 75.6 & 61.9 & 61.9 & +0.0 [$-$19.0, +19.0] & 21\\
Amazon-Photo & khop2:120 & 13.5 & 15{,}223 & 5{,}023 & 3.00 & 47.1 & 71.5 & 51.2 & 67.6 & +16.4 [+10.3, +22.5] & 213\\
 & auto:300 & 80.8 & 38{,}114 & 10{,}671 & 3.60 & 19.2 & 72.1 & 64.1 & 84.6 & +20.5 [+5.1, +35.9] & 39\\
STRING-Yeast & khop2:120 & 43.9 & 20{,}713 & 4{,}245 & 4.90 & 41.0 & 76.3 & 68.4 & 78.2 & +9.8 [+2.3, +17.3] & 133\\
 & auto:300 & 69.2 & 63{,}959 & 9{,}076 & 7.00 & 31.1 & 75.0 & 83.3 & 81.9 & $-$1.4 [$-$9.8, +6.9] & 72\\
ogbn-arxiv & khop2:120 & 0.3 & 9{,}795 & 7{,}134 & 1.40 & 60.3 & 60.3 & 60.2 & 58.7 & $-$1.5 [$-$6.2, +3.5] & 259\\
 & auto:300 & 36.9 & 20{,}425 & 14{,}369 & 1.40 & 38.1 & 58.0 & 55.0 & 60.3 & +5.3 [$-$1.3, +11.9] & 151\\
Cora & khop2:120 & 0.0 & 7{,}007 & 6{,}592 & 1.10 & 67.6 & 66.7 & 66.9 & 65.8 & $-$1.2 [$-$6.2, +4.2] & 260\\
 & auto:300 & 10.6 & 6{,}705 & 6{,}213 & 1.10 & 61.5 & 66.3 & 67.8 & 64.8 & $-$3.0 [$-$9.6, +3.5] & 230\\
Coauthor-CS & khop2:120 & 0.0 & 8{,}662 & 6{,}697 & 1.30 & 59.0 & 60.3 & 58.0 & 58.4 & +0.4 [$-$4.7, +5.4] & 257\\
 & auto:300 & 33.7 & 21{,}083 & 10{,}306 & 2.00 & 38.1 & 61.5 & 52.5 & 61.1 & +8.6 [+1.9, +15.4] & 162\\
\textit{all eight} & khop2:120 & 21.8 & 13{,}779 & 5{,}193 & & 49.7 & 69.5 & 60.7 & 66.8 & +6.1 [+3.9, +8.2] & 1{,}555\\
\textit{all eight} & auto:300 & 61.6 & 33{,}936 & 11{,}155 & & 29.3 & 71.7 & 63.0 & 66.9 & +3.9 [+0.6, +7.2] & 719\\
\textit{all eight} & both & 41.7 & 19{,}723 & 6{,}229 & & 39.5 & 70.6 & 61.4 & 66.8 & +5.4 [+3.6, +7.3] & 2{,}274\\
\midrule
Qwen3-32B, all eight & both & 53.0 & 26{,}587 & 6{,}595 & & 41.7 & 82.3 & 77.0 & 76.9 & $-$0.1 [$-$2.0, +1.8] & 1{,}011\\
Llama-3.1-8B, HuRI & auto:300, 40k window & 9.0 & & & & 51.9 & 36.5 & 64.0 & 62.0 & $-$2.0 [$-$16.0, +12.0] & 50\\
Llama-3.1-8B, HuRI & auto:300, 128k window & 0.0 & & & & 55.8 & 59.0 & 58.1 & 56.4 & $-$1.7 [$-$10.3, +6.0] & 117\\
Llama-3.1-8B, PP-Pathways & auto:300, 40k window & 78.8 & & & & 16.0 & 53.2 & 33.3 & 50.0 & +16.7 [$-$8.3, +41.7] & 24\\
Llama-3.1-8B, PP-Pathways & auto:300, 128k window & 3.2 & & & & 54.5 & 53.8 & 55.2 & 50.0 & $-$5.2 [$-$12.9, +3.4] & 116\\
\bottomrule
\end{tabular}
\end{table}

%% file: tables/tabS4_ladder.tex
\begin{table}[!htbp]
\centering
\caption{\textbf{Which part of the description does the work: the three rendering steps in every study.} Each step is the reasoning-controlled difference between two rungs of the R5 ladder (names $-$ R3 raw metrics; names + evidence $-$ names; BioGlyph $-$ names + evidence) on one common feasible set of all four rungs, with the two-stage (network, question) bootstrap interval; the total is BioGlyph $-$ R3 on the same units. Asterisks mark intervals that exclude zero. From \texttt{ladder.parquet} (scripts/ladder\_report.py).}
\label{tab:s_ladder}
\scriptsize
\setlength{\tabcolsep}{3pt}
\begin{tabular}{l l r r r r r}
\toprule
Study & Question split & Naming & Evidence & Consequence & Total & $n$\\
\midrule
main grid (8 networks, 3 models) & all & $-$3.9 [$-$6.8, +0.3] & +1.6 [$-$0.8, +3.7] & +5.4 [+3.4, +8.4]$^{*}$ & +3.1 [+0.1, +7.6]$^{*}$ & 3{,}012\\
 & direct readout & $-$5.8 [$-$9.1, $-$1.1]$^{*}$ & +1.8 [$-$1.6, +4.7] & +7.2 [+4.8, +10.3]$^{*}$ & +3.1 [$-$0.0, +7.9] & 1{,}812\\
 & compositional & $-$1.4 [$-$7.0, +4.6] & +1.6 [$-$2.3, +5.1] & +3.8 [$-$0.1, +8.8] & +4.0 [$-$1.3, +10.5] & 851\\
 & control & +0.3 [$-$1.6, +1.9] & +0.6 [+0.0, +2.2] & +0.0 [+0.0, +0.0] & +0.9 [+0.0, +2.5] & 349\\
\addlinespace[1pt]
human interactomes, 40k window & all & $-$4.0 [$-$6.9, $-$0.8]$^{*}$ & +0.5 [$-$2.1, +3.1] & +2.8 [+0.4, +5.1]$^{*}$ & $-$0.7 [$-$3.3, +2.1] & 850\\
 & direct readout & $-$3.7 [$-$9.2, +3.6] & +0.0 [$-$4.3, +4.6] & +5.2 [+0.0, +9.3] & +1.5 [$-$2.6, +6.7] & 481\\
 & compositional & $-$6.6 [$-$13.8, +0.0] & +1.5 [$-$3.9, +6.8] & +0.0 [$-$6.1, +5.9] & $-$5.0 [$-$11.4, +0.8] & 259\\
 & control & +0.9 [+0.0, +2.9] & +0.0 [+0.0, +0.0] & $-$0.9 [$-$2.9, +0.0] & +0.0 [+0.0, +0.0] & 110\\
\addlinespace[1pt]
human interactomes, 128k window (Llama) & all & +0.0 [$-$6.8, +6.8] & $-$3.9 [$-$10.2, +1.6] & +3.9 [$-$0.5, +8.4] & +0.0 [$-$4.7, +4.4] & 382\\
 & direct readout & $-$2.5 [$-$13.0, +7.8] & $-$8.0 [$-$15.5, $-$0.5]$^{*}$ & +8.5 [+1.0, +16.1]$^{*}$ & $-$2.0 [$-$9.5, +5.5] & 199\\
 & compositional & +2.9 [$-$7.6, +12.8] & +0.0 [$-$13.7, +13.0] & $-$1.0 [$-$10.8, +9.8] & +1.9 [$-$6.9, +12.0] & 105\\
 & control & +2.6 [$-$1.3, +7.7] & +1.3 [+0.0, +5.3] & $-$1.3 [$-$5.0, +0.0] & +2.6 [+0.0, +7.9] & 78\\
\addlinespace[1pt]
Gemma-3-12B and Mistral-Nemo (8 networks) & all & +1.0 [$-$4.1, +5.5] & $-$3.4 [$-$6.9, +0.4] & +4.5 [+2.2, +7.6]$^{*}$ & +2.1 [$-$1.6, +6.4] & 1{,}978\\
 & direct readout & $-$1.2 [$-$8.9, +5.3] & $-$4.0 [$-$9.8, +1.9] & +9.7 [+6.7, +13.8]$^{*}$ & +4.5 [$-$0.1, +9.6] & 1{,}246\\
 & compositional & +4.1 [$-$1.0, +10.1] & $-$2.8 [$-$6.4, +0.7] & $-$5.7 [$-$9.2, $-$2.5]$^{*}$ & $-$4.4 [$-$10.1, +2.1] & 562\\
 & control & +6.5 [+0.0, +14.4] & $-$0.6 [$-$6.1, +4.1] & +0.0 [$-$4.9, +6.2] & +5.9 [$-$0.8, +15.2] & 170\\
\addlinespace[1pt]
seven further networks (4 models) & all & $-$3.6 [$-$6.1, $-$1.1]$^{*}$ & $-$0.8 [$-$2.5, +1.9] & +5.1 [+3.4, +7.1]$^{*}$ & +0.6 [$-$1.6, +4.0] & 4{,}171\\
 & direct readout & $-$7.8 [$-$11.1, $-$4.0]$^{*}$ & $-$0.6 [$-$3.3, +3.4] & +7.5 [+4.9, +10.2]$^{*}$ & $-$0.9 [$-$3.3, +2.8] & 2{,}582\\
 & compositional & +4.2 [$-$0.2, +9.9] & $-$1.6 [$-$6.2, +2.6] & +1.9 [$-$1.5, +6.8] & +4.4 [$-$0.3, +11.1] & 1{,}131\\
 & control & +0.7 [$-$1.6, +1.9] & $-$0.4 [$-$1.4, +0.0] & $-$0.4 [$-$2.9, +0.7] & $-$0.2 [$-$4.0, +1.3] & 458\\
\addlinespace[1pt]
patient and disease graphs (4 models) & all & $-$9.6 [$-$12.2, $-$7.2]$^{*}$ & +6.3 [+3.4, +8.8]$^{*}$ & +7.4 [+5.5, +9.4]$^{*}$ & +4.0 [+0.5, +7.0]$^{*}$ & 1{,}816\\
 & direct readout & $-$18.6 [$-$22.7, $-$14.9]$^{*}$ & +11.2 [+5.7, +15.2]$^{*}$ & +14.1 [+11.1, +18.2]$^{*}$ & +6.8 [+1.6, +10.5]$^{*}$ & 947\\
 & compositional & $-$3.3 [$-$9.5, +3.8] & +4.7 [+0.9, +8.8]$^{*}$ & $-$1.3 [$-$5.4, +3.1] & +0.2 [$-$4.4, +4.6] & 549\\
 & control & +5.9 [+3.1, +9.1]$^{*}$ & $-$5.6 [$-$10.0, $-$2.2]$^{*}$ & +2.2 [+0.6, +4.1]$^{*}$ & +2.5 [$-$1.6, +5.9] & 320\\
\addlinespace[1pt]
harder patient questions & all & +9.4 [+1.9, +16.1]$^{*}$ & +0.9 [$-$2.8, +6.2] & +0.0 [$-$3.1, +3.8] & +10.3 [+5.0, +16.7]$^{*}$ & 320\\
 & direct readout & +9.4 [+1.9, +16.1]$^{*}$ & +0.9 [$-$2.8, +6.2] & +0.0 [$-$3.1, +3.8] & +10.3 [+5.0, +16.7]$^{*}$ & 320\\
\addlinespace[1pt]
PPR retrieval (8 networks) & all & +2.9 [$-$0.8, +7.0] & +2.6 [$-$0.1, +5.6] & +0.7 [$-$2.2, +3.3] & +6.2 [+2.7, +9.5]$^{*}$ & 1{,}399\\
 & compositional & +2.9 [$-$0.8, +7.0] & +2.6 [$-$0.1, +5.6] & +0.7 [$-$2.2, +3.3] & +6.2 [+2.7, +9.5]$^{*}$ & 1{,}399\\
\addlinespace[1pt]
\bottomrule
\end{tabular}
\end{table}

%% file: tables/tabS5_rungs.tex
\begin{table}[!htbp]
\centering
\caption{\textbf{The four rungs on every network, two 8B models pooled.} System accuracy of R3 raw metrics, R5 names only, R5 names + evidence and R5 BioGlyph, then the same four on the common feasible units of all four rungs (reasoning-controlled), the consequence step (BioGlyph $-$ names + evidence) with its bootstrap interval, and the number of feasible units. Rows marked s7ext are the harder question set on the two patient graphs.}
\label{tab:s_rungs}
\scriptsize
\setlength{\tabcolsep}{2.6pt}
\begin{tabular}{l rrrr rrrr r r}
\toprule
 & \multicolumn{4}{c}{System accuracy (\%)} & \multicolumn{4}{c}{Reasoning-controlled (\%)} & Consequence & \\
\cmidrule(lr){2-5}\cmidrule(lr){6-9}
Network & R3 & names & +evid. & BioGlyph & R3 & names & +evid. & BioGlyph & $\Delta$ [95\% CI] & $n$\\
\midrule
STRING-Yeast & 36.1 & 66.8 & 66.7 & 75.6 & 76.9 & 69.9 & 68.8 & 82.7 & +13.9 [+6.9, +21.4] & 208\\
ChCh-Miner & 25.3 & 66.8 & 70.4 & 77.7 & 60.0 & 65.4 & 66.2 & 74.6 & +8.5 [+0.8, +16.2] & 163\\
ego-Facebook & 29.3 & 71.3 & 72.8 & 79.3 & 68.2 & 71.5 & 71.5 & 80.8 & +9.3 [+1.3, +17.2] & 180\\
email-Eu-core & 26.0 & 68.6 & 70.8 & 78.8 & 66.4 & 62.3 & 64.8 & 74.6 & +9.8 [+1.6, +18.0] & 156\\
HuRI & 49.7 & 60.7 & 59.5 & 48.4 & 61.4 & 59.1 & 57.1 & 62.0 & +4.9 [+1.0, +9.1] & 359\\
PP-Pathways & 32.4 & 58.0 & 56.6 & 57.4 & 51.9 & 54.3 & 52.7 & 54.3 & +1.6 [$-$2.9, +6.2] & 285\\
STRING-Human & 41.2 & 61.1 & 61.9 & 69.7 & 65.0 & 58.8 & 61.5 & 65.8 & +4.2 [$-$0.8, +9.2] & 302\\
STRING-Ecoli & 39.3 & 69.6 & 72.0 & 79.3 & 81.2 & 74.0 & 80.1 & 87.8 & +7.7 [+2.2, +13.3] & 218\\
DG-AssocMiner & 59.5 & 62.5 & 60.3 & 48.9 & 67.8 & 63.7 & 61.9 & 67.2 & +5.3 [+1.6, +9.4] & 363\\
HuDiNe & 73.1 & 62.2 & 68.3 & 75.6 & 76.4 & 65.0 & 73.3 & 79.3 & +6.0 [+2.7, +9.6] & 509\\
TCGA-GBM-PSN & 75.2 & 62.7 & 70.2 & 77.4 & 77.6 & 64.6 & 71.8 & 80.2 & +8.4 [+4.1, +12.7] & 439\\
GBM-scRNA & 15.9 & 26.1 & 28.4 & 34.1 & 26.3 & 15.8 & 26.3 & 26.3 & +0.0 [+0.0, +0.0] & 20\\
Amazon-Photo & 33.2 & 59.9 & 61.7 & 71.8 & 55.8 & 62.5 & 59.8 & 70.5 & +10.7 [+5.4, +16.1] & 264\\
Wiki-Vote & 28.8 & 68.8 & 66.8 & 73.2 & 63.3 & 68.4 & 63.3 & 72.8 & +9.5 [+3.2, +16.5] & 191\\
email-Enron & 36.5 & 69.4 & 74.4 & 79.6 & 73.8 & 71.2 & 72.8 & 79.6 & +6.8 [+0.5, +13.1] & 230\\
Cora & 64.6 & 62.0 & 61.4 & 66.5 & 69.2 & 61.8 & 63.4 & 67.6 & +4.1 [+0.2, +8.3] & 491\\
CiteSeer & 76.9 & 66.3 & 67.1 & 75.6 & 79.6 & 71.1 & 70.0 & 78.2 & +8.2 [+4.7, +11.8] & 510\\
ogbn-arxiv & 49.2 & 56.1 & 57.2 & 59.1 & 59.5 & 55.5 & 57.4 & 60.1 & +2.7 [$-$0.8, +6.2] & 426\\
Coauthor-CS & 48.6 & 55.9 & 57.4 & 60.9 & 57.0 & 57.3 & 57.5 & 60.5 & +3.0 [$-$0.8, +6.8] & 417\\
Power-Grid & 69.4 & 63.5 & 60.7 & 64.7 & 68.9 & 64.4 & 62.1 & 65.2 & +3.1 [$-$0.2, +6.4] & 544\\
HuDiNe (s7ext) & 14.8 & 19.3 & 22.7 & 25.0 & 7.8 & 14.1 & 17.2 & 18.8 & +1.6 [$-$3.1, +6.2] & 70\\
TCGA-GBM-PSN (s7ext) & 23.9 & 38.6 & 37.5 & 38.6 & 20.0 & 36.9 & 35.4 & 38.5 & +3.1 [$-$3.1, +9.2] & 71\\
\bottomrule
\end{tabular}
\end{table}

%% file: tables/tabS6_more.tex
\begin{table}[!htbp]
\centering
\caption{\textbf{The sixteen further networks: system accuracy of every arm and the controlled advantage of R5 BioGlyph.} Two 8B models pooled (Qwen3-8B and Llama-3.1-8B are present in every study). The R4 column is the mean of the four encoders. The last three columns are R5 BioGlyph minus R3 raw metrics and minus R2 adjacency text on the answerable questions both arms fitted and finished, and R3's overflow share. A dash marks an arm not run on that network.}
\label{tab:s_more}
\scriptsize
\setlength{\tabcolsep}{2.4pt}
\begin{tabular}{l rrrrrrrrr rr r}
\toprule
Network & R0 & R1 & R2 & R3 & R4 & names & +evid. & BioGlyph & opaque & R5 $-$ R3 & R5 $-$ R2 & R3 overflow (\%)\\
\midrule
HuRI & 54.0 & 58.8 & 61.5 & 49.7 & 51.6 & 60.7 & 59.5 & 48.4 & 55.1 & +0.3 & $-$3.1 & 20.2\\
PP-Pathways & 53.5 & 51.3 & 50.0 & 32.4 & 49.4 & 58.0 & 56.6 & 57.4 & 55.1 & +1.8 & $-$6.3 & 45.8\\
STRING-Human & -- & -- & 48.1 & 41.2 & 47.9 & 61.1 & 61.9 & 69.7 & -- & +1.0 & +9.0 & 39.4\\
STRING-Ecoli & -- & -- & 39.4 & 39.3 & 50.7 & 69.6 & 72.0 & 79.3 & -- & +8.1 & +16.3 & 54.2\\
DG-AssocMiner & -- & -- & 64.3 & 59.5 & 51.3 & 62.5 & 60.3 & 48.9 & -- & $-$1.7 & $-$4.0 & 8.2\\
HuDiNe & -- & -- & 59.5 & 73.1 & 44.7 & 62.2 & 68.3 & 75.6 & -- & +1.4 & +12.8 & 0.6\\
TCGA-GBM-PSN & -- & -- & 65.3 & 75.2 & 46.3 & 62.7 & 70.2 & 77.4 & -- & +2.4 & +8.0 & 0.0\\
GBM-scRNA & -- & -- & 12.5 & 15.9 & 18.2 & 26.1 & 28.4 & 34.1 & -- & +0.0 & +18.0 & 61.4\\
Amazon-Photo & 47.8 & 48.9 & 50.2 & 33.2 & 48.2 & 59.9 & 61.7 & 71.8 & 53.0 & +17.1 & +11.0 & 47.1\\
Wiki-Vote & 58.3 & 45.4 & 43.1 & 28.8 & 52.0 & 68.8 & 66.8 & 73.2 & 57.7 & +11.5 & +4.3 & 60.1\\
email-Enron & -- & -- & 43.1 & 36.5 & 48.8 & 69.4 & 74.4 & 79.6 & -- & +5.4 & +19.6 & 55.0\\
Cora & 52.7 & 60.3 & 65.1 & 64.6 & 50.1 & 62.0 & 61.4 & 66.5 & 56.1 & $-$2.0 & $-$4.4 & 5.3\\
CiteSeer & -- & -- & 65.4 & 76.9 & 50.5 & 66.3 & 67.1 & 75.6 & -- & $-$3.3 & +6.2 & 2.1\\
ogbn-arxiv & 52.1 & 56.7 & 60.1 & 49.2 & 48.8 & 56.1 & 57.2 & 59.1 & 51.9 & +1.0 & $-$4.9 & 18.6\\
Coauthor-CS & 48.7 & 56.4 & 57.5 & 48.6 & 46.0 & 55.9 & 57.4 & 60.9 & 51.0 & +3.6 & $-$1.3 & 16.8\\
Power-Grid & -- & -- & 63.6 & 69.4 & 50.5 & 63.5 & 60.7 & 64.7 & -- & $-$4.7 & $-$3.3 & 0.0\\
HuDiNe (s7ext) & -- & -- & 14.8 & 14.8 & 13.9 & 19.3 & 22.7 & 25.0 & -- & +10.0 & +10.8 & 0.0\\
TCGA-GBM-PSN (s7ext) & -- & -- & 35.2 & 23.9 & 18.5 & 38.6 & 37.5 & 38.6 & -- & +16.9 & $-$3.2 & 0.0\\
\bottomrule
\end{tabular}
\end{table}

%% file: tables/tabS9_controls.tex
\begin{table}[!htbp]
\centering
\caption{\textbf{Further controls.} Top: hosted frontier readers on three networks (khop2:120; nothing overflows), system accuracy per arm. Middle: two further open model families on the eight main networks (system accuracy pooled over the eight, and the controlled advantage of R5 BioGlyph over R3 and R2). Bottom: fusion of learned roles with BioGlyph (R6) against R5 alone, and the oracle-free PPR retrieval.}
\label{tab:s_controls}
\scriptsize
\setlength{\tabcolsep}{3pt}
\begin{tabular}{l l r r r r r r r}
\toprule
\multicolumn{9}{l}{\textit{Frontier readers}}\\
Reader & Network & R2 & R3 & names & +evid. & BioGlyph & R5 $-$ R3 & R5 $-$ R2\\
\midrule
claude-fable-5 & Cora & 81.5 & 81.5 & 76.5 & 80.2 & 81.5 & +0.0 & +0.0\\
claude-fable-5 & HuDiNe & 83.2 & 84.1 & 75.7 & 77.6 & 77.6 & $-$6.5 & $-$5.6\\
claude-fable-5 & STRING-Yeast & 81.2 & 79.7 & 78.1 & 79.7 & 79.7 & +0.0 & $-$1.5\\
claude-opus-5 & Cora & 75.4 & 76.8 & 73.9 & 76.1 & 74.6 & $-$2.2 & $-$0.8\\
claude-opus-5 & HuDiNe & 84.7 & 85.3 & 76.0 & 80.0 & 78.7 & $-$6.6 & $-$6.0\\
claude-opus-5 & STRING-Yeast & 87.8 & 87.1 & 85.7 & 84.4 & 85.0 & $-$2.1 & $-$2.8\\
gpt-5.5 & Cora & 76.9 & 77.6 & 75.0 & 77.6 & 78.2 & +0.6 & +1.3\\
gpt-5.5 & HuDiNe & 84.6 & 85.9 & 78.2 & 82.7 & 82.1 & $-$3.8 & $-$2.5\\
gpt-5.5 & STRING-Yeast & 87.2 & 85.9 & 80.8 & 84.6 & 85.3 & $-$0.6 & $-$1.9\\
gpt-5.6-sol & Cora & 76.3 & 77.6 & 75.0 & 77.6 & 77.6 & +0.0 & +1.3\\
gpt-5.6-sol & HuDiNe & 85.3 & 85.9 & 80.1 & 82.1 & 81.4 & $-$4.5 & $-$3.9\\
gpt-5.6-sol & STRING-Yeast & 87.2 & 85.9 & 82.7 & 85.3 & 85.9 & +0.0 & $-$1.3\\
\midrule
\multicolumn{9}{l}{\textit{Other model families, eight main networks}}\\
Model & & R2 & R3 & R4 mean & names & +evid. & BioGlyph & R5 $-$ R3 / R2 (ctrl)\\
\midrule
gemma-3-12b-it & & 49.2 & 35.2 & 53.9 & 66.5 & 64.0 & 68.0 & $-$2.3 / +2.8\\
Mistral-Nemo-Instruct-2407 & & 41.5 & 30.1 & 48.8 & 57.7 & 57.9 & 60.3 & +6.5 / +8.5\\
\midrule
\multicolumn{9}{l}{\textit{Fusion with learned roles (R6), two 8B models pooled; R5 alone = 70.6}}\\
Arm & & sys. & arm $-$ R5 (sys) & arm $-$ R5 (ctrl) & \multicolumn{4}{l}{}\\
\midrule
R6 GCN+BioGlyph & & 69.7 & $-$0.9 & $-$1.6 & \multicolumn{4}{l}{}\\
R6 GraphSAGE+BioGlyph & & 70.0 & $-$0.6 & $-$1.1 & \multicolumn{4}{l}{}\\
R6 GAT+BioGlyph & & 70.0 & $-$0.6 & $-$1.2 & \multicolumn{4}{l}{}\\
R6 GIN+BioGlyph & & 69.6 & $-$1.0 & $-$1.8 & \multicolumn{4}{l}{}\\
R6 compiled-global+BioGlyph & & 70.0 & $-$0.6 & $-$1.0 & \multicolumn{4}{l}{}\\
\midrule
\multicolumn{9}{l}{\textit{PPR retrieval (ppr:120), two 8B models pooled}}\\
Arm & & sys. & overflow \% & \multicolumn{5}{l}{}\\
\midrule
R0 no graph & & 29.1 & 0.0 & \multicolumn{5}{l}{}\\
R1 Edge list & & 42.6 & 1.1 & \multicolumn{5}{l}{}\\
R2 adjacency text & & 47.5 & 4.0 & \multicolumn{5}{l}{}\\
R3 Raw metrics & & 33.0 & 27.2 & \multicolumn{5}{l}{}\\
R5 names only & & 47.1 & 0.0 & \multicolumn{5}{l}{}\\
R5 names + evidence & & 48.5 & 0.0 & \multicolumn{5}{l}{}\\
R5 BioGlyph & & 49.9 & 0.0 & \multicolumn{5}{l}{}\\
R5 BioGlyph (opaque) & & 38.6 & 0.0 & \multicolumn{5}{l}{}\\
\multicolumn{9}{l}{controlled R5 $-$ R3 +5.8 [+2.7, +8.7] ($n$ = 1{,}562); R5 $-$ R2 $-$1.3 [$-$5.8, +3.5]}\\
\bottomrule
\end{tabular}
\end{table}

%% file: tables/tabS7_conversation.tex
\begin{table}[!htbp]
\centering
\caption{\textbf{The conversation studies.} Top: the eight-turn threads (built for the compiled description alone, because their lookup turns ask about roles the description names), accuracy by turn class, which are never pooled, and what the push-back did against the turn it disputes. Middle: the three-turn study, R5 BioGlyph against R3 raw metrics on the same conversations, accuracy at each turn on the conversations that fitted and finished, and the number of R3 conversations that never started. Bottom: the same questions asked in benchmark style and as prose.}
\label{tab:s_conv}
\scriptsize
\setlength{\tabcolsep}{3pt}
\begin{tabular}{l l r r r r r r r r}
\toprule
\multicolumn{10}{l}{\textit{Eight-turn threads (719 threads per model)}}\\
Network & Model & Threads & Graded & Lookup & Push-back & Held & Gave up & Corrected & Wrong twice\\
\midrule
HuDiNe & Llama-3.1-8B & 250 & 64.6 & 71.6 & 50.6 & 30.9 & 28.5 & 19.7 & 20.9\\
HuDiNe & Qwen3-8B & 250 & 89.5 & 82.0 & 65.6 & 58.7 & 28.9 & 7.0 & 5.4\\
STRING-Yeast & Llama-3.1-8B & 250 & 59.4 & 52.0 & 43.4 & 21.7 & 27.7 & 21.7 & 28.9\\
STRING-Yeast & Qwen3-8B & 250 & 83.8 & 78.4 & 69.2 & 62.3 & 23.1 & 7.3 & 7.3\\
email-Eu-core & Llama-3.1-8B & 219 & 60.0 & 56.6 & 36.1 & 22.5 & 36.2 & 13.3 & 28.0\\
email-Eu-core & Qwen3-8B & 219 & 90.4 & 69.4 & 61.6 & 56.0 & 31.5 & 6.0 & 6.5\\
\midrule
\multicolumn{10}{l}{\textit{Three-turn study (map at turn 1, a facts-only follow-up, a push-back)}}\\
Network & Model & Arm & Turn 1 & Turn 2 & Turn 3 & $n$ & Held & Gave up & Wrong twice\\
\midrule
HuDiNe & Llama-3.1-8B & R3 raw metrics & 42.3 & 46.6 & 17.0 & 137 & 8.8 & 37.4 & 45.6\\
HuDiNe & Llama-3.1-8B & R5 BioGlyph & 44.2 & 62.8 & 19.6 & 138 & 10.1 & 52.7 & 27.7\\
HuDiNe & Qwen3-8B & R3 raw metrics & 75.4 & 89.2 & 84.5 & 138 & 83.8 & 5.4 & 10.1\\
HuDiNe & Qwen3-8B & R5 BioGlyph & 76.1 & 95.3 & 89.2 & 138 & 85.8 & 9.5 & 1.4\\
STRING-Yeast & Llama-3.1-8B & R3 raw metrics & 50.0 & 34.3 & 6.1 & 60 & 4.5 & 30.3 & 63.6\\
STRING-Yeast & Llama-3.1-8B & R5 BioGlyph & 47.2 & 41.8 & 9.2 & 89 & 4.1 & 37.8 & 53.1\\
STRING-Yeast & Qwen3-8B & R3 raw metrics & 62.2 & 52.1 & 50.0 & 45 & 45.8 & 6.2 & 43.8\\
STRING-Yeast & Qwen3-8B & R5 BioGlyph & 75.3 & 76.5 & 74.5 & 89 & 68.4 & 8.2 & 17.3\\
email-Eu-core & Llama-3.1-8B & R3 raw metrics & 54.0 & 35.8 & 9.9 & 113 & 3.3 & 32.5 & 57.5\\
email-Eu-core & Llama-3.1-8B & R5 BioGlyph & 52.2 & 49.2 & 16.4 & 113 & 10.7 & 38.5 & 45.1\\
email-Eu-core & Qwen3-8B & R3 raw metrics & 76.1 & 89.9 & 86.8 & 67 & 86.8 & 2.9 & 10.3\\
email-Eu-core & Qwen3-8B & R5 BioGlyph & 67.5 & 94.3 & 88.5 & 114 & 87.7 & 6.6 & 4.9\\
\multicolumn{10}{l}{R3 conversations that never started (turn 1 over budget): 131 of 736 (17.8\%); R5: 0}\\
\midrule
\multicolumn{10}{l}{\textit{Benchmark style against prose (same questions; all answerable with overflow scored wrong, and finished replies only)}}\\
Network & model & arm & bench. all & prose all & bench. finished & prose finished & \multicolumn{3}{l}{}\\
\midrule
HuDiNe & Llama-3.1-8B & R3 raw metrics & 60.0 & 50.0 & 68.9 & 49.2 & \multicolumn{3}{l}{}\\
HuDiNe & Llama-3.1-8B & R5 BioGlyph & 65.7 & 59.3 & 68.5 & 59.9 & \multicolumn{3}{l}{}\\
HuDiNe & Qwen3-8B & R3 raw metrics & 74.3 & 71.4 & 74.3 & 71.4 & \multicolumn{3}{l}{}\\
HuDiNe & Qwen3-8B & R5 BioGlyph & 81.4 & 77.9 & 81.4 & 77.9 & \multicolumn{3}{l}{}\\
STRING-Yeast & Llama-3.1-8B & R3 raw metrics & 34.3 & 31.4 & 63.2 & 46.7 & \multicolumn{3}{l}{}\\
STRING-Yeast & Llama-3.1-8B & R5 BioGlyph & 56.4 & 59.3 & 62.6 & 60.7 & \multicolumn{3}{l}{}\\
STRING-Yeast & Qwen3-8B & R3 raw metrics & 35.7 & 35.0 & 82.0 & 80.3 & \multicolumn{3}{l}{}\\
STRING-Yeast & Qwen3-8B & R5 BioGlyph & 87.9 & 80.7 & 87.9 & 80.7 & \multicolumn{3}{l}{}\\
email-Eu-core & Llama-3.1-8B & R3 raw metrics & 46.4 & 50.0 & 56.6 & 55.2 & \multicolumn{3}{l}{}\\
email-Eu-core & Llama-3.1-8B & R5 BioGlyph & 60.7 & 61.4 & 65.4 & 62.2 & \multicolumn{3}{l}{}\\
email-Eu-core & Qwen3-8B & R3 raw metrics & 8.6 & 8.6 & 80.0 & 80.0 & \multicolumn{3}{l}{}\\
email-Eu-core & Qwen3-8B & R5 BioGlyph & 76.4 & 74.3 & 76.4 & 74.3 & \multicolumn{3}{l}{}\\
\bottomrule
\end{tabular}
\end{table}

%% file: tables/tabS8_bio.tex
\begin{table}[!htbp]
\centering
\caption{\textbf{Biological validation.} Top: SGD essentiality of yeast proteins by BioGlyph role, with the odds ratio against the background rate, and the essential share of the same number of proteins picked by a single centrality. Middle: the Reactome-FI knockout screen, deterministic leg (top-ranked cut nodes and how many proteins they detach) and the language-model leg (each arm asked how many proteins detach when the named protein is removed from the retrieved module; a count within $\pm$2 of the oracle is scored right).}
\label{tab:s_bio}
\scriptsize
\setlength{\tabcolsep}{3pt}
\begin{tabular}{l r r r r l}
\toprule
\multicolumn{6}{l}{\textit{STRING-Yeast, SGD essentiality (background 30.9\%)}}\\
Role & $n$ & Essential (\%) & Odds ratio [95\% CI] & $p$ (Fisher) & Matched-size baseline\\
\midrule
\glyph{HUB} & 190 & 42.1 & 1.67 [1.24, 2.26] & $8.8\times10^{-4}$ & top-190 degree 42.1\%\\
\glyph{CUT\_NODE} & 363 & 35.0 & 1.23 [0.98, 1.55] & $8.2\times10^{-2}$ & \\
\glyph{COMMUNITY\_CORE} & 522 & 46.9 & 2.27 [1.88, 2.75] & $8.8\times10^{-17}$ & \\
\glyph{BOUNDARY\_NODE} & 473 & 34.0 & 1.18 [0.96, 1.45] & $1.2\times10^{-1}$ & \\
\glyph{CROSS\_COMMUNITY\_CONNECTOR} & 85 & 57.6 & 3.14 [2.03, 4.86] & $2.9\times10^{-7}$ & top-85 degree 24.7\%, betweenness 56.5\%\\
\glyph{PERIPHERAL} & 674 & 12.9 & 0.27 [0.21, 0.34] & $3.3\times10^{-33}$ & \\
\midrule
\multicolumn{6}{l}{\textit{Reactome-FI knockout screen, deterministic leg; every target below is a \glyph{CUT\_NODE}}}\\
Protein & Rank & Detached proteins & DepMap dependent (\%) & Class & \multicolumn{1}{l}{}\\
\midrule
EP300 & 1 & 76 & 28.4 & selective & \\
GPLD1 & 2 & 53 & 0.0 & too few & \\
CTCF & 3 & 51 & 99.7 & common & \\
YY1 & 4 & 29 & 86.8 & common & \\
TP53 & 5 & 26 & 0.3 & too few & \\
SPI1 & 6 & 21 & 2.6 & too few & \\
ESR1 & 7 & 21 & 1.1 & too few & \\
HNF4A & 8 & 21 & 0.4 & too few & \\
SIX5 & 9 & 21 & 0.1 & too few & \\
JUN & 10 & 20 & 17.5 & selective & \\
\multicolumn{6}{l}{cut nodes selectively essential: 13.3\% against 8.8\% of other screened proteins (fold 1.51, one-sided Fisher $p=2.4\times10^{-3}$)}\\
\midrule
\multicolumn{6}{l}{\textit{Language-model leg, 60 modules, three models pooled}}\\
Arm & within $\pm$2 \% & median $|$error$|$ & faithful \% & \multicolumn{2}{l}{}\\
\midrule
R1 Edge list & 22.2 & 9.0 & 72.3 & \multicolumn{2}{l}{}\\
R2 adjacency text & 36.7 & 4.0 & 69.5 & \multicolumn{2}{l}{}\\
R3 Raw metrics & 10.6 & 13.0 & 77.9 & \multicolumn{2}{l}{}\\
R5 BioGlyph & 47.8 & 2.0 & 87.1 & \multicolumn{2}{l}{}\\
\bottomrule
\end{tabular}
\end{table}

%% file: tables/tabS10_cko.tex
\begin{table}[p]
\centering
\caption{\textbf{The candidate-knockout screen: picking the biologically labeled protein out of five anonymous candidates.} One question shows one retrieved module and five candidate node ids, exactly one of which carries the label; the pick is graded by SGD or DepMap, never by the graph oracle. Model rows pool Qwen3-8B and Llama-3.1-8B (Wilson 95\% intervals); policy rows are the deterministic reference picks stored per question at build time. The three experiments are analyzed separately, as pre-registered.}
\label{tab:s_cko}
\scriptsize
\setlength{\tabcolsep}{4pt}
\begin{tabular}{l r r r r r r}
\toprule
 & \multicolumn{2}{c}{STRING-Yeast / SGD} & \multicolumn{2}{c}{Reactome / selective} & \multicolumn{2}{c}{Reactome / common}\\
Arm or policy & Accuracy & 95\% CI & Accuracy & 95\% CI & Accuracy & 95\% CI\\
\midrule
R0 no graph (leak control) & 7.8 & [5.9, 10.3] & 9.5 & [7.4, 12.1] & 8.5 & [6.5, 11.0]\\
R0 degree only & 24.5 & [21.2, 28.1] & 39.3 & [35.5, 43.3] & 46.2 & [42.2, 50.2]\\
R2 adjacency text & 24.7 & [21.4, 28.3] & 33.0 & [29.4, 36.9] & 46.5 & [42.5, 50.5]\\
R3 raw measurements & 24.7 & [21.4, 28.3] & 34.8 & [31.1, 38.7] & 40.8 & [37.0, 44.8]\\
R5 BioGlyph & 27.2 & [23.8, 30.9] & 36.7 & [32.9, 40.6] & 39.3 & [35.5, 43.3]\\
\midrule
policy: chance & 18.3 &  & 16.0 &  & 19.3 & \\
policy: degree & 25.0 &  & 42.0 &  & 47.3 & \\
policy: betweenness & 29.3 &  & 35.0 &  & 37.0 & \\
policy: glyph priority & 25.7 &  & 37.3 &  & 35.3 & \\
policy: region degree & 23.0 &  & 32.3 &  & 37.7 & \\
\bottomrule
\end{tabular}
\end{table}